\documentclass[accepted]{article}

\usepackage{microtype}
\usepackage{graphicx}
\usepackage{subcaption}
\usepackage{booktabs} 

\usepackage{hyperref}
\usepackage{color}
\usepackage{xspace}
\usepackage{mathtools,soul}
\usepackage[flushleft]{threeparttable}
\usepackage{multirow}
\usepackage{enumitem}
\usepackage{xurl}
\usepackage{caption}
\usepackage{pgfplots}
\usepackage[raggedrightboxes]{ragged2e}
\usepackage{makecell, rotating}
\usepackage[justification=centering]{caption}
\usepackage{makecell}
\usepackage{tipa}
\usepackage{xr}
\usepackage{tipa}

\newcommand{\xn}[1]{{\small\color{red}{\xspace#1}}}

\newcommand{\AVE}{\mbox{$\mathop{\mathtt{AVE}}\limits$}\xspace}
\newcommand{\IRP}{\mbox{$\mathop{\mathtt{PRP}}\limits$}\xspace}
\newcommand{\EM}{\mbox{$\mathop{\mathtt{PM}}\limits$}\xspace}

\newcommand{\SA}{\mbox{$\mathop{\mathtt{SA}}\limits$}\xspace}
\newcommand{\SR}{\mbox{$\mathop{\mathtt{SR}}\limits$}\xspace}

\newcommand{\MPC}{\mbox{$\mathop{\mathtt{MPC}}\limits$}\xspace}
\newcommand{\PSI}{\mbox{$\mathop{\mathtt{PSI}}\limits$}\xspace}
\newcommand{\QPR}{\mbox{$\mathop{\mathtt{QPR}}\limits$}\xspace}

\newcommand{\AP}{\mbox{$\mathop{\mathtt{AP}}\limits$}\xspace}
\newcommand{\AG}{\mbox{$\mathop{\mathtt{AG}}\limits$}\xspace}

\newcommand{\dataset}{\mbox{$\mathop{\mathtt{ECInstruct}}\limits$}\xspace}
\newcommand{\method}{\mbox{$\mathop{\mathtt{eCeLLM}}\limits$}\xspace}

\newcommand{\methodL}{\mbox{$\mathop{\mathtt{\method\text{-}L}}\limits$}\xspace}
\newcommand{\methodB}{\mbox{$\mathop{\mathtt{\method\text{-}M}}\limits$}\xspace}
\newcommand{\methodS}{\mbox{$\mathop{\mathtt{\method\text{-}S}}\limits$}\xspace}

\newcommand{\hs}[1]{\textcolor{magenta}{{[HS: #1]}}}
\newcommand{\rc}[1]{\textcolor{cyan}{{[Ron: #1]}}}

\usepackage[accepted]{icml2024}

\usepackage{amsmath}
\usepackage{amssymb}
\usepackage{mathtools}
\usepackage{amsthm}
\usepackage{listings}
\usepackage{inconsolata}
\usepackage{pifont}

\usepackage[capitalize,noabbrev]{cleveref}

\theoremstyle{plain}

\theoremstyle{definition}

\theoremstyle{remark}

\usepackage[textsize=tiny]{todonotes}

\graphicspath{ {./figures/} {./plots/} }

\icmltitlerunning{\method: Generalizing Large Language Models for \mbox{E-commerce} from Large-scale, 
High-quality Instruction Data}

\begin{document}
\twocolumn[
\icmltitle{\method: Generalizing Large Language Models for \mbox{E-commerce} \\from Large-scale, 
High-quality Instruction Data}




\icmlsetsymbol{equal}{*}

\begin{icmlauthorlist}
\icmlauthor{Bo Peng}{equal,1}
\icmlauthor{Xinyi Ling}{equal,1}
\icmlauthor{Ziru Chen}{1}
\icmlauthor{Huan Sun}{1,2}
\icmlauthor{Xia Ning}{1,2,3}
\end{icmlauthorlist}

\icmlaffiliation{1}{Department of Computer Science and Engineering, The Ohio State University, USA.}
\icmlaffiliation{2}{ Translational Data Analytics Institute, The Ohio State University, USA.}
\icmlaffiliation{3}{Department of Biomedical Informatics, The Ohio State University, USA}

\icmlcorrespondingauthor{Xia Ning}{ning.104@osu.edu}

\icmlkeywords{Machine Learning, ICML}

\vskip 0.3in
]



\printAffiliationsAndNotice{\icmlEqualContribution} 

\begin{abstract}
%
%
%
%

%
%
%
%
With tremendous efforts in developing effective \mbox{e-commerce} models, 
conventional \mbox{e-commerce} models 
show limited success in generalist \mbox{e-commerce} modeling, 
and suffer from unsatisfactory performance on new users and new products -- 
a typical out-of-domain generalization challenge. 
Meanwhile, large language models (LLMs) demonstrate
outstanding performance in generalist modeling and \mbox{out-of-domain generalizability} in many fields.
%
%
Toward fully unleashing their power for \mbox{e-commerce}, 
in this paper, 
we construct \dataset, the first \mbox{open-sourced}, \mbox{large-scale}, and \mbox{high-quality} 
benchmark instruction dataset for \mbox{e-commerce}.
%
%
Leveraging \dataset, we develop \method, a series of 
\mbox{e-commerce} LLMs, by instruction-tuning general-purpose LLMs.
%
Our comprehensive experiments and evaluation demonstrate that 
%
\method models substantially outperform baseline models, including the most advanced \mbox{GPT-4},
and the state-of-the-art task-specific models in \mbox{in-domain} evaluation. 
Moreover, \method exhibits excellent generalizability to out-of-domain settings, 
including unseen products and unseen instructions, highlighting its superiority as a generalist \mbox{e-commerce} model.
Both the \dataset dataset and the \method models show great potential in 
empowering versatile and effective LLMs for \mbox{e-commerce}.
%
%
%
%
\dataset and \method models are publicly accessible through 
\href{https://ninglab.github.io/eCeLLM/}{\textcolor{blue}{https://ninglab.github.io/eCeLLM/}}.
\end{abstract}

%
%
%


\vspace{-25pt}

\section{Introduction}
\label{sec:introduction}
\begin{figure*}
    \centering
    \vspace{-3pt}
    \includegraphics[width=.9\linewidth]{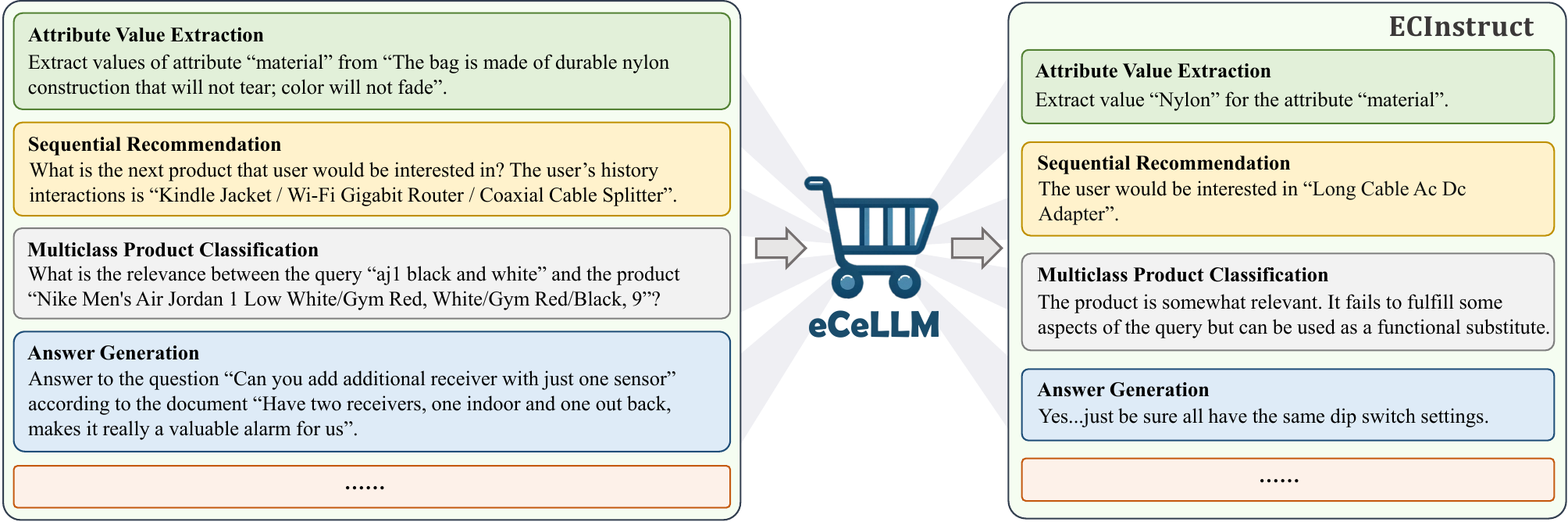} 
    \vspace{-6pt}
    \caption{Overall scheme of \method instruction-tuned with \dataset}
    \label{fig:overview}
    \vspace{-12pt}    
\end{figure*}



The Internet's evolution and the rise of the digital economy have made \mbox{e-commerce} an integral part of daily life, drawing considerable attention from researchers. 
In recent years, tremendous research efforts have been dedicated to developing effective e-commerce models~\cite{yang2022mave,geng2022recommendation}.
Though promising, 
conventional \mbox{e-commerce} models generally suffer from two issues.
\textbf{(1)} Limited success in generalist \mbox{e-commerce} modeling~\cite{Zhang2023multitask}:
Conventional \mbox{e-commerce} models are typically task-specific (e.g., \mbox{gSASRec} for 
sequential recommendation~\cite{petrov2023gsasrec}, \mbox{SUOpenTag} for attribute value extraction~\cite{xu-etal-2019-scaling}).
However, contemporary \mbox{e-commerce} platforms, such as Amazon and eBay, are 
highly complex and consistently expanding, with many interdependent tasks. 
In this context, generalist modeling is particularly suitable and desired on these platforms
due to its cost-effectiveness and extensibility, 
while the task-specific modeling scheme struggles with scalability. 
\textbf{(2)} Unsatisfactory performance on new users and new products~\cite{hou2023large}: 
%
Cold start (i.e., performing \mbox{e-commerce} tasks on new users or new products)~\cite{LIKA20142065}, 
the typical \mbox{out-of-domain} (OOD) generalization challenge in \mbox{e-commerce},
has been a long-standing and difficult problem~\cite{ding2021zero,yang2022mave}.
Existing task-specific \mbox{e-commerce} models are typically tailored to existing users and 
products and lack the ability to effectively extrapolate to new users and new products~\cite{LIKA20142065}. 
However, new users and new products are very common and highly wanted in the dynamic landscape of \mbox{e-commerce}. 
Recently, large language models (LLMs), 
such as 
\mbox{GPT-4}~\cite{GPT4}, Claude 2~\cite{Claude2}, \mbox{Gemini}~\cite{team2023gemini}, and Llama 2~\cite{touvron2023llama}, 
have demonstrated exceptional performance in natural language processing~\cite{zhao2023survey}, information retrieval~\cite{spatharioti2023comparing}, and many other fields~\cite{frieder2023large,zeng2023large}.  
However, 
their power for \mbox{e-commerce} is not fully unleashed. 
While limited efforts~\cite{yang2023large,zhang2023chatgpt, li2023ecomgpt} 
have been dedicated to leveraging LLMs for \mbox{e-commerce}, most of them focus on studying the utility of pre-trained LLMs in single or homogenous tasks (e.g., e-commerce authoring).
Some recent efforts employ instruction tuning to adapt LLMs for \mbox{e-commerce} 
applications~\cite{shi2023llama, li2023ecomgpt, geng2022recommendation}. 
However, they fall short in the covered
\mbox{e-commerce} tasks and instruction data. 
%
%
We aim to bridge the gap and 
develop \mbox{e-commerce} foundation models with real-world utilities for a large variety of \mbox{e-commerce}
applications. 
%
%

To this end, we construct \dataset, an open-sourced, \mbox{large-scale}, and \mbox{high-quality} benchmark instruction dataset tailored for developing and evaluating LLMs in \mbox{e-commerce} realm.
\dataset covers 116,528 samples from 10 real and widely performed \mbox{e-commerce} tasks of 4 categories.  %
Each data sample comprises an instruction, an input, and an output. Some samples also incorporate a list of options.
All the 10 tasks have in-domain (IND) test samples, and 6 tasks also have OOD test samples which consist of products 
unseen in the training samples of the respective tasks.
\dataset undergoes rigorous and thorough scrutiny and is carefully crafted to enable a wide spectrum of 
empirical testing and exploration, including IND evaluation, OOD evaluation, and task-specific studies. 

Leveraging \dataset, we develop a series of \mbox{\underline{e}-\underline{C}ommerc\underline{e}} \underline{LLM}s, denoted as \method 
(pronounce: \mbox{e-sell \textprimstress em}, /\mbox{{\textsecstress\textipa{I-\textprimstress sel@m}/}})
models, by instruction-tuning 6 \mbox{general-purpose} LLMs, such as Llama 2~\cite{touvron2023llama}
and Mistral~\cite{jiang2023mistral}. 
%
%
The \method models are extensively evaluated 
across various settings, including IND and OOD data, unseen instructions, and different training 
sample sizes of \dataset. 
Overall, our experimental results demonstrate the following findings:

\textbf{(1)}
\method models substantially outperform baseline models, including the most advanced GPT-4~\cite{GPT4} and the state-of-the-art (SoTA) task-specific models, on almost all the 10 tasks in IND evaluation. 
On average, \method models show a substantial improvement of 10.7\%  over the best baseline models. 

\textbf{(2)} Moreover, \method exhibits excellent generalizability to OOD settings, including unseen products and unseen instructions, highlighting its superiority as a generalist \mbox{e-commerce} model.
Particularly, \method models establish an improvement of 9.3\% over the best baselines on the OOD 
(new) products.  
These results indicate the great potential of both the \dataset dataset and the \method models in 
empowering versatile and effective LLMs for \mbox{e-commerce}, and validate the potential of 
LLMs in doing \mbox{e-commerce} tasks. 

Figure~\ref{fig:overview} depicts the overall scheme of \method.
To the best of our knowledge,  this study is the first comprehensive and systematic study of 
instruction-tuning LLMs for \mbox{e-commerce} applications, open-sources the first  
large-scale, high-quality benchmark dataset (\dataset), and develops the state-of-the-art generalist LLMs (\method series) for \mbox{e-commerce}.
\dataset and \method models are publicly accessible through
\href{https://ninglab.github.io/eCeLLM/}{\textcolor{blue}{https://ninglab.github.io/eCeLLM/}}.

\section{Related Work}
\label{sec:related_work}

\textbf{Instruction Tuning with LLMs~~}
%
Instruction tuning enables the transfer of general knowledge captured in LLMs to specific application domains, facilitating the generalizability of LLMs.
FLAN~\cite{wei2021finetuned} and T0~\cite{sanh2021multitask} are the early work to investigate instruction tuning by evaluating the zero-shot performance of fine-tuned LLMs on numerous datasets. Both of the models show encouraging performance over the original \mbox{GPT-3}~\cite{brown2020language}, indicating the effectiveness of instruction tuning.
The importance of datasets, model scale, and instructions for instruction tuning are explored in a later work \mbox{FLAN-v2}~\cite{chung2022scaling}.
Considering the significant impact of datasets for instruction tuning, several studies collect sizable benchmarks consisting of numerous tasks, such as Super-NaturalInstructions~\cite{wang2022super}, \mbox{Self-instruct}~\cite{wang2022self} and Flan \mbox{Collection}~\cite{longpre2023flan}, for fine-tuning general-purpose LLMs on NLP tasks.
Meanwhile, instruction-tuned LLMs are employed in specific domains. For example, Platypus~\cite{lee2023platypus} is fine-tuned for reasoning in STEM, and Mammoth~\cite{yue2023mammoth} is for general math problem-solving. 
Our \method incorporates instruction tuning with LLMs for \mbox{e-commerce}. 

%
%

\textbf{LLMs for E-commerce~~} 
LLMs are emerging in the general e-commerce realm.
RecMind~\cite{wang2023recmind} utilizes LLMs as an autonomous recommender agent by enhancing their capability to comprehend user behaviors.
\mbox{LLaMA-E}~\cite{shi2023llama} is fine-tuned on LLaMA~\cite{touvron2023llama1} for various e-commerce authoring tasks, such as ads generation and general product QA.
EcomGPT~\cite{li2023ecomgpt} is a pioneer instruction-tuned LLM on its dataset EcomInstruct. 
However, most of its tasks are human-constructed by repurposing data from 
other tasks (e.g., from the ``query product matching" task,
a task is constructed to generate user queries based on the matched products),
and thus, could have limited utility in real applications.
%
A recent study~\cite{geng2022recommendation} develops a fine-tuned LLM, P5, with personalized prompts to conduct a few specific recommendation tasks (e.g., sequential recommendation).
These studies demonstrate the utility of LLMs and instruction tuning in \mbox{e-commerce} applications.
To the best of our knowledge, \method is the 
most comprehensive instruction-tuned \mbox{e-commerce} LLM.

\textbf{Comparison among Instruction-tuned LLMs for \mbox{E-commerce}~~}
%
%
Table~\ref{tbl:paper_comparison} summarizes the difference between our \method from LLaMA-E and EcomGPT,
the two existing fine-tuned LLMs capable of performing multiple \mbox{e-commerce} tasks
(P5 only performs recommendations).  
Fundamentally
different from \mbox{LLaMA-E} and EcomGPT, \method
focuses on comprehensive real-world e-commerce tasks and
data, 
includes very extensive evaluation and benchmarking, 
and 
open-sources
instruction data and models. 
 
\begin{table}[!t]
  \vspace{-10pt}
  \caption{Comparison among \mbox{E-commerce} LLMs} 
  \label{tbl:paper_comparison}
  \begin{footnotesize}
  \begin{threeparttable}
      \begin{tabular}{
	@{\hspace{0pt}}p{0.15\textwidth}@{\hspace{0pt}}
	@{\hspace{-4pt}}c@{\hspace{1pt}}
	@{\hspace{1pt}}c@{\hspace{1pt}}
	@{\hspace{8pt}}c@{\hspace{0pt}}
      }
      \toprule
       \multirow{2}{*}{Comparison} & LLaMA-E & EcomGPT &\method \\
       & ~\cite{shi2023llama} & ~\cite{li2023ecomgpt} & (ours)\\

	 \midrule
       Instruction tuning & \checkmark & \checkmark & \checkmark \\
       \mbox{Open-sourced data} & \ding{55} & \ding{55} & \checkmark \\
	\cmidrule(lr){1-4}
       Real-world tasks & \checkmark & \ding{55} & \checkmark \\
       ~~~~~\# training tasks & 5 & 122$^\ddag$ & 10 \\
       ~~~~~\mbox{\# In-domain tests} & 5 & 0 & 10 \\
       ~~~~~\mbox{\# out-of-domain tests }& {0} & {12}$^\ddag$ & {6} \\
       \cmidrule(lr){1-4}
       \# general-purpose & \multirow{2}{*}{5} 
       & \multirow{2}{*}{3} 
       & \multirow{2}{*}{5} \\
       LLMs evaluated & & & \\
       \# task-specific \mbox{SoTA models evaluated} & 
       \multirow{2}{*}{0} 
       & \multirow{2}{*}{0} 
       & \multirow{2}{*}{11} \\
       \mbox{\# base LLMs tuned} & 3 & 4 & 6 \\
       \cmidrule(lr){1-4}
       \mbox{Open-sourced models} & \ding{55} & \checkmark & \checkmark \\
      \bottomrule
      \end{tabular}
      \begin{tablenotes}[normal,flushleft]
      \begin{footnotesize}
      \item 
      $^\ddag$EcomGPT has 122 training tasks, most of which are manipulated from data
      of different other tasks.
      The training data is not publicly available.
      EcomGPT releases 12 test tasks (8 in Chinese) for only out-of-domain evaluation. 
      %
      
          \par
      \end{footnotesize}
      \end{tablenotes}
  \vspace{-15pt}
  \end{threeparttable}
  \end{footnotesize}
\end{table}

\section{\dataset Dataset}
\label{sec:dataset}

We introduce \dataset, an instruction dataset for adapting LLMs to e-commerce tasks.
\dataset features 3 key design principles.
\textbf{(1)} \textbf{Broad coverage}: \dataset includes 10 diverse tasks of 4 categories. 
%
Comprehensive and wide-ranging related tasks are critical in enabling versatile LLMs for a specific domain as shown in the literature~{\cite{lee2023beyond}}.
%
\textbf{(2)} \textbf{Realistic tasks}: 
\dataset focuses on real-world \mbox{e-commerce} tasks with real-world data, 
not human-manipulated tasks with synthetic data.
This ensures \mbox{e-commerce} LLMs tuned on \dataset a potentially high utility in real-world applications.
\textbf{(3)} \textbf{High quality}: \dataset considers only real-world data and undergoes rigorous scrutiny to ensure its accuracy and high quality. 
%
High data quality plays a pivotal role in building effective 
LLMs~\cite{hoffmann2022training,gadre2023datacomp}.
\vspace{-2pt}
\subsection{\mbox{E-commerce} Tasks}
\label{sec:dataset:preprocessing}

\dataset comprises 10 real-world tasks 
with \mbox{real-world} data that are widely performed in e-commerce applications.
These tasks fall within 4 categories: \textbf{(i)} product understanding, \textbf{(ii)} user understanding, \textbf{(iii)} query product matching, and \textbf{(iv)} product question answering. 
Such tasks are ubiquitous and essential on \mbox{e-commerce} platforms; success on these tasks would enable key functionalities in providing 
excellent user experience, promoting online retail, and driving sustainable revenues, among others. 
%


Particularly, 
\dataset includes 3 tasks for product understanding: 
\textbf{(1)} attribute value extraction (\AVE)~\cite{xu-etal-2019-scaling, AVEQA, yang2022mave}, \textbf{(2)} product matching (\EM)~\cite{kopcke2010evaluation,Rahm2010EM} and \textbf{(3)} product relation prediction (\IRP)~\cite{ahmed2021graph,xu2020product}.
For user understanding, 
\dataset includes \textbf{(4)} sentiment analysis (\SA)~\cite{wankhade2022survey} and \textbf{(5)} sequential recommendation (\SR)~\cite{petrov2023gsasrec,li2023text}.
\dataset also covers 3 query product matching tasks~\cite{reddy2022shopping}: \textbf{(6)} multi-class product classification (\MPC), \textbf{(7)} product substitute identification (\PSI), and \textbf{(8)} query-product ranking (\QPR).
For product question answering, \dataset contains the tasks of \textbf{(9)} answerability prediction (\AP)~\cite{gupta2019amazonqa} and \textbf{(10)} answer generation (\AG)~\cite{deng2023product}.
More details are available in Appendix~\ref{sec:appendix:preprocessing}.
Table~\ref{tbl:task_data} summarizes the tasks and their data sources. 
Fundamentally different from the instruction data of LLaMA-E~\cite{shi2023llama} and EcomGPT~\cite{li2023ecomgpt}, 
\dataset contains tasks all from real \mbox{e-commerce} platforms, and data all extracted 
from these real tasks. 

\vspace{-2pt}
\subsection{Diverse Instructions}
\label{sec:dataset:instruction}



%

Recent work~{\cite{xu2022multiinstruct}} shows that diverse instructions for LLM tuning improve LLM 
generalizability to new instructions. 
Inspired by this, we construct diverse instructions in \dataset.
For each task, we start with a clear and concise human-written seed instruction.
We then generate diverse instructions synonymous with the seed instruction via \mbox{GPT-4}~\cite{GPT4}, 
and select into \dataset 5 of those with writing styles (e.g., wording) distinct from the seed instruction.
Thus, \dataset involves 6 high-quality and diverse instructions, including the seed instruction, on each task. 
%

%
To test instruction understanding and model generalizability to new instructions, 
%
we hold out one instruction for each task as its ``unseen" instruction, 
which is inaccessible during model training. 
Thus, each task has 5 diverse instructions for instruction tuning, and 1
unseen instruction for testing. 
%
%
All the instructions are presented in Appendix~\ref{sec:appendix:instruction}.
%

\subsection{Data Quality Control}
\label{sec:dataset:quality}

%
In \dataset, we carry out the following procedures to ensure its accuracy and high quality.
Specifically, we 
\textbf{(1)} remove overlapping data between training and test sets to avoid data leakage;  
\textbf{(2)} retain only data in English to ensure the unity of languages in texts;  
\textbf{(3)} eliminate non-English notations such as HTML tags and Unicode;  
\textbf{(4)} only select products with detailed information to allow sufficient product knowledge that 
LLMs can learn from; 
%
\textbf{(5)} keep texts within a reasonable length following the convention 
in the literature~\cite{hou2022towards}; 
%
\textbf{(6)} manually inspect all processed data.
We also conduct task-specific quality control on individual tasks. 
All the quality control procedures are detailed in Appendix~\ref{sec:appendix:preprocessing}. 
%

\subsection{Training, Validation, and Test Sets}
\label{sec:dataset:split}

%

\dataset is split into training sets, validation sets, IND test sets, and OOD test sets as follows. 
Table~\ref{tbl:data_summary} summarizes the \dataset dataset.
Appendix~\ref{sec:appendix:preprocessing} describes the details of the data split process.

\textbf{Training Set~~}
%
\dataset has a training set of 10K for each individual task (except for \EM,
which has 2,022 training samples).
The training sets from all the tasks are combined into a dataset of 92,022 samples as the training set of \dataset. 

\textbf{Validation Set~~}
%
\dataset has a validation set of 1K for each individual task (except for \EM, which has 253 
validation samples).
Similarly, the validation sets of each task are combined into a dataset of 9,253 samples 
as the validation set of \dataset.

\textbf{In-Domain Test Set~~}
%
For each of the 10 tasks, \dataset also includes an in-domain (IND) test set of 1K samples (except for \EM, which has 253 IND samples).  
We define IND in terms of products that are within the same category as the training products.

\textbf{Out-of-Domain Test Set~~}
%
To evaluate the generalizability of LLMs tuned on \dataset to unseen samples -- a critical capacity desired 
to address the cold-start problem in \mbox{e-commerce},  we create OOD test sets in \dataset. 
We define OOD in terms of products, that is, new products unseen during LLM training 
are considered as OOD.
We determine OOD products using their category information. 
In \dataset, 6 tasks -- \AVE, \IRP, \SA, \SR, \AP, and \AG, have different product categories. 
For each of these tasks, 
all products from one particular category are held out as the OOD data for that task, 
as summarized in Table~\ref{tbl:data_summary}
(each product belongs to only one category). 
%
%
%
We focus on OOD (new) products, because
promoting and recommending new products to users are effective strategies in \mbox{e-commerce}
in driving sales and revenue, enhancing user experience, and increasing user engagement. 
Note we do not define OOD test sets over users
(i.e., hold out users into OOD test sets), because users are anonymous and user identifiers are absent in existing datasets. 
Even though, the unseen instructions could be used in proximity to new users -- a new user may have 
a completely new style of instructions to \method. 



\section{\method Models}
\label{sec:exp_protocol:base}

We build a series of \method models on top of 6 base models:
\textbf{(1)} \methodL: trained from large base models, 
Flan-T5 XXL ({11B parameters})~\cite{chung2022scaling} and Llama-2 13B-chat~\cite{touvron2023llama}, 
\textbf{(2)} \methodB: trained from medium-sized base models, 
Llama-2 7B-chat~\cite{touvron2023llama} and Mistral-7B Instruct-v0.2~\cite{jiang2023mistral}, and 
\textbf{(3)} \methodS: trained from small base models,
Flan-T5 XL (3B)~\cite{chung2022scaling} and Phi-2 (3B)~\cite{Phi2}. 
These \method models are instruction-tuned over \dataset training data. 

We fine-tune all the base models with LoRA~\cite{hu2021lora} and Huggingface transformers library~\cite{wolf2019huggingface}.
During fine-tuning, the learning rate and batch size of all the models are set as 1e-4 and 128, respectively.
We apply a cosine learning rate scheduler with a 5\% warm-up period for 3 epochs.
%
%
We set $\alpha$ and the rank in LoRA as 16, and add LoRA adaptors to all the projection layers and the language modeling head.
We perform 0-shot evaluations (i.e., without in-context examples)  on all the tasks. 
%

\section{Experimental Setup}
\label{sec:exp_protocol}


\begin{table*}[h]
\vspace{-10pt}
  \caption{Summary of Baseline Models}
  \centering
  \label{tbl:baseline}
  \footnotesize
  \begin{threeparttable}
      \begin{tabular}{
	@{\hspace{1pt}}c@{\hspace{4pt}}
	@{\hspace{4pt}}p{0.18\textwidth}@{\hspace{0pt}}
	@{\hspace{0pt}}p{0.80\textwidth}@{\hspace{1pt}}
    }
    \toprule
	\multicolumn{2}{l}{\multirow{2}{*}{General-purpose LLMs}} & 
	\textbf{(1)} GPT-4 Turbo~\cite{GPT4}, 
	\textbf{(2)} Gemini Pro~\cite{team2023gemini}, 
	\textbf{(3)} Claude 2.1~\cite{Claude2}, 
	\textbf{(4)} Llama-2 13B-chat~\cite{touvron2023llama},
	\textbf{(5)} Mistral-7B Instruct-v0.2~\cite{jiang2023mistral}\\
	\cmidrule(lr){1-3}
	\multicolumn{2}{l}{\parbox{0.15\textwidth}{E-commerce LLM}} & 	
	 \textbf{(1)} EcomGPT~\cite{li2023ecomgpt}: To the best of our knowledge, it is the only open-source \mbox{e-commerce} LLM.\\
	\cmidrule(lr){1-3}
	\multicolumn{3}{l}{\mbox{SoTA Task-specific Models}}  \\	
	\cmidrule(lr){3-3}
	&
	\multirow{2}{*}{\AVE (Section~\ref{sec:appendix:preprocessing:ave})} & 
	\textbf{(1)} SUOpenTag~\cite{xu-etal-2019-scaling}, 
	\textbf{(2)}AVEQA~\cite{AVEQA}: Both methods scan the product information, token by token, to extract the values of the specified attributes.\\
	\cmidrule(lr){3-3}
	& 
	\multirow{2}{*}{\IRP (Section~\ref{sec:appendix:preprocessing:irp})} & 
	\textbf{(1)} RGCN~\cite{schlichtkrull2018modeling}: It utilizes a graph convolutional network to capture relations between products.
    \textbf{(2)} DeBERTaV3~\cite{he2021debertav3}: It predicts product relations from product titles.\\
	\cmidrule(lr){3-3} 
	& \multirow{3}{*}{\SA (Section~\ref{sec:appendix:preprocessing:sa})} & 
	\textbf{(1)} BERTweet~\cite{nguyen2020bertweet}: It is a pre-trained language model for English tweets, and achieves superior performance over RoBERTa~\cite{liu2019roberta} and XLM-R~\cite{conneau2019unsupervised} on \SA. 
	\textbf{(2)} P5~\cite{geng2022recommendation}: It is a pre-trained LLM for \SA from \mbox{e-commerce} user reviews. \\
	\cmidrule(lr){3-3}
	& 
	\multirow{3}{*}{\SR (Section~\ref{sec:appendix:preprocessing:sr})}& 
	\textbf{(1)} gSASRec~\cite{petrov2023gsasrec}, \textbf{(2)} Recformer~\cite{li2023text}:
    Both methods leverage Transformer~\cite{vaswani2017attention} to predict the next product of users' interest based on users' historical activities on products.\\
    \cmidrule(lr){3-3}
    	& 
	\multirow{3}{*}{\AG (Section~\ref{sec:appendix:preprocessing:ag})} & 
	\textbf{(1)} GPT-4 Turbo~\cite{GPT4}: To the best of our knowledge, there are no
    models specifically designed for \AG. Thus, GPT-4 Turbo is used as the SoTA task-specific model in this task 
    as it has outstanding performance in general question answering~\cite{GPT4}.\\
    \cmidrule(lr){3-3}
    & 
   \mbox{\EM, \MPC, \PSI,  \QPR, \AP} 
   (Section~\ref{sec:appendix:preprocessing:em}, ~\ref{sec:appendix:preprocessing:mpc}, 
    ~\ref{sec:appendix:preprocessing:psi}, ~\ref{sec:appendix:preprocessing:qpr}, 
    ~\ref{sec:appendix:preprocessing:ap}).
    &
	{\textbf{(1)} BERT~\cite{devlin2018bert}, \textbf{(2)} DeBERTaV3~\cite{he2021debertav3}: Both methods generate predictions \newline from the textual product information (e.g., titles).} \\
    %
	%
    \bottomrule
    \end{tabular}
  \end{threeparttable}
  \vspace{-10pt}
\end{table*}


%
%
We compare \method against 3 categories of baseline models: \textbf{(1)} general-purpose LLMs, 
\textbf{(2)} \mbox{e-commerce} LLMs, and \textbf{(3)} SoTA task-specific models. 
Table~\ref{tbl:baseline} lists all the baseline models. 
%
%
Note that for the task-specific models, we only use the best (i.e., SoTA) models for each individual task based on 
literature, so as to enable schematic comparison between generalist modeling from \method and the 
task-specific modeling. 
We conduct IND and OOD tests on respective test datasets (Section~\ref{sec:dataset:split}) for all the models. 
%
Table~\ref{tbl:data_summary} lists the evaluations conducted for different tasks. 
Details on using/training baseline models are in Appendix~\ref{sec:appendix:training}.  
The prompt templates used in the evaluations are in Appendix~\ref{sec:appendix:Template}.

%
\textbf{General-purpose LLMs~~} 
%
For all the general-purpose LLMs, 
we use the checkpoints released by their developers.
For GPT-4 Turbo, Gemini Pro, and Claude 2.1, we access the checkpoints using their official APIs.
For \mbox{Llama-2 13B-chat} and \mbox{Mistral-7B Instruct-v0.2}, we use the checkpoints released in Huggingface~\cite{wolf2019huggingface}.

We perform 1-shot evaluations on all the general-purpose LLMs.
Existing work~\cite{li2023ecomgpt} tests LLMs on \mbox{e-commerce} tasks 
in a 0-shot setting.
Meanwhile, in-context examples can notably benefit LLMs~\cite{brown2020language}.
Thus, in our case, we conduct 1-shot evaluations, balancing the computing cost and model performance, 
to enable stronger performance from the \mbox{general-purpose} LLMs.
Although extensive prompt engineering and many in-context examples
could enable better performance from LLMs,
it is less practical in real \mbox{e-commerce} applications; for example, asking users to provide many 
in-context examples of the \mbox{e-commerce} platform, if ever feasible, can reduce user engagement. 
Also, it is not cost- and energy-efficient to do large-scale, few-shot in-context learning on LLMs over large test sets like \dataset's. 
%

\textbf{\mbox{E-commerce} LLMs~~} 
%
For EcomGPT, we use the checkpoint released by its authors.
We conduct both 0-shot and 1-shot evaluations on EcomGPT, as we empirically observe
that 1-shot may result in better performance than 0-shot for EcomGPT 
(EcomGPT originally conducts 0-shot evaluations).  
Therefore, we report the best performance of the two evaluations on each task.
%

%

\textbf{SoTA Task-specific Models} 
%
We train the SoTA task-specific models \mbox{SUOpenTag}, \mbox{AVEQA}, \mbox{RGCN}, and \mbox{gSASRec}
from scratch using \dataset training data of individual tasks and the model implementations
published by their respective authors. 
%
%
For P5, we use the checkpoint released by its authors (it has already been pre-trained on e-commerce data).
For GPT-4 Turbo, we access the checkpoint via its API.
For \mbox{Recformer}, \mbox{BERT}, \mbox{BERTweet}, and \mbox{DeBERTaV3}, we tune the checkpoints on specific tasks. 

\begin{table*}[!h]
\vspace{-10pt}
  \caption{Overall Performance in IND Evaluation}
  \centering
  \vspace{2pt}
  \label{tbl:overall_indomain}
  \footnotesize
  \begin{threeparttable}
      \begin{tabular}{
        @{\hspace{6pt}}l@{\hspace{3pt}}  
	@{\hspace{6pt}}c@{\hspace{5pt}} 
 	@{\hspace{5pt}}c@{\hspace{6pt}} 
	@{\hspace{4pt}}c@{\hspace{6pt}} 
	@{\hspace{4pt}}c@{\hspace{4pt}} 
	@{\hspace{3pt}}c@{\hspace{7pt}} 
        @{\hspace{0pt}}c@{\hspace{6pt}}  
        @{\hspace{4pt}}c@{\hspace{7pt}}  
        @{\hspace{8pt}}c@{\hspace{7pt}}  
        @{\hspace{8pt}}c@{\hspace{9pt}}  
        @{\hspace{8pt}}c@{\hspace{6pt}}  
      }
      \toprule
      \multirow{2.5}{*}{Model} 
      & \AVE~~ & \IRP~~ & \EM~~ & \SA~~ & \SR~~ & \MPC & \PSI & \QPR & \AP & \AG \\ 
      \cmidrule(lr){2-11}
      & F1* & Macro F1 & F1 & Macro F1 & HR@1 & Accuracy & F1 & NDCG & F1 & F$_{\text{BERT}}$ \\ 
      \midrule
      GPT-4 Turbo & 0.495 & 0.326 & 0.753 & 0.516 & \underline{0.387} & 0.611 & 0.195 & \underline{0.875} & 0.649 & \underline{\textbf{0.858}} \\ 
      Gemini Pro & 0.396 & 0.136 & 0.867 & 0.470 & 0.269 & 0.584 & 0.248 & 0.821 & 0.506 & 0.855 \\ 
      Claude 2.1 & 0.381 & 0.275 & 0.523 & 0.415 & 0.066 & 0.655 & 0.273 & 0.821 & 0.280 & 0.841 \\ 
      Llama-2 13B-chat & 0.002 & 0.333 & 0.434 & 0.188 & 0.056 & 0.504 & 0.252 & 0.815 & 0.623 & 0.811 \\ 
      Mistral-7B Instruct-v0.2 & 0.369 & 0.324 & 0.613 & 0.470 & 0.164 & 0.529 & 0.305 & 0.842 & 0.588 & 0.853 \\ 
      \cmidrule(lr){2-11}
      EcomGPT & 0.000 & 0.091 & 0.648 & 0.188 & 0.042 & 0.540 & 0.170 & 0.000 & 0.086 & 0.669 \\ 
      \cmidrule(lr){2-11}
      SoTA task-specific model & \underline{0.546} & \underline{0.588} & \underline{\textbf{0.995}} & \underline{0.573} & 0.265 & \underline{\textbf{0.703}} & \underline{0.389} & 0.859 & \underline{0.830} & \underline{\textbf{0.858}} \\ 
	\midrule
      \methodL & 0.582 & \textbf{0.611} & \textbf{0.995} & \textbf{0.648} & 0.526 & 0.684 & \textbf{0.501} & 0.870 & \textbf{0.851} & 0.841 \\ 
      \methodB & \textbf{0.662} & 0.558 & \textbf{0.995} & 0.639 & \textbf{0.542} & 0.696 & 0.305 & \textbf{0.876} & 0.846 & {0.842} \\ 
      \methodS & 0.509 & 0.518 & 0.991 & 0.596 & 0.479 & 0.650 & 0.392 & 0.870 & 0.846 & {0.842} \\ 
	\midrule
      improvement (\%, avg: 10.7) &  21.2 &  3.9 &  0.0 &  13.1 &  40.1 &  -1.0 &  28.8 &  0.1 &  2.5 &  -1.9 \\ 
      \bottomrule
      \end{tabular}
      \begin{tablenotes}[normal,flushleft]
      \begin{footnotesize}
      \item
      In this table, ``F1*", ``Macro F1", ``F1", ``HR@1", ``Accuracy", ``NDCG" and ``F$_{\text{BERT}}$" are the primary evaluation metrics in respective tasks (Appendix~\ref{sec:appendix:preprocessing}).
      For each task, 
      the best baseline performance is \underline{underlined}, and the overall best performance is in \textbf{bold}.
      The row ``improvement” presents the percentage improvement of the best-performing \method model over the best-performing baseline model (\underline{underlined}) in each task.
      We also include the average (`avg') improvement across all the tasks in the table.

      \par
      \end{footnotesize}
      \end{tablenotes}
  \end{threeparttable}
  \vspace{-10pt}
\end{table*}


\section{Experimental Results}
\label{sec:exp_results}

%

We evaluate the models using the test sets of individual tasks.
For each task, we employ multiple metrics for a comprehensive evaluation (Appendix~\ref{sec:appendix:preprocessing}).
For the sake of clarity, 
in this section, 
we present the performance only in terms of the primary evaluation metric for each task (Table~\ref{tbl:task_data}).
%
%
%
As discussed in Appendix~\ref{sec:appendix:base_model}, 
Llama-2 13B-chat, Mistral-7B Instruct-v0.2, and Phi-2 are the best base models for \methodL, \methodB, and \methodS, respectively.
Thus, by default, \methodL, \methodB, and \methodS refer to those tuned from these base models, respectively. 

\textbf{Main Results~~}
%
Our comprehensive experiments yield the following main results: 
\textbf{(1)} \method models demonstrate the best performance on almost all the IND tasks, with a significant average improvement of 10.7\%
over the \mbox{general-purpose} LLMs, \mbox{e-commerce} LLMs, and the SoTA \mbox{task-specific} models 
across the 10 tasks
(Section~\ref{sec:exp_results:in_domain}).
\textbf{(2)} \method models show outstanding generalizability to OOD products 
and surpass the best baselines with a remarkable average improvement of 9.3\% in OOD evaluation
 (Section~\ref{sec:exp_results:out_domain}).
\textbf{(3)} By training over diverse instructions, 
\method is equipped with strong generalizability to unseen instructions (Section~\ref{sec:exp_results:instruction}).
%
%
\textbf{(4)} Trained on all the tasks in \dataset together,
\method exhibits similar or better performance than models trained on each individual task (Section~\ref{sec:exp_results:joint}).
%
\textbf{(5)} \method models benefit from larger instruction training data for \mbox{e-commerce} tasks
(Section~\ref{sec:exp_results:data}).

\vspace{-2pt}
\subsection{In-domain Evaluation}
\label{sec:exp_results:in_domain}

%

Table~\ref{tbl:overall_indomain}
shows the performance of \method and all the baseline methods in IND evaluation, 
where \method models are trained using \dataset training set (i.e., including all the tasks), 
and the SoTA task-specific models are trained using the task-specific training data.  
%
%
%
Among the SoTA task-specific models, we report the results of only the \mbox{best-performing} 
model on each task. 
%
%
The complete results of each task are presented in Appendix~\ref{sec:appendix:performance}.
%
%
%


\textbf{Overall Comparison~~}
\label{sec:exp_results:in_domain:overall}
%
%
As shown in Table~\ref{tbl:overall_indomain}, 
overall, \method substantially outperforms baseline models across the 10 \mbox{e-commerce} tasks
at 10.7\% on average.
Particularly, \method models (i.e., \methodL, \methodB, and \methodS) achieve superior performance over 
the baselines on 7 out of the 10 tasks with an average improvement of 15.7\%.
On the rest 3 tasks \EM, \MPC, and \AG, \method models achieve the same or comparable performance as the baselines (e.g., maximum difference of 1.9\% as on \AG).
These results demonstrate the remarkable effectiveness of \method compared with 
the general-purpose LLMs, the SoTA task-specific models, and the existing e-commerce LLM across the
\mbox{e-commerce} tasks.

\textbf{Comparison between \method and General-purpose LLMs~~}
\label{sec:exp_results:in_domain:llm}
%
Table~\ref{tbl:overall_indomain} shows that \method models substantially outperform the 
general-purpose LLMs by a remarkable margin. 
For example, across the 10 tasks, 
\methodL achieves a significant average improvement of 39.6\% over \mbox{GPT-4 Turbo}. 
A key difference between \method and GPT-4 Turbo is that  \method is specifically tuned on our instruction dataset \dataset for \mbox{e-commerce}. 
The remarkable improvement of {\method} over general-purpose LLMs suggests that there could be a significant gap between general knowledge and the knowledge required for \mbox{e-commerce} tasks, highlighting the significance of \dataset in imparting knowledge pertinent to \mbox{e-commerce} into LLMs.
%

%
We also observe that 
general-purpose LLMs lag behind the SoTA task-specific models by a large margin.
For example, the best general-purpose LLM GPT-4 Turbo considerably underperforms the SoTA task-specific models on 7 out of the 10 tasks (e.g., 0.495 vs 0.546 on \AVE).
This underscores the critical need to deliberately accommodate general-purpose LLMs for \mbox{e-commerce} tasks.

\textbf{Comparison between \method and \mbox{E-commerce} LLMs~~}
\label{sec:exp_results:in_domain:ecomgpt}
%
According to Table~\ref{tbl:overall_indomain}, \method models 
demonstrate substantial improvement over the existing \mbox{e-commerce} LLM EcomGPT on all the tasks.
For example, \methodL surpasses EcomGPT remarkably by  244.6\% on \SA, and \EM by 53.5\%. 
%
%
%
Both \method and EcomGPT are \mbox{instruction-tuned} LLMs for \mbox{e-commerce}. 
However, our instruction dataset \dataset for \method fundamentally differs from EcomGPT's EcomInstruct dataset:
\dataset includes only real-world data and real-world \mbox{e-commerce} tasks, while EcomInstruct
incorporates a considerable amount of synthetic data and tasks, which hinders its applicability in real \mbox{e-commerce} tasks. 
%
%
%
%

\textbf{Comparison between \method and SoTA task-specific models~~}
\label{sec:exp_results:in_domain:expert}
%
%
Table~\ref{tbl:overall_indomain} shows the SoTA task-specific models perform best among the baselines
on each respective task. 
%
%
%
%
The substantial improvement of \method over the SoTA \mbox{task-specific} models serves as strong evidence that \method, with specific tuning over \dataset, could leverage knowledge shared across multiple \mbox{e-commerce} tasks and 
boost performance on each individual task. 
\vspace{-1pt}
\subsection{Out-of-domain Evaluation}
\label{sec:exp_results:out_domain}

\begin{table}[!h]
\vspace{-12pt}
  \caption{Overall Performance in OOD Evaluation}
  \centering
  \vspace{2pt}
  \label{tbl:overall_ood}
  \footnotesize
  \begin{threeparttable}
      \begin{tabular}{
        %
        @{\hspace{2pt}}l@{\hspace{2pt}}
	@{\hspace{2pt}}c@{\hspace{3pt}}
	@{\hspace{3pt}}c@{\hspace{3pt}}
	@{\hspace{3pt}}c@{\hspace{2pt}}
	@{\hspace{2pt}}c@{\hspace{2pt}}
	@{\hspace{2pt}}c@{\hspace{2pt}}
        @{\hspace{4pt}}c@{\hspace{2pt}}
      }
      \toprule
      \multirow{2.5}{*}{Model} & \AVE & \IRP & \SA & \SR & \AP & \AG \\ 
      \cmidrule(lr){2-7}
      & F1* & M-F1 & M-F1 & HR@1 & F1 & F$_{\text{BERT}}$ \\ 
      \midrule
      GPT-4 Turbo & 0.397 & 0.392 & 0.510 & \underline{0.198} & 0.680 & \underline{\textbf{0.860}} \\ 
      Gemini Pro & 0.275 & 0.123 & 0.454 & 0.116 & 0.552 & 0.856 \\ 
      Claude 2.1 & \underline{\textbf{0.410}} & 0.277 & 0.369 & 0.036 & 0.245 & 0.842 \\ 
      Llama-2 13B-chat & 0.000 & 0.324 & 0.178 & 0.050 & 0.644 & 0.808 \\ 
      Mistral-7B & \multirow{2}{*}{0.264} & \multirow{2}{*}{0.327} & \multirow{2}{*}{0.438} & \multirow{2}{*}{0.108} & \multirow{2}{*}{0.608} & \multirow{2}{*}{0.851} \\
      Instruct-v0.2 & & & & & & \\
      \cmidrule(lr){2-7}
      EcomGPT & 0.001 & 0.096 & 0.178 & 0.023 & 0.140 & 0.722 \\ 
      \cmidrule(lr){2-7}
      SoTA  & \multirow{2}{*}{0.269} & \multirow{2}{*}{\underline{0.507}} & \multirow{2}{*}{\underline{0.567}} 
      & \multirow{2}{*}{0.081} & \multirow{2}{*}{\underline{0.853}} & \multirow{2}{*}{\underline{\textbf{0.860}}} \\
      task-specific model & & & & & &\\
      \midrule
      \methodL & 0.335 & \textbf{0.558} & 0.629 & 0.273 & 0.867 & {0.841} \\ 
      \methodB & {0.367} & 0.502 & \textbf{0.640} & \textbf{0.280} & 0.878 & 0.840 \\ 
      \methodS & 0.302 & 0.520 & 0.565 & 0.241 & \textbf{0.879} & 0.840 \\ 
      \midrule
      improvement  & \multirow{2}{*}{-10.5} & \multirow{2}{*}{10.1} & \multirow{2}{*}{14.1} & \multirow{2}{*}{41.4} & \multirow{2}{*}{3.0} & \multirow{2}{*}{-2.2} \\
      (\%, avg: 9.3) & & & & & &\\
      \bottomrule
      \end{tabular}
      \begin{tablenotes}[normal,flushleft]
      \begin{footnotesize}
      \item
      In this table, ``M-F1" represents macro F1. The columns in this table have the same meanings as those in Table~\ref{tbl:overall_indomain}. 
      \par
      \end{footnotesize}
      \end{tablenotes}
  \end{threeparttable}
\vspace{-8pt}  
\end{table}


Table~\ref{tbl:overall_ood} shows the performance of \method and baselines in OOD evaluation.  
%
%
%
Overall, we observe a similar trend to that in IND evaluation.
Specifically, \methodL outperforms the SoTA task-specific models on OOD evaluation by a wide margin on 5 out of the 6 tasks, 
except for \AG task, on which \method is comparable with the SoTA task-specific model (e.g., 2.2\% difference).
%
Similarly, \methodL outperforms the general-purpose LLMs on 4 out of the 6 tasks (i.e., \IRP, \SA, \SR, and \AP).
Across the 6 tasks, \methodL demonstrates a substantial average improvement of 18.9\% over GPT-4 Turbo.
Compared to EcomGPT, \methodL again shows terrific advantages on all the tasks in OOD evaluation.
As shown in the literature~\cite{LIKA20142065}, 
cold start for new products has been an unsolved issue but a key driver in 
\mbox{e-commerce} applications. 
The OOD generalizability of \method to new products 
as demonstrated in Table~\ref{tbl:overall_ood} makes \method a highly viable tool 
for \mbox{e-commerce} applications. 

Note that GPT-4 and Claude 2.1 exhibit robust performance on \AVE and \AG, 
indicating that these general-purpose LLMs are inherent with rich knowledge for solving extraction and question-answering problems.
%
However,  
they may still lack comprehensive knowledge to effectively perform multiple, diverse \mbox{e-commerce} tasks, such as on the other 4 tasks. 




%
%
%

\subsection{\mbox{\method Generalizability to Unseen Instructions}}
\label{sec:exp_results:instruction}

\begin{table*}[!h]
  \vspace{-13pt}
  \caption{Performance on Unseen Instructions in IND Evaluation}
  \vspace{2pt}
  \centering
  \label{tbl:instruction_indomain}
  \footnotesize
  \begin{threeparttable}
      \begin{tabular}{
        @{\hspace{4pt}}l@{\hspace{4pt}}
        @{\hspace{4pt}}c@{\hspace{4pt}}
	  @{\hspace{4pt}}c@{\hspace{4pt}}
	  @{\hspace{4pt}}c@{\hspace{5pt}}
	  @{\hspace{5pt}}c@{\hspace{4pt}}
	  @{\hspace{4pt}}c@{\hspace{4pt}}
	  @{\hspace{4pt}}c@{\hspace{4pt}}
        @{\hspace{5pt}}c@{\hspace{4pt}}
        @{\hspace{1pt}}c@{\hspace{1pt}}
        @{\hspace{6pt}}c@{\hspace{4pt}}
        @{\hspace{8pt}}c@{\hspace{8pt}}
        @{\hspace{8pt}}c@{\hspace{4pt}}
      }
      \toprule
      \multirow{2.5}{*}{Model}  
      & \multirow{2.5}{0.1\textwidth}{\centering{Training Instructions}}
      & \AVE & \IRP & \EM & \SA & \SR & \MPC & \PSI & \QPR & \AP & \AG \\ 
      \cmidrule(lr){3-12}
      && F1* & Macro F1 & F1 & Macro F1 & HR@1 & Accuracy & F1 & NDCG & F1 & F$_{\text{BERT}}$ \\ 
      \midrule
      \multirow{2}{*}{\methodL} 
      & single 
      & 0.046 & 0.619 & 0.995 & 0.610 & 0.526 & 0.696 & 0.206 & 0.870 & 0.846 & 0.841 \\ 
       & diverse
      & \textbf{0.553} & \textbf{0.638} & 0.995 & \textbf{0.639} & 0.524 & 0.694 & \textbf{0.335} & 0.870 & 0.842 & 0.841 \\ 
      \midrule
      \multirow{2}{*}{\methodB} 
      & single
      & 0.000 & \textbf{0.618} & 0.995 & 0.554 & 0.543 & 0.696 & 0.241 & 0.878 & \textbf{0.852} & 0.850 \\ 
      & diverse
      & \textbf{0.622} & 0.540 & 0.995 & \textbf{0.643} & 0.540 & 0.695 & \textbf{0.253} & 0.878 & 0.822 & 0.844 \\ 
      \midrule
      \multirow{2}{*}{\methodS} 
      & single
      & 0.447 & 0.535 & 0.991 & 0.577 & \textbf{0.478} & 0.652 & 0.314 & 0.867 & 0.841 & 0.838 \\ 
       & diverse
      & \textbf{0.488} & \textbf{0.552} & 0.991 & 0.577 & 0.457 & \textbf{0.660} & \textbf{0.381} & 0.871 & 0.845 & 0.842 \\ 
      \bottomrule
      \end{tabular}
      \begin{tablenotes}[normal,flushleft]
      \begin{footnotesize}
      \item
      In this table, ``single" and ``diverse" indicate that the \method models are tuned over single and diverse instructions, respectively.
      The best performance of each \method model (e.g., \methodL, \methodB abd \methodS) is in \textbf{bold}, if the performance difference between the \method model tuned over single and 
      diverse instructions exceeds 1\%.
      %
      %
      %
      %
      \par
      \end{footnotesize}
      \end{tablenotes}
  \end{threeparttable}
  \vspace{-10pt}
\end{table*}

   
\begin{table*}[!h]
  \vspace{-5pt}
  \caption{Performance of Generalist and Task-specific {\method} Models in IND Evaluation}
  \centering
  \footnotesize
  \label{tbl:model_indomain}
  \begin{threeparttable}
      \begin{tabular}{
        @{\hspace{4pt}}l@{\hspace{4pt}}
        @{\hspace{4pt}}c@{\hspace{4pt}}
	  @{\hspace{4pt}}c@{\hspace{4pt}}
	  @{\hspace{4pt}}c@{\hspace{5pt}}
	  @{\hspace{5pt}}c@{\hspace{4pt}}
	  @{\hspace{4pt}}c@{\hspace{4pt}}
	  @{\hspace{4pt}}c@{\hspace{4pt}}
        @{\hspace{5pt}}c@{\hspace{4pt}}
        @{\hspace{1pt}}c@{\hspace{1pt}}
        @{\hspace{6pt}}c@{\hspace{4pt}}
        @{\hspace{8pt}}c@{\hspace{8pt}}
        @{\hspace{8pt}}c@{\hspace{4pt}}
      }
      \toprule
      \multirow{2.5}{*}{Model} & \multirow{2.5}{*}{Training Tasks}
      & \AVE & \IRP & \EM & \SA & \SR & \MPC & \PSI & \QPR & \AP & \AG \\ 
      \cmidrule(lr){3-12}
      && F1* & Macro F1 & F1 & Macro F1 & HR@1 & Accuracy & F1 & NDCG & F1 & F$_{\text{BERT}}$ \\ 
      \midrule
      \multirow{2}{*}{\methodL} & Task-specific & \textbf{0.599} & 0.521 & 0.995 & 0.616 & 0.518 & 0.655 & 0.000 & \textbf{0.879} & 0.854 & 0.841 \\ 
      & Generalist & 0.582 & \textbf{0.611} & 0.995 & \textbf{0.648} & \textbf{0.526} & \textbf{0.684} & \textbf{0.501} & 0.870 & 0.851 & 0.841 \\ 
      \midrule
      \multirow{2}{*}{\methodB} & Task-specific & \textbf{0.757} & 0.543 & 0.987 & \textbf{0.655} & 0.535 & 0.681 & 0.000 & 0.883 & \textbf{0.864} & 0.841\\
      & Generalist & 0.662 & \textbf{0.558} & 0.995 & 0.639 & \textbf{0.542} & \textbf{0.696} & \textbf{0.305} & 0.876 & 0.846 & 0.842\\
      \midrule
      \multirow{2}{*}{\methodS} &
      Task-specific & 0.397 & 0.348 & 0.991 & \textbf{0.608} & 0.413 &  0.646 & 0.000 & 0.858 & 0.835 & 0.835\\
      & Generalist & \textbf{0.509} & \textbf{0.518} & 0.991 & 0.596 & \textbf{0.479} & 0.650 & \textbf{0.392} & \textbf{0.870} & \textbf{0.846} & 0.842\\
      \bottomrule
      \end{tabular}
      \begin{tablenotes}[normal, flushleft]
      \begin{footnotesize}
      \item
      In this table, ``Task-specific" indicates that the \method models are tuned on individual tasks;
      ``Generalist" represents tuning \method models using all tasks together.
      The best performance of generalist and task-specific \method models on each task is in \textbf{bold}, if the performance difference between the 
      generalist and task-specific \method model exceeds 1\%.      
      \par
      \end{footnotesize}
      \end{tablenotes}
  \end{threeparttable}
  \vspace{-10pt}  
\end{table*}

     
%
To evaluate the generalizability of \method to unseen instructions, 
we compare \method models tuned from diverse instructions 
and tuned from single instructions per task, and test their performance over unseen instructions. 
%
%
%
%

%
As shown in Table~\ref{tbl:instruction_indomain}, overall, 
\method models tuned over multiple, diverse instructions 
exceed or resemble those tuned over single instructions. 
For example, 
\methodL performs much better on unseen instructions on \AVE and \PSI when trained from 
diverse instructions. 
On the other 8 tasks, \methodL trained from diverse instructions 
is either better than or comparable to that from single instructions.
Similarly, in the case of \methodB, using diverse instructions shows very strong effects on performance 
on \AVE and \SA, while for other tasks, performance remains comparable (except for \IRP). 
For \methodS, diverse instructions enhance performance on almost all the tasks. 

These results illustrate the pivotal role of diverse instructions for \mbox{e-commerce} tasks, and 
underscore the utility of \dataset in generalizing LLMs to new instructions in \mbox{e-commerce} applications. 
The results also showcase the generalizability of \method to unseen instructions. 
As new users may apply instructions unseen in existing \mbox{e-commerce} data, such generalizability 
also indicates the potential of \method for cold-start users in \mbox{e-commerce}. 
It is noticeable that \method is able to generalize to unseen instructions (Table~\ref{tbl:instruction_indomain}) 
with a similar performance as it has on ``seen" instructions (Table~\ref{tbl:overall_indomain}). 
For example, on \SA, \methodL shows only a 1.4\% difference in its performance 
in the former setting and in the latter setting (0.639 vs 0.648). 
This further demonstrates the strong generalizability of \method.

 
%
%


\subsection{Generalist vs Task-specific \method Models}
\label{sec:exp_results:joint}
%

We compare the {\method} models tuned using all tasks (generalist models) with those tuned on individual tasks (task-specific models) and present the performance comparison in IND evaluation 
in Table~\ref{tbl:model_indomain}.
%
%
%
The results of the OOD evaluation are presented in Appendix~\ref{sec:appendix:ood:joint}.  
As shown in Table~\ref{tbl:model_indomain},  
generalist {\method} demonstrates slightly stronger 
performance compared to \mbox{task-specific} {\method} on each task
(e.g., 2.7\% average improvement of generalist {\methodL} over all tasks except {\PSI}).
%
For example, on \AVE, generalist \methodL is comparable to the \mbox{task-specific} {\methodL} (0.582 vs 0.599).
%
On \IRP, generalist \methodL demonstrates a substantial improvement of 17.3\% compared to \mbox{task-specific} {\methodL} (0.611 vs 0.521).
%
%
As evidenced by the results, 
by training on all tasks together, 
{\method} models enjoy strong versatility and 
could enable knowledge transfer across tasks for improved performance.

%



\subsection{Impact of Training Data Size on \method}
\label{sec:exp_results:data}

\begin{figure}[h]
	\vspace{-2pt}
	\centering
	\footnotesize
	\begin{minipage}{0.48\linewidth}
		\centering
		\includegraphics[width=\linewidth]{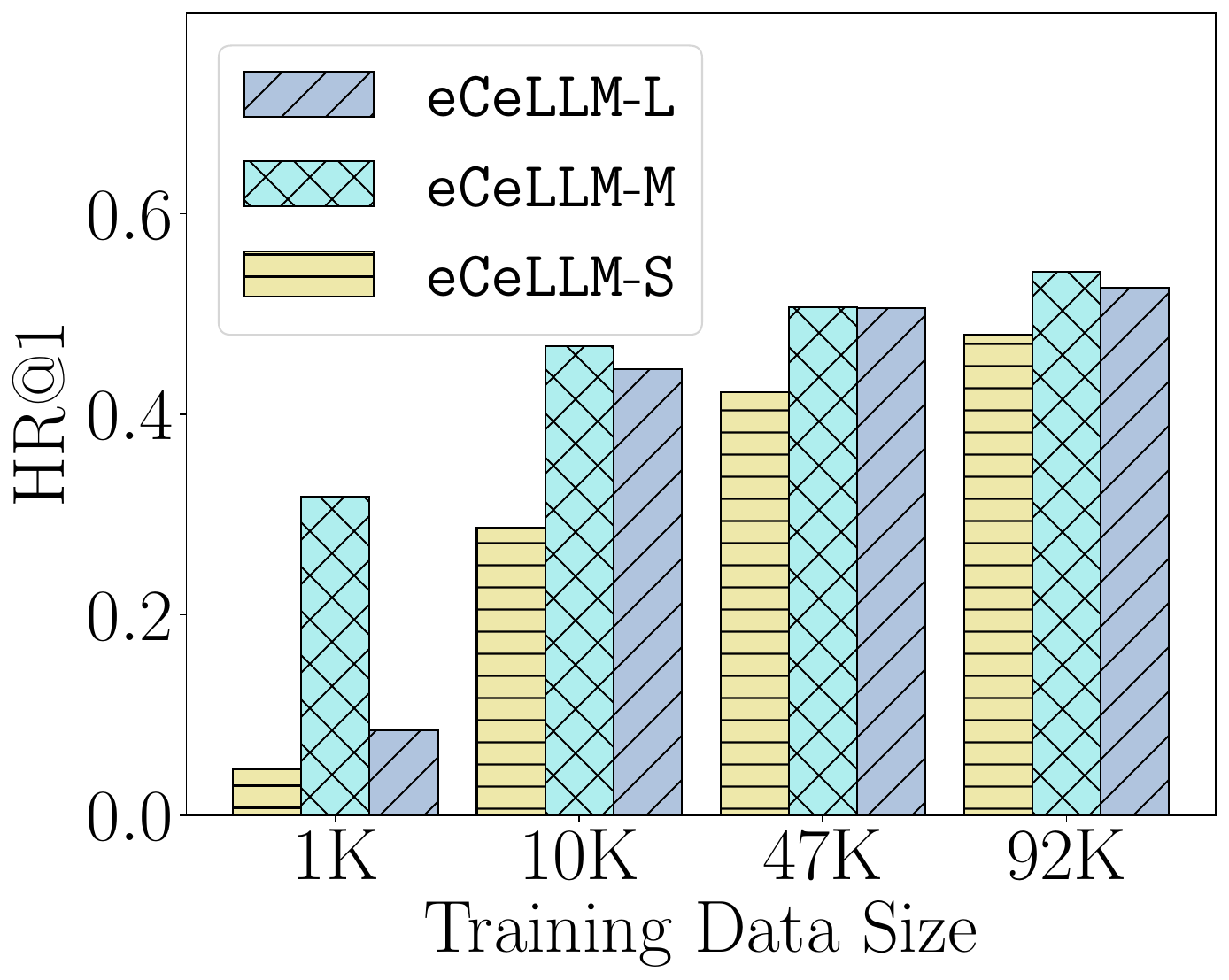}
		\vspace*{-12pt}
		\subcaption{IND Evaluation}
		\label{fig:indomain}
	\end{minipage}
	\begin{minipage}{0.48\linewidth}
		\centering
		\includegraphics[width=\linewidth]{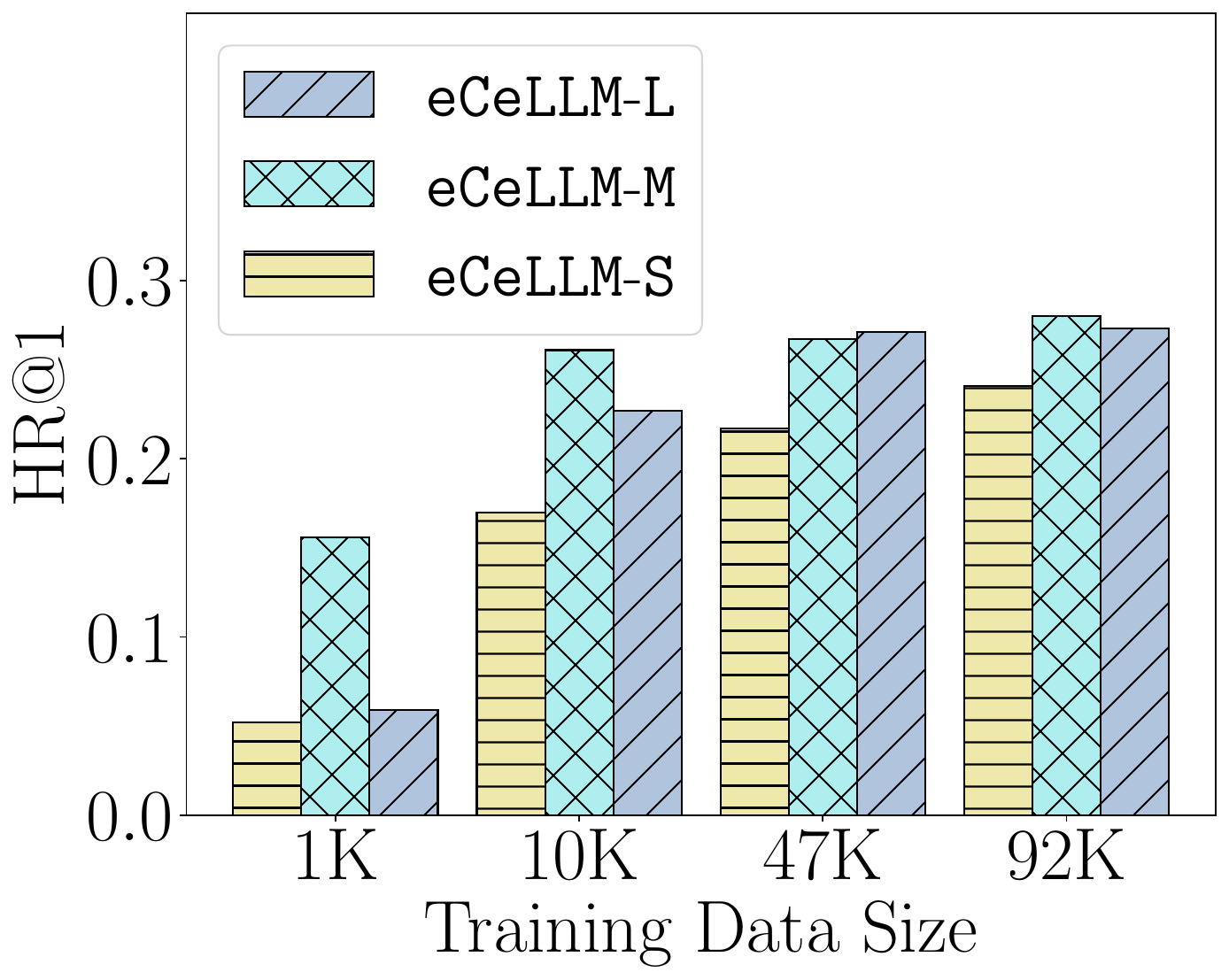}
		\vspace*{-12pt}
		\subcaption{OOD Evaluation}
		\label{fig:ood}
	\end{minipage}
\vspace{-8pt}
\caption{\method performance on \SR}
\label{fig:data_scaling}
\vspace{-20pt}
\end{figure}

We investigate how the training data size impacts the performance of \method models.
As presented in Section~\ref{sec:dataset:split}, 
\dataset has more than 92 thousand (92K) training samples.
In this experiment, we generate 3 smaller training sets of 1K, 10K, and 47K samples. 
The 1K and 10K samples are generated by randomly selecting 0.1K and 1K samples, respectively, from the training set of each of the 10 tasks. 
The 47K samples are constructed by randomly selecting 5K samples from each task
except for \EM, for which we include all its 2K training samples. 
%
%
The complete results are presented in Appendix~\ref{sec:appendix:data}.

Figure~\ref{fig:data_scaling} presents the \method model performance on \SR. 
As it shows, increasing training data size typically benefits \method model performance.
For example, in IND evaluation, the performance of \methodL increases from 0.085 to 0.526 when the training data size is increased from 1K to 92K.
Similarly, in OOD evaluation, increasing training samples from 1K to 92K significantly elevates \methodL performance from 0.059 to 0.273.
A similar trend could be observed in \methodB and \methodS.
As evidenced by our results, 
the large-scale training data is critical in developing effective \mbox{e-commerce} LLMs.
This further highlights the significance of our extensive, comprehensive, and high-quality e-commerce instruction dataset, {\dataset}.

%
%
%
%
%

\section{Conclusion and Limitation}
\label{sec:conclusions}

This paper open-sources the first  
large-scale, \mbox{high-quality} benchmark dataset (\dataset), 
and develops the \mbox{state-of-the-art} generalist LLMs (\method series) for \mbox{e-commerce}.
\method models are extensively evaluated against the most advanced baseline models including \mbox{GPT-4 Turbo}, EcomGPT, and SoTA task-specific models, in both IND and OOD evaluations.
Our experimental results demonstrate that 
\method substantially outperforms the best baseline models with an average improvement of 10.7\% in IND evaluation.
Our results also show that 
\method exhibits excellent generalizability to OOD settings, evidenced by its considerable improvement of 9.3\% over the best baselines on the OOD products.
%
%
To the best of our knowledge, this study is the first comprehensive and systematic study of instruction-tuning LLMs for e-commerce applications.

This paper acknowledges certain limitations and future work. 
(1) Recently emerged \mbox{e-commerce} tasks such as explanation generation could be included to further improve the comprehensiveness of \dataset.
(2) User profiling could be better enabled once metadata is available, and advanced foundation models 
could be developed to address unique and fundamental challenges and tasks in \mbox{e-commerce}. 

%
%
%

%



\section*{Impact Statement}
This paper presents work whose goal is to advance the field of Machine Learning. There are many potential societal consequences of our work, none of which we feel must be specifically highlighted here.

\bibliography{paper}
\bibliographystyle{icml2024}

\newpage
\appendix
\onecolumn

\setcounter{table}{0}
\setcounter{figure}{0}
\renewcommand{\thetable}{A\arabic{table}}
\renewcommand{\thefigure}{A\arabic{figure}}

\section{Task Definition and Data Preprocessing}
\label{sec:appendix:preprocessing}

All tasks can be characterized into 4 categories: 
product understanding, user understanding, query product matching, and product question answering (product QA).  
Table~\ref{tbl:task_data} summarizes the task definitions, evaluation metrics, data sources for each task, 
and their OOD test settings, if any. 
The details are articulated below.
\begin{table*}[h]
  \caption{Tasks and \dataset Datasets}
  \centering
  \vspace{2pt}
  \footnotesize
  \label{tbl:task_data}
  \begin{threeparttable}
      \begin{tabular}{
	@{\hspace{0pt}}l@{\hspace{10pt}}
	@{\hspace{0pt}}l@{\hspace{5pt}}
	@{\hspace{5pt}}p{0.35\textwidth}@{\hspace{5pt}}
	@{\hspace{5pt}}p{0.1\textwidth}@{\hspace{5pt}}
	@{\hspace{5pt}}p{0.15\textwidth}@{\hspace{5pt}}	
	@{\hspace{5pt}}p{0.25\textwidth}@{\hspace{0pt}}	
      }
      \toprule
      & Task & Definition  & Type & Metrics & Data \\
	\midrule
	\parbox[t]{2mm}{\multirow{11}{*}{\rotatebox[origin=c]{90}{Product Understanding}}}
       & \multirow{4}{*}{\AVE} 
       & Given the titles, descriptions, features, and brands of the products, extract values for the specific target attributes.
       & Information extraction
       & precision$^*$, recall$^*$, \underline{F1$^*$} (Section~\ref{sec:appendix:preprocessing:ave})
       & {MAVE~\cite{yang2022mave} based 
       on Amazon Review 2018~\cite{ni2019justifying}; 
       \textbf{OOD}: 7 held-out attributes}\\
       \cmidrule(lr){2-6}
       & \multirow{3}{*}{\IRP}
       & Given the titles of two products, predict their relation from \textit{``also buy", ``also view"}, and \textit{``similar"}.
       & Multi-class classification
       & accuracy, macro precision, macro recall, \underline{macro F1}
       & Amazon Review 2018~\cite{ni2019justifying}; 
       \textbf{OOD}: Tools category\\
       \cmidrule(lr){2-6}
       & \multirow{4}{*}{\EM}
       & Given the titles, descriptions, manufacturers, and prices of the products from two different platforms, predict if they are the same product.
       & Binary \mbox{classification}
       & accuracy, precision, recall, \underline{F1}, specificity, negative prediction rate
       & Amazon-Google Product~\cite{kopcke2010evaluation, Rahm2010EM} \\
	\midrule	
	\parbox[t]{2mm}{\multirow{8.5}{*}{\rotatebox[origin=c]{90}{User Understanding}}}
       &\multirow{3}{*}{\SA}
       & Given a product review by a user, identify the sentiment that the user expressed on the product.
       & Multi-class classification
       & accuracy, macro precision, macro recall, \underline{macro F1}
       & Amazon Review 2018~\cite{ni2019justifying}; \textbf{OOD}: Tools category\\
       \cmidrule(lr){2-6}
       & \multirow{5}{*}{\SR}
       & Given the interactions of a user over the products, predict the next product that the user would be interested in.
       & Ranking
       & \underline{HR@1}
       & Amazon Review 2018~\cite{ni2019justifying} and Amazon Review 2014~\cite{he2016ups, mcauley2015image}; \textbf{OOD}: Tools category\\
	\midrule
	\parbox[t]{2mm}{\multirow{11}{*}{\rotatebox[origin=c]{90}{Query Product Matching}}}
       &\multirow{3}{*}{\MPC}
       & Given a query and a product title, predict the relevance between the query and the product.
       & Multi-class classification
       & accuracy, macro precision, macro recall, \underline{macro F1}
       & Shopping Queries Dataset~\cite{reddy2022shopping} \\
       \cmidrule(lr){2-6}
       & \multirow{4}{*}{\PSI}
       & Given a user query and a potentially relevant product, predict if the product can serve as a substitute for the user’s query.
       & Binary \mbox{classification}
       & accuracy, precision, recall, \underline{F1}, specificity, negative prediction rate
       & Shopping Queries Dataset~\cite{reddy2022shopping}\\
       \cmidrule(lr){2-6}
       & \multirow{3}{*}{\QPR}
       & Given a user query and a list of potentially relevant products to the query, rank the products according to their relevance to the query.
       & Ranking
       & \underline{NDCG}$^{[1]}$
       & Shopping Queries Dataset~\cite{reddy2022shopping}\\
	\midrule
	\parbox[t]{2mm}{\multirow{7}{*}{\rotatebox[origin=c]{90}{Product QA}}}
       &\multirow{4}{*}{\AP}
       & Given a product-related question and reviews of this product, predict if the question is answerable.
       & Binary \mbox{classification}
       & accuracy, precision, recall, \underline{F1}, specificity, negative prediction rate
       & AmazonQA~\cite{gupta2019amazonqa}; 
       \textbf{OOD}: Cells category\\
       \cmidrule(lr){2-6}
       & \multirow{3}{*}{\AG}
       & Given a product-related question and reviews as supporting documents, generate the answer to the question.
       & Generation
       & P$_{\text{BERT}}$$^{[2]}$, R$_{\text{BERT}}$$^{[2]}$, \underline{F$_{\text{BERT}}$}$^{[2]}$, BLEURT$^{[3]}$
       & AmazonQA~\cite{gupta2019amazonqa}; 
       \textbf{OOD}: Cells category\\
      \bottomrule
      \end{tabular}
      \begin{tablenotes}[normal,flushleft]
      \begin{footnotesize}
      \item \underline{Underline} indicates the primary metrics. Higher values of the primary metrics indicate better model performance. \textbf{OOD} refers to the out-of-domain product category/attributes.
      [1] NDCG: normalized discounted cumulative gain~\cite{wang2013theoretical} 
      [2] P$_{\text{BERT}}$, R$_{\text{BERT}}$, F$_{\text{BERT}}$:
      BERTScore evaluates the quality of generated texts by measuring the similarity between the embeddings of tokens in the generated text and the \mbox{ground-truth} text~\cite{zhang2019bertscore}. 
      [3] BLEURT measures the text quality by comparing the similarity of sentences using contextual embeddings from pre-trained models.
      ~\cite{sellam2020bleurt}.
    \par
      \end{footnotesize}
      \end{tablenotes}
  \end{threeparttable}
\end{table*}


To pursue adherence to data usage requirements, we check the licenses of \dataset data sources, ensuring their permission to publish. Table \ref{tbl:license} presents the licenses of our curated dataset sources.

\begin{table*}[!h]
  \caption{Details of Data Source License}
  \vspace{2pt}
  \centering
  \label{tbl:license}
  \footnotesize
  \begin{threeparttable}
      \begin{tabular}{
        @{\hspace{2pt}}l@{\hspace{4pt}}
        @{\hspace{4pt}}l@{\hspace{3pt}}
        @{\hspace{3pt}}l@{\hspace{0pt}}
      }
      \toprule
      Dataset & License Type & Source \\
      \midrule
      Amazon-Google Products & CC-by-4.0 & \href{{https://dbs.uni-leipzig.de/research/projects/benchmark-datasets-for-entity-resolution}}{https://dbs.uni-leipzig.de/research/projects/benchmark-datasets-for-entity-resolution} \\
      Amazon Review & Not Specified & \href{https://cseweb.ucsd.edu/~jmcauley/datasets.html#amazon_reviews}{https://cseweb.ucsd.edu/\~jmcauley/datasets.html\#amazon\_reviews} \\
      AmazonQA & Not Specified & \href{https://github.com/amazonqa/amazonqa}{https://github.com/amazonqa/amazonqa} \\
      Shopping Queries Dataset & Apache License 2.0 & \href{https://github.com/amazon-science/esci-data}{https://github.com/amazon-science/esci-data} \\
      
      \bottomrule
      \end{tabular}
  \end{threeparttable}
\end{table*}


\paragraph{Data Split}
Raw datasets of the attribute value extraction (\AVE, discussed below in Section \ref{sec:appendix:preprocessing:ave}), product matching (\EM, discussed in Section \ref{sec:appendix:preprocessing:em}), product relation prediction (\IRP, discussed in Section \ref{sec:appendix:preprocessing:irp}), and sentiment analysis (\SA, discussed in Section \ref{sec:appendix:preprocessing:sa}) tasks are first split into training, validation, and test data at 8:1:1 ratio. 
For multi-class product classification (\MPC, discussed in Section \ref{sec:appendix:preprocessing:mpc}), query product ranking (\QPR, discussed in Section \ref{sec:appendix:preprocessing:qpr}), product substitute identification (\PSI, discussed in Section \ref{sec:appendix:preprocessing:psi}), answerability prediction (\AP, discussed in Section \ref{sec:appendix:preprocessing:ap}), and answer generation (\AG, discussed in Section \ref{sec:appendix:preprocessing:ag}) tasks, the raw datasets are already split.
All split data are processed as detailed below and converged with instruction templates into structured data, which will be directly used by LLMs.
Based on prior research~\cite{wei2021finetuned} and considering the high computing demands, 
we uniformly downsample training sets to a size of 10K, validation sets to 1K, and test sets to 1K size, optimizing data volumes for efficient processing and affordable LLM evaluation. 

%
\paragraph{In-domain (IND) Data Selection}
In \IRP, \SA, and \SR tasks, we utilize the Amazon Review 2018 dataset~\cite{ni2019justifying} as the data source, which contains product reviews and metadata in 29 different product categories. 
For these three tasks, the data from the Electronics, Home, and Sports categories are employed as the IND data sources. We further downsample and combine the processed structured data for each task from these sources at a 1:1:1 ratio into a training, validation, and test set, respectively.
For \AP and \AG tasks, we use the data from the AmazonQA dataset~\cite{gupta2019amazonqa}, which has 17 product categories. To avoid data leakage, we use the data from the Sports and Tools categories for the \AP task, and Electronics and Home for \AG task as IND data sources, respectively. 
Similarly to the previous three tasks, for each of the tasks, we sample and combine the processed data from the IND sources at a 1:1:1 ratio into training, validation, and test sets.
For the \EM task, we process all data from Amazon-Google Products~\cite{kopcke2010evaluation, Rahm2010EM} and get a 2K training set, 0.2K validation set, and 0.2K test set. Since this dataset is small, we do not do the downsampling for the \EM task.
For \AVE, \MPC, \QPR, and \PSI tasks, we directly process and downsample them from the corresponding split in original datasets, separately.

\paragraph{Out-of-domain (OOD) Data Selection}
As for OOD datasets, only test data are utilized. The data from the Tools category from the Amazon Review dataset is used as the OOD data source in \IRP, \SA, and \SR tasks, while the Cells category from AmazonQA serves as the OOD source in \AP and \AG tasks. For the \AVE task, we hold out 7 attributes as the OOD dataset. All OOD sources are processed into 1K test sets.

\begin{table}[!h]
  \caption{Summary of the   \dataset Dataset and Tests}
  \centering
  \label{tbl:data_summary}
  \begin{footnotesize}
  \begin{threeparttable}
      \begin{tabular}{
	@{\hspace{3pt}}l@{\hspace{3pt}}
	@{\hspace{3pt}}c@{\hspace{3pt}}
	@{\hspace{3pt}}c@{\hspace{3pt}}
	@{\hspace{3pt}}c@{\hspace{3pt}}
	@{\hspace{3pt}}c@{\hspace{3pt}}
      }
      \toprule
       Task & {Training}  & {Validation}  & {IND Test}  & {OOD Test}  \\
	 \midrule
       \AVE & 10,000 & 1,000 & 1,000 & 1,000  \\
       \EM & \textcolor{white}{0}2,022 & \textcolor{white}{0,}253 & \textcolor{white}{0,}253 & \ding{55} \\
       \IRP & 10,000 & 1,000 & 1,000 & 1,000  \\
       \SA & 10,000 & 1,000 & 1,000 & 1,000  \\
       \SR & 10,000 & 1,000 & 1,000 & 1,000  \\
       \MPC & 10,000 & 1,000 & 1,000 & \ding{55} \\
       \QPR & 10,000 & 1,000 & 1,000 & \ding{55}\\
       \PSI & 10,000 & 1,000 & 1,000 & \ding{55}\\
       \AP & 10,000 & 1,000 & 1,000 & 1,000  \\
       \AG & 10,000 & 1,000 & 1,000 & 1,000  \\
       \midrule
       \dataset & 92,022 & 9,253 & 10 tasks & 6 tasks\\
      \bottomrule
      \end{tabular}
      \begin{tablenotes}[normal,flushleft]
      \begin{footnotesize}
      \item 
      In this table, IND and OOD refers to the in-domain evaluation and out-of-domain evaluation, respectively. 
      \par
      \end{footnotesize}
      \end{tablenotes}
  \end{threeparttable}
  \end{footnotesize}
\end{table}

\subsection{Product Understanding Tasks}
\label{sec:appendix:preprocessing:pu}

Three tasks are defined as follows to understand the different aspects of products.

\subsubsection{Attribute Value Extraction (\AVE)}
\label{sec:appendix:preprocessing:ave}
    \paragraph{Definition:} Given the titles, descriptions, features, and brands of the products, extract values for the specific target attributes. 
    By understanding product attribute values, models can extract key properties of the products and build the profiling for them, which is beneficial in many e-commerce scenarios, such as customer service agents and explanations.
    \paragraph{Data Processing:} The Amazon Review 2018 dataset~\cite{ni2019justifying} is used as the raw data. We extract the ground-truth attribute-value pairs following the protocol of MAVE~\cite{yang2022mave} and get 4,767,579 data entries with 683 attributes in total. 
    We randomly hold out 7 attributes from the dataset as the OOD data source and process both the OOD data and remaining IND data with instructions. 
    The samples that have ground-truth values and corresponding sources for the specific attribute in the dataset are referred to as positive samples, while the samples that have no extracted ground-truth values are denoted as negative samples.
    \paragraph{Evaluation Metrics:} Following the evaluation definition in previous work MAVE~\cite{yang2022mave}, the model can predict null value (NV) or incorrect value (IV) for negative samples.
    For positive samples, the model predictions can be correct value (CV), wrong value (WV), and null value (NL). 
    In MAVE, customized precision and recall are calculated as the percentage of correctly predicted positive samples in all predictions and ground truths, respectively.
    This formulation, however, fails to evaluate model results on negative samples.
    Therefore, with the definition above, we improve the formulation by considering both positive and negative samples as follows:
    \begin{equation}
        \label{eqn:Fave}
        \text{precision}^\ast = \frac{\text{NV + CV}}{\text{NV + IV + CV + WV}},\ 
        \text{recall}^\ast = \frac{\text{NV + CV}}{\text{N}},\ 
        \text{F1}^\ast = \frac{2 \times \text{precision}^\ast \times \text{recall}^\ast}{\text{precision}^\ast + \text{recall}^\ast},
    \end{equation}
    where N refers to the total number of entries in ground-truth data. 
    Note that we only consider the prediction result as the correct value (VC) when the ground truth is fully contained in the prediction (e.g. ``bright yellow'' is considered as a correct value with the ground truth of ``yellow'').
    In our formulation, the precision$^\ast$ is calculated as the percentage of correctly predicted positives and negatives in all predictions, and the recall$^\ast$ is computed as the percentage of correct positive and negative predictions in all ground truths. We use F1$^\ast$ as the primary metric for the AVE task.

    \paragraph{Example:} 
        \begin{itemize}
            \item {\small \texttt{Input: }}
                \begin{itemize}
                    \item {\small \texttt{Product title: Bencore Multi Functional Molle Tactical Messenger Bag.}}
                    \item {\small \texttt{Product description: This rugged/durable tactical shoulder bag provides perfect and stylish solution for almost any scenario. The bag is made of durable nylon construction that will not tear, color will not fade. The bag has many MOLLE straps through the bag for all your MOLLE accessories. The bag contains many roomy compartments as pictured and comes in many stylish colors. Design, comfort and functionality was the emphasis of this bag which is why we made sure the bag is fully ergonomic, lightweight and has many roomy pockets and Velcro patches throughout the bag. The product comes with the Bencore Life Time warranty and is satisfaction guaranteed. Bencore is a leading manufacturer in outdoor Apparel/Accessories, from Par cords to Backpacks to basic outdoor essentials.}}
                    \item {\small \texttt{Product feature: Durable heavy-duty, lightweight Nylon construction, will not tear or break even under extreme conditions - Lifetime Warranty, Rugged, roomy main drawstring-closed compartment provides secure storage space for your gear; MOLLE System, works with most MOLLE accessory, Front pocket provides quick access, roomy interior pocket for convenient separated storage, concealed back pocket with zipper closure, Padded and fully ergonomic System, adjustable shoulder strap for comfortable handling.}}
                    \item {\small \texttt{Product brand: Bencore}}
                    \item {\small \texttt{Target attributes: Material}}
                \end{itemize}
            \item {\small \texttt{Output:}}
                \begin{itemize}
                    \item {\small \texttt{Attribute: material; Value: nylon; Source: product description.}}
                    \item {\small \texttt{Attribute: material; Value: nylon; Source: product feature.}}
                \end{itemize}
        \end{itemize}
    Note that in the above example, for the ``target attribute" ``material", the model should extract its value ``nylon" from the ``product description" and ``product feature".

\subsubsection{Product Relation Prediction (\IRP)}
\label{sec:appendix:preprocessing:irp}

\paragraph{Definition:} Given the titles of two products, predict their relation. Studying the relations between products can help models generate better results when conducting other e-commerce tasks such as recommendations.
    
\paragraph{Data Processing:} To learn the relationship between products, we use the product metadata of Electronics, Home, and Sports categories from the Amazon Review 2018 dataset~\cite{ni2019justifying} as the IND sources and Tools as the OOD source. 
We collect product IDs from metadata and remove the products without detailed information in the metadata. In this task, the product titles are used to represent products.
The product pairs that appear more than once with different relations are eliminated. 
After filtering and combining data with instruction templates, the three relations (\textit{also buy, also view, and similar}) in the structured dataset are roughly 7:10:1.
    
\paragraph{Evaluation Metrics:} With three different relations in this task, accuracy, macro precision, macro recall, and macro F1 are employed as metrics. 
Meanwhile, macro F1 is used as the primary metric for combining evaluation results of different labels and providing a comprehensive measurement of the model performances.
    
\paragraph{Example:} 
    \begin{itemize}
        \item {\small \texttt{Input: }}
            \begin{itemize}
                \item {\small \texttt{Product 1: Monoprice 11952 Polyurethane Replacement Ear Pads for PID 8323 type Headphones - Red}}
                \item {\small \texttt{Product 2: Monoprice Hi-Fi Light Weight Over the Ear Headphones - Black with a 50mm driver and a 47in 3.5mm cable for Apple iPhone iPod Android Smartphone Samsung Galaxy Tablets MP3}}
            \end{itemize}
        \item {\small \texttt{Options:}}
            \begin{itemize}
                \item {\small \texttt{A. Users who view product 1 may also buy product 2.}}
                \item {\small \texttt{B. Users who view product 1 may also view product 2.}}
                \item {\small \texttt{C. The product 1 is similar with the product 2.}}
            \end{itemize}
        \item {\small \texttt{Output:}}
            \begin{itemize}
                \item {\small \texttt{B}}
            \end{itemize}
    \end{itemize}
    Note that in the above example, the relation of Product 1 and Product 2 is \textit{also\_view}.

\subsubsection{Product Matching (\EM)}
\label{sec:appendix:preprocessing:em}

\paragraph{Definition:} Given the titles, descriptions, manufacturers, and prices of the products from two different platforms, predict if they are the same product. This task enables the model to learn the similarities among products.
    
\paragraph{Data Processing:} We use the original data~\cite{kopcke2010evaluation, Rahm2010EM}, which contains detailed product information from Amazon and Google platforms, as well as 1.3K matching product pairs between the two platforms. After deduplication, we randomly sample the same amount of unmatched pairs since there are only matched samples in the raw data and thus get 2,530 product pairs in total. 
The product pairs are split into the training, validation, and test sets with a ratio of 8:1:1, and processed with instruction templates respectively.
    
\paragraph{Evaluation Metrics:} For this binary classification task, we use precision (positive prediction rate), recall (sensitivity), F1, specificity (recall of negative labels), negative prediction rate (NPR, precision of negative labels), and accuracy as metrics. 
We choose F1 as the primary metric since it serves as a balanced evaluation metric by combining precision and recall to generally reflect model performance.
    
\paragraph{Example:} 
    \begin{itemize}
        \item {\small \texttt{Input: }}
            \begin{itemize}
                \item {\small \texttt{Product 1: title - marine aquarium 2.5 virtual undersea paradise win/mac, description - marine aquarium 2.0 is like having a small piece of an aquatic paradise in your home -- without having to take care of actual fish, manufacturer - encore software, price - 19.99}}
                \item {\small \texttt{Product 2: title - encore software 25020 - marine aquarium 2.5 (hybrid) - win 95 98 me 2000 xp/mac 10.1 or higher, description - encore software 25020: marine aquarium 2.5 hybrid discover the virtual fish tank phenomenon that has everyone talking! marine aquarium 2.5 delivers a stunning undersea paradise through your desktop with 26 exotic species of fish, manufacturer - encore software, price - 19.97}}
            \end{itemize}
        \item {\small \texttt{Output:}}
            \begin{itemize}
                \item {\small \texttt{Yes}}
            \end{itemize}
    \end{itemize}
    In the above example, the two products (Product 1, Product 2) are identified as the same product.

\subsection{User Understanding}
\label{sec:appendix:preprocessing:uu}
The two tasks in this section aim to help the models comprehend the users' needs and preferences.
\subsubsection{Sentiment Analysis (\SA)}
\label{sec:appendix:preprocessing:sa}

    \paragraph{Definition:} Given a product review by a user, identify the sentiment that the user expressed on the product. The task will help models understand what sentiment users express and recommend more proper products to the user
    
    \paragraph{Data Processing:} For the sentiment analysis, we also use the review data of Electronics, Home, and Sports categories from the Amazon Review 2018 dataset~\cite{ni2019justifying} as the IND sources and Tools category as the OOD source. 
    We only keep the data with reviews of at least 10 words. 
    The ratings are used as the ground-truth sentiment level, while 5.0 refers to very positive and 1.0 refers to very negative. 
    After downsampling and combining the raw data with instruction templates, the five labels (from 1.0 to 5.0) in the structured dataset are roughly 4:2:3:8:25.
    
    \paragraph{Evaluation Metrics:} With five different labels in the \SA task, accuracy, macro precision, macro recall, and macro F1 are employed as metrics. macro F1 is used as the primary metric.
    \paragraph{Example:} 
        \begin{itemize}
            \item {\small \texttt{Input: }}
                \begin{itemize}
                    \item {\small \texttt{This is really perfect for my kids who have a thick hair. I cam be able to create a beautiful hair bun with them. I would like to recommend this to all.}}
                \end{itemize}
            \item {\small \texttt{Options:}}
                \begin{itemize}
                    \item {\small \texttt{A. Very positive}}
                    \item {\small \texttt{B. Positive}}
                    \item {\small \texttt{C. Neutral}}
                    \item {\small \texttt{D. Negative}}
                    \item {\small \texttt{E. Very negative}}
                \end{itemize}
            \item {\small \texttt{Output:}}
                \begin{itemize}
                    \item {\small \texttt{A}}
                \end{itemize}
        \end{itemize}
        The user from the above example expressed a very positive sentiment in the review.

\subsubsection{Sequential Recommendation (\SR)}
\label{sec:appendix:preprocessing:sr}
    \paragraph{Definition:} Given the interactions of a user over the products, predict the next product that the user would be interested in. By learning on this task, the models will have a comprehensive view of user preferences, which enables models to cater to users' future needs.
    
    \paragraph{Data Processing:} In the \SR task, we use both product reviews and metadata from the Amazon Review 2018 dataset~\cite{ni2019justifying}. Meanwhile, we use metadata from the Amazon Review 2014 dataset~\cite{he2016ups, mcauley2015image} as a supplement. 
    The data of Electronics, Home, and Sports categories from both datasets serve as IND sources and the Tools category is used for the OOD test.
    Moreover, we consider users' review histories as their interactions with products.
    Following the data processing protocol of UnisRec~\cite{hou2022towards}, we remove the products without metadata in the 2018 version dataset and conduct the 5-core filter to ensure all products and users appear at least 5 times. 
    After filtering, we sort every user's interactions chronologically and truncate the history with a maximum of 50 products, retaining the least recent history.
    For the text information of products, we use the metadata from Amazon Review 2014 to fill in the missing information of the same products in the 2018 version metadata.

    We also combine the product title, category, and brand to represent a product. The average length of the combined texts is about 21 words. Thus, we retain the first, maximum of 25 words of each combined text for computational efficiency. 
    %
    As in conventional sequential recommendation tasks, we split the last product of the user interactions into the test set as the ground truth next product of the user's interest, the second last product into the validation set, and the remaining products into the training set.
    When processing the user interactions with instruction templates, for each sample, we randomly select 19 candidate products and mix them up with one ground-truth product. These 20 products serve as the options for the sample.

    \paragraph{Evaluation Metrics:} We evaluate the \SR task on hit rate at top 1 (HR@1), which is a popular metric in sequential recommendation and measures if the top-ranked product matches the ground-truth user interaction. HR@1 also serves as the primary metric in this task.

    \paragraph{Example:} 
        \begin{itemize}
            \item {\small \texttt{Input: }}
                \begin{itemize}
                    \item {\small \texttt{1st: M-Edge Latitude Kindle Jacket, Pink (Fits Kindle Keyboard). Electronics. Computers \& Accessories. M-Edge.}}
                    \item {\small \texttt{2nd: Marware jurni Kindle Fire Case Cover, Black (will not fit HD or HDX models). Electronics. Computers \& Accessories. Marware.}}
                    \item {\small \texttt{3rd: NETGEAR AC1600 Dual Band Wi-Fi Gigabit Router (R6250). Electronics. Computers \& Accessories. NETGEAR.}}
                    \item {\small \texttt{4th: iMBAPrice 110014-1 (1-Pack) Glod Plated 2.4 Ghz 3-Way Coaxial Cable Splitter F-Type Screw for Video Satellite Splitter/VCR/Cable Splitter/TV Splitter/Antenna Splitter/RG6 Splitter. Electronics. Accessories \& Supplies...}}
                \end{itemize}
            \item {\small \texttt{Options:}}
                \begin{itemize}
                    \item {\small \texttt{A: T POWER 9v~12v (6.6ft Long Cable) Ac Dc Adapter Compatible with X Rocker Pro Series H3 51259 Video Gaming Chair 51231,51396 \& V Rocker 5130301...}}
                    \item {\small \texttt{B: Boys Floatsafe Flotie Soft Fabric Armbands Floatie Blue For Kids Ages 1 To 3. Floatsafe Floatie}}
                    \item {\small \texttt{C: Anker iPhone Charger, Powerline Lightning Cable (3ft), MFi Certified for iPhone Xs/XS Max/XR/X}}
                    \item {\small \texttt{D: Curtain Drapery Rod w/brackets Small - Wrought Iron Hand Made. Home \& Kitchen. Home Dcor. Hand Crafted \& American Made!}}
                    \item {\small \texttt{\dots}}
                    \item {\small \texttt{T: Lorex ACCMIC1 Indoor Audio Microphone Accessory for Surveillance DVR's (Black). Electronics. Camera \& Photo. Lorex}}
                \end{itemize}
            \item {\small \texttt{Output:}}
                \begin{itemize}
                    \item {\small \texttt{A}}
                \end{itemize}
        \end{itemize}
        Given the interactions shown in the above input, the next product of user's interest for this user is C: Anker iPhone Charger, Powerline Lightning Cable (3ft), MFi Certified for iPhone Xs/XS Max/XR/X

\subsection{Query Product Matching}
\label{sec:appendix:preprocessing:qps}

The following three tasks seek to study the relations between the user queries and the potential relevant products to the queries.
We use the data from the Shopping Queries Dataset~\cite{reddy2022shopping}, which contains 48,300 queries and potentially relevant products to each query. The dataset was originally collected from different market locales in various languages. All products are labeled in four categories: {\emph{Exact, Substitute, Complement, and Irrelevant}}, based on their relevance to the query. As in the Shopping Queries Dataset, the four labels are defined as follows:

\begin{itemize}
    \item \emph{Exact}: The item is relevant for the query, and satisfies all the query specifications (e.g., a water bottle matching all attributes of a query “plastic water bottle 24oz”, such as material and size).
    \item \emph{Substitute}: The item is somewhat relevant, i.e., it fails to fulfill some aspects of the query, but the item can be used as a functional substitute (e.g., fleece for a “sweater” query)
    \item \emph{Complement}: The item does not fulfill the query, but could be used in combination with an exact item (e.g., track pants for “running shoes” query).
    \item \emph{Irrelevant}: The item is irrelevant, or it fails to fulfill a central aspect of the query (e.g., socks for a “telescope” query, or a wheat flour bread for a “gluten-free bread” query)
\end{itemize}

In the following query-related tasks, we only utilize the data that is both in English and from the market of the U.S. locale.

\subsubsection{Multi-class Product Classification (\MPC)}
\label{sec:appendix:preprocessing:mpc}

    \paragraph{Definition:} Given a query and a product title, predict the relevance between the query and the product (\emph{Exact, Substitute, Complement, Irrelevant}). This task helps models learn the fine-grained relevance between queries and products, promoting better recommendation results.
    \paragraph{Data Processing:} We use the product titles to represent products and translate the Unicode into English. The translated query-product pairs are processed with instruction templates. The ratio of the four labels ({\emph{Exact, Substitute, Complement, and Irrelevant}}) in the structured dataset is about 35:10:1:5.
    \paragraph{Evaluation Metrics:} Accuracy, macro precision, macro recall, and macro F1 are employed as metrics. Accuracy is used as the primary metric in the \MPC task.
    \paragraph{Example:} 
        \begin{itemize}
            \item {\small \texttt{Input: }}
                \begin{itemize}
                    \item {\small \texttt{Query: aj1 black and white}}
                    \item {\small \texttt{Product: Nike Men's Air Jordan 1 Low White/Gym Red, White/Gym Red/Black, 9}}
                \end{itemize}
            \item {\small \texttt{Options:}}
                \begin{itemize}
                    \item {\small \texttt{A: The product is relevant to the query, and satisfies all the query specifications.}}
                    \item {\small \texttt{B: The product is somewhat relevant. It fails to fulfill some aspects of the query but the product can be used as a functional substitute.}}
                    \item {\small \texttt{C: The product does not fulfill the query, but could be used in combination with a product exactly matching the query.}}
                    \item {\small \texttt{D: The product is irrelevant to the query.}}
                \end{itemize}
            \item {\small \texttt{Output:}}
                \begin{itemize}
                    \item {\small \texttt{B}}
                \end{itemize}
        \end{itemize}
        In the above example, the product serves as a substitute for the product described in the query.

\subsubsection{Product Substitute Identification (\PSI)}
\label{sec:appendix:preprocessing:psi}

\paragraph{Definition:} Given a user query and a potentially relevant product, predict if the product can serve as a substitute for the user’s query.
\paragraph{Data Processing:} The preprocessing of the \PSI is similar to that of \MPC, except that the PSI is a binary classification task. The query-product pairs with {\emph{Exact, Complement, or Irrelevant}} labels are relabeled as non-substitute. After combining the query-product-label triples with instruction templates, the ratio of \emph{substitute} (positive) and \emph{non-substitute} (negative) labels is approximately 4:1.
\paragraph{Evaluation Metrics:} For this binary classification task, we use precision (positive prediction rate), recall (sensitivity), F1, specificity (recall of negative labels), negative prediction rate (NPR, precision of negative labels), and accuracy as metrics. We choose F1 as the primary metric.
\paragraph{Example:} 
    \begin{itemize}
        \item {\small \texttt{Input: }}
            \begin{itemize}
                \item {\small \texttt{Query: fissler magic smooth-edge can opener}}
                \item {\small \texttt{Product: KUKINO Manual Can Opener, Multifunction Handheld Food Grade Stainless Steel Can Openers, Black.}}
            \end{itemize}
        \item {\small \texttt{Output:}}
            \begin{itemize}
                \item {\small \texttt{No}}
            \end{itemize}
    \end{itemize}
    The product does not serve as a substitute for the query in the above example.

\subsubsection{Query-product Ranking (\QPR)}
\label{sec:appendix:preprocessing:qpr}

\paragraph{Definition:} Given a user query and a list of potentially relevant products to the query, rank the products according to their relevance to the query.
\paragraph{Data Processing:} In the \QPR task, we utilize the product titles to represent the products. A query and a list of potentially relevant products with different relevance labels to this query constitute a sample. We sort the list of products according to their relevance and generate the ground truth for each sample.
\paragraph{Evaluation Metrics:} We evaluate this task using normalized discounted cumulative gain (NDCG)~\cite{wang2013theoretical}, which is a prevalent evaluation metric for assessing the ranking quality in the existing literature. NDCG is also used as the primary metric in this task.
\paragraph{Example:} 
    \begin{itemize}
        \item {\small \texttt{Input: }}
            \begin{itemize}
                \item {\small \texttt{Query: high heel shoe chair}}
                \item {\small \texttt{Product A: ORE International HBB1826 High Heel Shoe Display with Hooks Jewelry Box, Cheetah Print.}}
                \item {\small \texttt{Product B: Coconut Float Red High Heel Gigantic Pool Float for Adults, 91.}}
                \item {\small \texttt{Product C: Wildkin Kids Wooden Bench Seat with Storage for Boys and Girls, Toy Box Bench Seat Features Safety Hinge, Backrest, and Two Carrying Handles, Measures 32 x 15.5 x 27 Inches (Wild Side) (LOD71001).}}
            \end{itemize}
        \item {\small \texttt{Output:}}
            \begin{itemize}
                \item {\small \texttt{A, C, B}}
            \end{itemize}
    \end{itemize}
    In the above example, Product A is more relevant to the query than Product C, and Product C is more relevant than Product B.

\subsection{Product Question Answering}
\label{sec:appendix:preprocessing:pqa}

These two tasks aim to learn from the products and answer the product-related questions by using the AmazonQA dataset~\cite{gupta2019amazonqa}. This dataset involves 923,685 product-related questions. There are around 9 product-related reviews and 4 answers provided by Amazon users for each question. There are also \textit{questionType} and \textit{is\_answerable} annotations for each question. We eliminate all meaningless notations such as HTML tags or emoji from questions, reviews, and answers.

\subsubsection{Answerability Prediction (\AP)}
\label{sec:appendix:preprocessing:ap}
    \paragraph{Definition:} Given a product-related question and reviews of this product, predict if the question is answerable.

    \paragraph{Data Processing:} We use the data of the Sports and Tools categories from AmazonQA as the IND sources and the Cell category as the OOD source. The \textit{is\_answerable} annotations are used as the ground truth. 
    The ratio of answerable (positive) and unanswerable (negative) samples in the structured dataset is around 2:3.
    
    \paragraph{Evaluation Metrics:} For the \AP task, we use precision (positive prediction rate), recall (sensitivity), F1, specificity (recall of negative labels), negative prediction rate (NPR, precision of negative labels), and accuracy as metrics. We use F1 as the primary metric when evaluating this task.
    
    \paragraph{Example:} 
        \begin{itemize}
            \item {\small \texttt{Input: }}
                \begin{itemize}
                    \item {\small \texttt{Question: Where do you purchase the paddles or do paddles come with it?}}
                    \item {\small \texttt{Document: Very happy with purchase and price! \textbackslash My son spends hours playing with this. It was easy to assemble and he loves it! Very happy with the purchase. \textbackslash You won't regret this purchase! A little awkward to assemble, as the instructions say you need 2 people to do it. \textbackslash This is well built. Great value. Fun and exercise for the whole family! Buy it today and have it for years to come. \textbackslash Sturdy, well made and will be around for many years to come! Totally awesome for my son to be able to play ball on his own sometimes ;)}}
                \end{itemize}
            \item {\small \texttt{Output:}}
                \begin{itemize}
                    \item {\small \texttt{No}}
                \end{itemize}
        \end{itemize}
        From the above example, the question is unanswerable based on the document.

\subsubsection{Answer Generation (AG)}
\label{sec:appendix:preprocessing:ag}
\paragraph{Definition:} Given a product-related question and reviews as supporting documents, generate the answer to the question.
\paragraph{Data Processing:} This task only involves answerable and open-ended questions. To avoid data leakage, we use the Electronics and Home categories, which are different from the \AP task, as the IND sources and the Cells category as the OOD source.
Every data entry in the AmazonQA has a list of answers labeled with \textit{helpfulness} levels.
We choose the most helpful answer as the ground truth.
\paragraph{Evaluation Metrics:} The \AG task is evaluated on precision, recall, and F1 of the BERTScore~\cite{zhang2019bertscore}, and BLEURT~\cite{sellam2020bleurt}. BERTScore and BLEURT both evaluate the similarity of sentences by leveraging contextual embeddings from pre-trained models and are widely used in the NLP field. In this task, we use F1 BERTScore as the primary metric.
\paragraph{Example:} 
    \begin{itemize}
        \item {\small \texttt{Input: }}
            \begin{itemize}
                \item {\small \texttt{Question: Can you add additional receiver with just one sensor? So one sensor picks up the the signal and sends it to two receiver.}}
                \item {\small \texttt{Document: I have a 1200ft driveway and the unit works perfectly. The feature that is missing is the option to have more than one notification pattern if you have more than one sensor. For example, 1, 2, 3 or 4 beeps would tell you which area the motion is coming from. If you want a reliable motion sensor that works for a long distance then this is your unit. \textbackslash Installed this system about two weeks ago, 300 feet from house toward end of driveway and it has never failed. Worked in rain with no problems or false alarms. There is almost 40 feet of drive remaining and I installed this at slight angle up the driveway. Larger vehicles (e.g. garbage truck, tractors mowers, etc) that drive slowly by at end of drive will also initiate the alarm. Faster or smaller vehicles on road will not be picked up, makes it really nice. Have two receivers, one indoor and one out back, makes it really a valuable alarm for us. \textbackslash I bought several brands of alarm systems. for the money I can't see how you would be disappointed for the cost. I have this unit about 200' from the receiver, and it works great. \textbackslash $\dots$}}
            \end{itemize}
        \item {\small \texttt{Output:}}
            \begin{itemize}
                \item {\small \texttt{Yes...just be sure all have the same dip switch settings.}}
            \end{itemize}
    \end{itemize}
    Note that the answer to the above question that is semantically similar to the output should be generated from the document.

\section{Data Statistics}
\label{sec:appendix:datastats}

The comprehensive and wide-ranging tasks in \dataset are critical for developing versatile e-commerce foundation models. 
\dataset includes products from 21 categories, such as Sports, Electronics, and Home. 
This broad coverage ensures that \dataset encompasses diverse product information, facilitating the creation of generalist e-commerce models.


\begin{figure}[htbp]
    \centering
      \begin{subfigure}{0.24\textwidth}
        \centering   
        \includegraphics[width=0.9\linewidth]{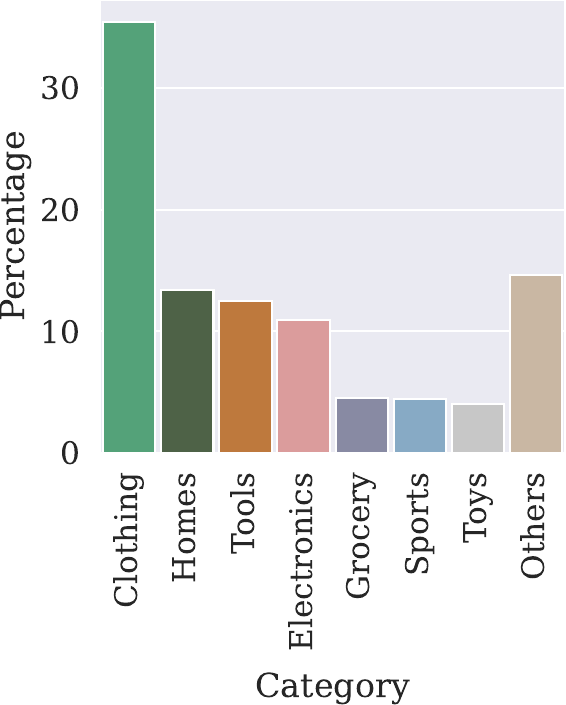}
          \caption{\AVE}
          \label{fig:cat_ave}
      \end{subfigure}
      \begin{subfigure}{0.24\textwidth}
        \centering   
        \includegraphics[width=0.9\linewidth]{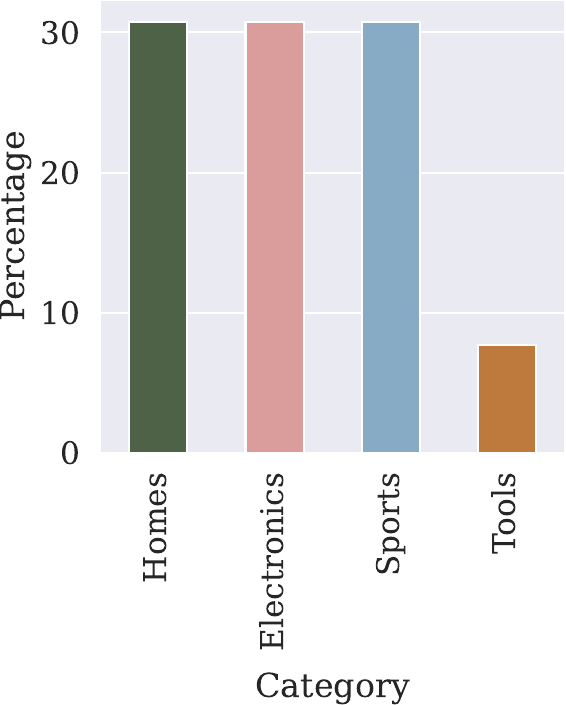}
          \caption{\IRP, \SA, and \SR}
          \label{fig:cat_irp}
      \end{subfigure}
	  \begin{subfigure}{0.24\textwidth}
        \centering   
        \includegraphics[width=0.9\linewidth]{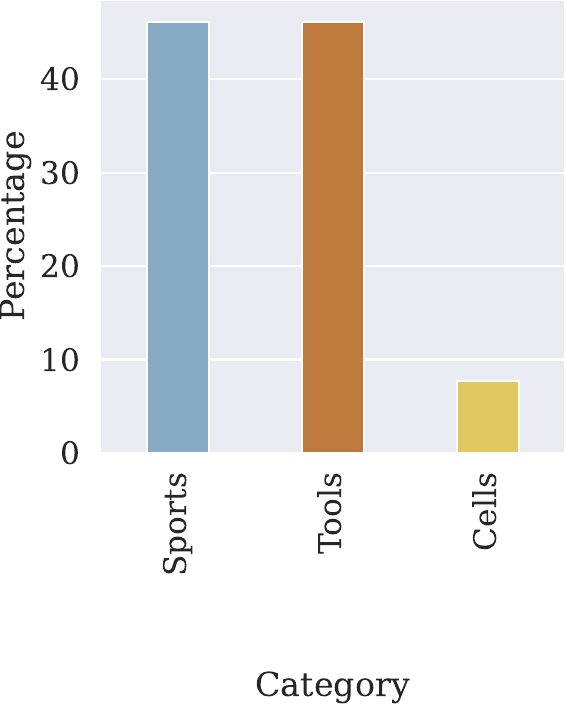}
          \caption{\AP}
          \label{fig:cat_ap}
      \end{subfigure}
      \begin{subfigure}{0.24\textwidth}
        \centering   
        \includegraphics[width=0.9\linewidth]{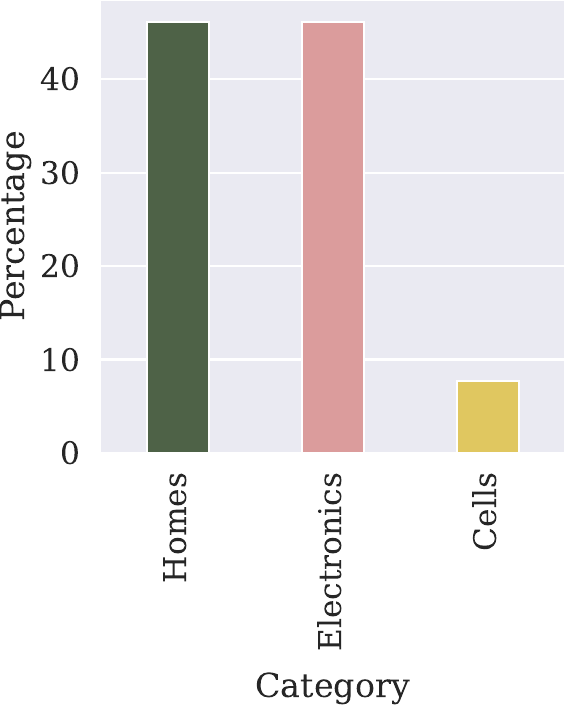}
          \caption{\AG}
          \label{fig:cat_ag}
      \end{subfigure}
    \caption{
    \label{fig:cat_dist}
    Distribution of Product Category
    }
\end{figure}
  
Figure \ref{fig:cat_dist} shows the distribution of product categories for \AVE, \IRP, \SA, \SR, \AP, and \AG. The diverse categories in \dataset significantly enhance the effectiveness of \method. Note that the data of \EM, \MPC, \PSI, and \QPR tasks lacks category information, thus their distributions are not shown.
%
\begin{figure}[htbp]
	\centering
	\begin{subfigure}{0.19\linewidth}
		\centering
		\includegraphics[width=0.9\linewidth]{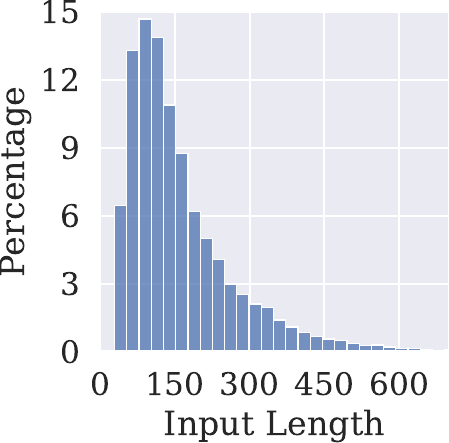}
		\caption{~\AVE}
	\end{subfigure}
	\begin{subfigure}{0.19\linewidth}
		\centering
		\includegraphics[width=0.9\linewidth]{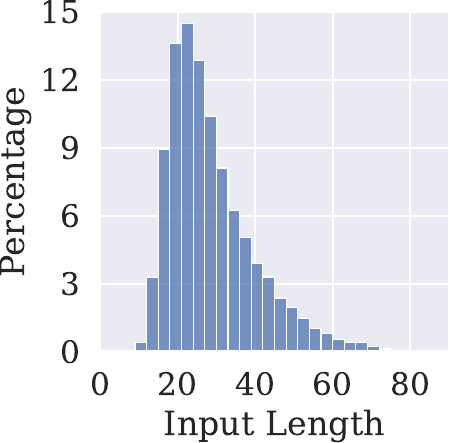}
		\caption{~\IRP}
	\end{subfigure}
  \begin{subfigure}{0.19\linewidth}
		\centering
		\includegraphics[width=0.9\linewidth]{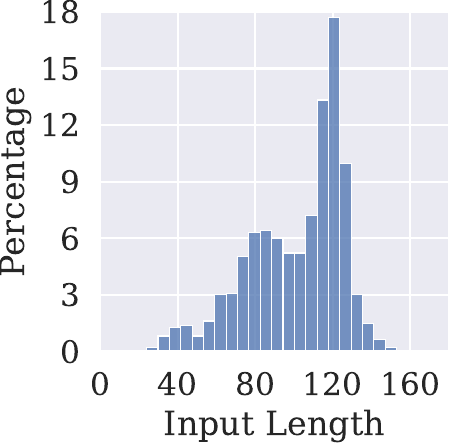}
		\caption{~\EM}
	\end{subfigure}
  \begin{subfigure}{0.19\linewidth}
		\centering
		\includegraphics[width=0.9\linewidth]{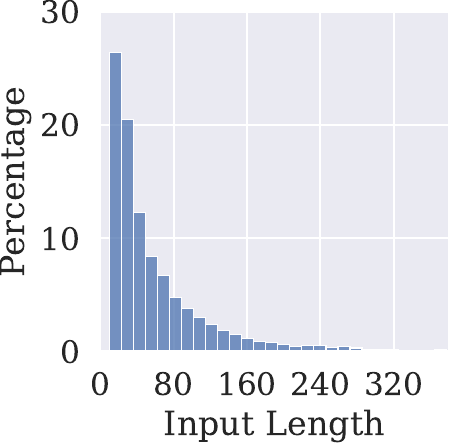}
		\caption{~\SA}
	\end{subfigure}
  \begin{subfigure}{0.19\linewidth}
		\centering
		\includegraphics[width=0.9\linewidth]{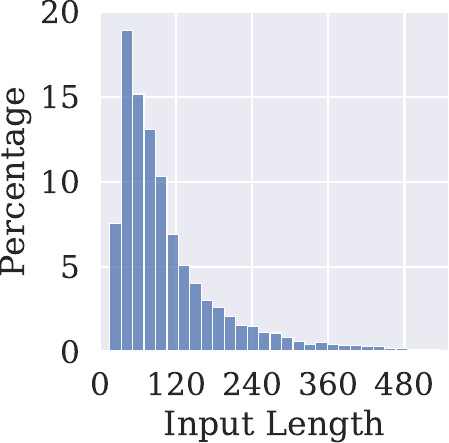}
		\caption{~\SR}
	\end{subfigure}
	\newline
	\vspace{3pt}
	\begin{subfigure}{0.19\linewidth}
		\centering
		\includegraphics[width=0.9\linewidth]{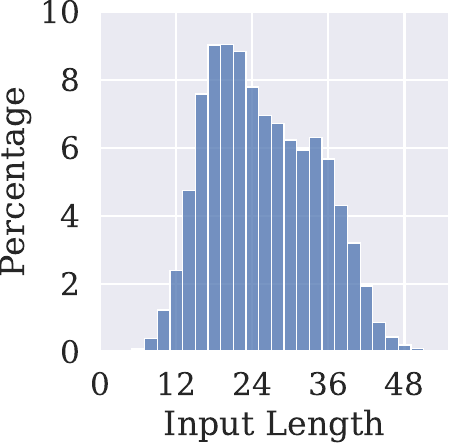}
		\caption{~\MPC}
	\end{subfigure}
  \begin{subfigure}{0.19\linewidth}
		\centering
		\includegraphics[width=0.9\linewidth]{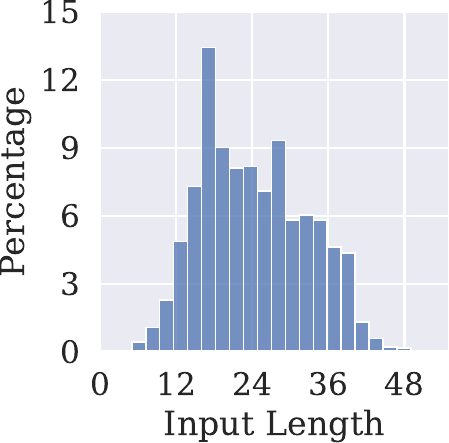}
		\caption{~\PSI}
	\end{subfigure}
  \begin{subfigure}{0.19\linewidth}
		\centering
		\includegraphics[width=0.9\linewidth]{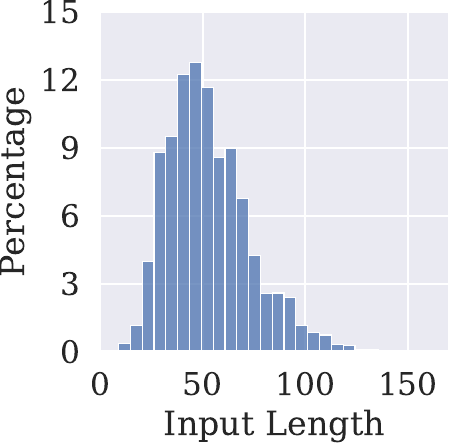}
		\caption{~\QPR}
	\end{subfigure}
  \begin{subfigure}{0.19\linewidth}
		\centering
		\includegraphics[width=0.9\linewidth]{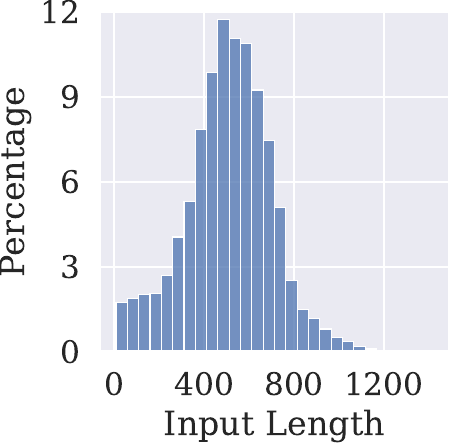}
		\caption{~\AP}
	\end{subfigure}
  \begin{subfigure}{0.19\linewidth}
		\centering
		\includegraphics[width=0.9\linewidth]{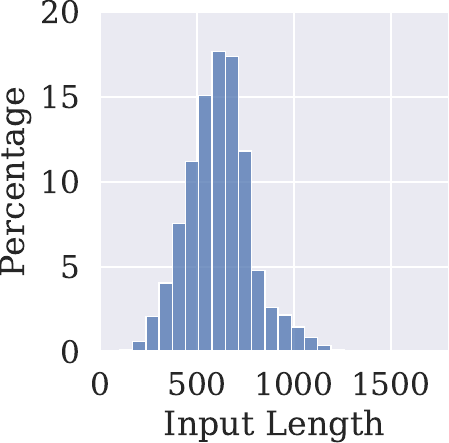}
		\caption{~\AG}
	\end{subfigure}
  \caption{
    \label{fig:input_len}
    Distribution of Input Length}
\end{figure}
Besides product categories, we also present the distributions of input lengths for each task, measured by word count, in Figure \ref{fig:input_len}. For better clarity, we exclude very long inputs (those representing at most 1\% of samples) in the \AVE, \IRP, \SA, and \SR tasks. 
%

Due to limited computing resources, especially in academic settings, the current \dataset dataset 
has at most 10k training samples per task to fit within the model training capacity.
However, scaling up \dataset is straightforward by following our data processing procedures detailed in Section \ref{sec:dataset} and Appendix \ref{sec:appendix:preprocessing}. For example, approximately 4.8 million high-quality samples could be further included in the \AVE task as outlined in Appendix \ref{sec:appendix:preprocessing:ave}.
To allow researchers and developers to customize \dataset to their specific needs, all data processing scripts are available at \href{https://github.com/ninglab/eCeLLM/tree/main/data_processing}{https://github.com/ninglab/eCeLLM/tree/main/data\_processing}. 

Table \ref{tbl:sample_num} provides the number of samples at different stages: in the raw data, after data filtering and quality checks, and after sampling. The ``After Filtering'' column shows the remaining data after applying quality checks. The ``After Sampling'' column shows the data included in \dataset. Detailed statistics for the ``After Sampling'' column, including training, validation, and testing samples for both in-domain and out-of-domain evaluation, are available in Table \ref{tbl:data_summary}.

Even after filtering and quality checks, a substantial number of samples remain available for each task, making \dataset easily extendable and scalable. We believe \dataset can significantly benefit e-commerce foundation model research in various applications.

\begin{table*}[!h]
  \caption{Number of Samples in Each Task}
  \centering
  \label{tbl:sample_num}
  \footnotesize
  \begin{threeparttable}
      \begin{tabular}{
        @{\hspace{6pt}}l@{\hspace{10pt}}
        @{\hspace{6pt}}r@{\hspace{6pt}}
        @{\hspace{6pt}}r@{\hspace{6pt}}
        @{\hspace{6pt}}r@{\hspace{6pt}}
      }
      \toprule
      Task & Raw Data & After Filtering & After Sampling \\
      \midrule
      \AVE & 15,177,901 & 4,767,579 & 13,000 \\
      \IRP & 10,868,853 & 7,705,363 & 13,000 \\
      \EM & 2,600 & 2,528 & 2,528 \\
      \SA & 800,000 & 617,200 & 13,000 \\
      \SR & 31,013,687 & 1,946,393 & 13,000 \\
      \MPC & 2,621,288 & 935,757 & 12,000 \\
      \PSI & 2,621,288 & 935,757 & 12,000 \\
      \QPR & 130,652 & 57,707 & 12,000 \\
      \AP & 220,541 & 184,814 & 13,000 \\
      \AG & 398,654 & 213,612 & 13,000 \\
      
      \bottomrule
      \end{tabular}
      \begin{tablenotes}[normal, flushleft]
        \begin{footnotesize}
          \item The ``After Sampling'' column indicates the data included in \dataset. 
        \end{footnotesize}
      \end{tablenotes}
  \end{threeparttable}
\end{table*}


\section{Instructions}
\label{sec:appendix:instruction}

Each task has 6 instructions. One of them is the human-written seed instruction, 
4 of them are GPT-4 generated diverse instructions, and the last one is the ``unseen" instruction.

\subsection{Instructions on Attribute Value Extraction (\AVE)}

\textbf{Seed Instruction}
{\emph{Given the title, description, feature, price, and brand of a product and a set of target attributes, extract the value of each target attribute from the product information. Output the extracted value and the corresponding source (e.g., title or feature) denoting where the value is extracted.}}

\textbf{Generated Instruction 1}
{\emph{Extract the value of the target attribute from the given product information and output it along with the corresponding source.}}

\textbf{Generated Instruction 2}
{\emph{Parse the product information to locate the target attribute, and then provide the extracted value of the target attribute and its source in the output, specifying None if the attribute is not present.}}

\textbf{Generated Instruction 3}
{\emph{First, identify the target attributes from the provided list. Then, scan the product title, description, feature, and brand to extract the values associated with each target attribute. Finally, create a list of dictionaries, each containing the extracted attribute, its corresponding value, and the source where it was found.}}

\textbf{Generated Instruction 4}
{\emph{Using the product's title, description, features, price, and brand, identify and retrieve the values associated with a specified set of target attributes. Output the extracted values along with their respective sources (e.g., title or feature) indicating where each value was found.}}

\textbf{Unseen Instruction}
{\emph{Retrieve the value associated with the target attribute from the product information and specify the source (e.g., title, description, feature, or title) where the value was found.}}

\subsection{Instructions on Product Relation Prediction (\IRP)}
\textbf{Seed Instruction}
{\emph{Given the title of two products, predict if the two products are similar, if the two products will be purchased or viewed together. Answer only from the options.}}

\textbf{Generated Instruction 1}
\emph{Analyze the titles of Product 1 and Product 2 to determine if they are similar, if they will be purchased or viewed together, and choose the corresponding option.}

\textbf{Generated Instruction 2}
\emph{Evaluate the titles of Product 1 and Product 2, then choose the option that best describes the relation between the two products.}

\textbf{Generated Instruction 3}
\emph{Evaluate the titles of Product 1 and Product 2 to assess their similarity and whether they are likely to be purchased or viewed together. Then, select the appropriate option.}

\textbf{Generated Instruction 4}
\emph{Predict whether two products are similar, whether two products are likely to be purchased or viewed together based on their titles. Choose your answer from the provided options.}

\textbf{Unseen Instruction}
\emph{Analyze the titles of Product 1 and Product 2 and select the option that indicates the relation of the two products.}

\subsection{Instructions on Product Matching (\EM)}
\textbf{Seed Instruction}
{\emph{Given the title, description, manufacturer, and price of two products, identify if they are the same product. Only output yes or no.}}

\textbf{Generated Instruction 1}
{\emph{Analyze the title, description, manufacturer, and price between the two products below and generate an output of yes if the two products are the same, otherwise respond with no.}}

\textbf{Generated Instruction 2}
{\emph{Check the details of the two products to see if they refer to the same product. Output only yes or no.}}

\textbf{Generated Instruction 3}
{\emph{Based on the product information, predict if the two products are identical or not. Output yes if they are identical or no otherwise.}}

\textbf{Generated Instruction 4}
{\emph{Compare the details of two given products to determine if they are identical. Output yes if they are identical or no otherwise.}}

\textbf{Unseen Instruction}
{\emph{Determine whether the two products are the same by comparing their title, description, manufacturer, and price, and provide a simple yes or no answer as the output.}}

\subsection{Instructions on Sentiment Analysis (\SA)}
\textbf{Seed Instruction}
{\emph{Given the user's review, identify the user's sentiment from the listed options. Answer using one of the options.}}

\textbf{Generated Instruction 1}
\emph{Assess the user's sentiment in the provided review and select the appropriate sentiment option from the list as the answer.}

\textbf{Generated Instruction 2}
\emph{Determine the sentiment expressed by the user in her review from the provided choices, and respond by selecting one of the available options.}

\textbf{Generated Instruction 3}
\emph{Carefully assess the user's review for any strong expressions of sentiment, either positive or negative. Based on your analysis, select the most fitting sentiment option from the provided list as output.}

\textbf{Generated Instruction 4}
\emph{Analyze the user's review text and determine the overall sentiment expressed, then choose the corresponding sentiment option from the provided list (A: very positive, B: positive, C: neutral, D: negative, E: very negative) based on the identified sentiment.}

\textbf{Unseen Instruction}
\emph{Analyze the user's review and determine the sentiment based on the listed options.}

\subsection{Instructions on Sequential Recommendation (\SR)}
\textbf{Seed Instruction}
{\emph{Given the products the user has purchased in history, rank the items in the listed options and output the item that the user is most likely to purchase next. Answer from one of the options.}}

\textbf{Generated Instruction 1}
\emph{Based on the user’s historical purchases, rank the items in options and predict the next product of the user’s interest from the provided options.}

\textbf{Generated Instruction 2}
\emph{Rank the items in options and predict the user's next purchase from the listed options by analyzing her historical purchases.}

\textbf{Generated Instruction 3}
\emph{The user's purchase history implies her preferences. Rank the items in the options based on the user’s preferences. Output the item that the user is most likely to purchase next from the options.}

\textbf{Generated Instruction 4}
\emph{Rank items in listed options based on the user’s purchase history to determine the item that the user is most likely to purchase next. Output the item with the highest likelihood of being the next purchase.}

\textbf{Unseen Instruction}
\emph{Estimate the user's intent based on the user's purchase history, and predict the next product that the user is most likely to purchase from the given options.}

\subsection{Instructions on Multi-class Product Classification (\MPC)}
\textbf{Seed Instruction}
{\emph{What is the relevance between the query and the product title below? Answer from one of the options.}}

\textbf{Generated Instruction 1}
\emph{Analyze the query and product title to determine the relevance between the query and product, and select the appropriate option from the provided options.}

\textbf{Generated Instruction 2}
\emph{Evaluate the relevance between the query and product title, and choose the most accurate option from the given options.}

\textbf{Generated Instruction 3}
\emph{Analyze the query and product title to assess the level of relevance between them, and then output the corresponding option that best describes this relevance.}

\textbf{Generated Instruction 4}
\emph{Determine the relevance between the query and the product title provided, and select your response from one of the available options.}

\textbf{Unseen Instruction}
\emph{Compare the query and the product title to determine if the product fully meets the query specifications. Choose the option that best describes the relevance between them.}

\subsection{Instructions on Product Substitute Identification (\PSI)}
{\bf{Seed Instruction}}
{\emph{Given a query and a product, identify if the product is somewhat relevant to the query. It fails to fulfill some aspects of the query but the product can be used as a functional substitute. Only output yes or no.}}

\textbf{Generated Instruction 1}
\emph{Answer yes if the product is a substitute for the query and no otherwise.}

\textbf{Generated Instruction 2}
\emph{Please respond with yes if the product is a suitable substitute for the query, and no if it is not.}

\textbf{Generated Instruction 3}
\emph{Check if a product can function as a substitute for a given query, even if it doesn't fully meet all requirements. Output yes if it can or no otherwise.}

\textbf{Generated Instruction 4}
\emph{Assess the relevance of a product to a given query by determining if it can function as a substitute, despite not fully meeting certain aspects of the query. Provide a binary output of yes or no based on this evaluation.}

\textbf{Unseen Instruction}
\emph{Assess whether the product is a substitute for the query and provide a yes or no response.}

\subsection{Instructions on Query Product Ranking (\QPR)}
\textbf{Seed Instruction}
{\emph{Given a query and a list of products denoted as A, B, C, ... with their titles, rank the products according to their relevance to the query. Output only a ranked list in which the most relevant product is at the top of the list.}}

\textbf{Generated Instruction 1}
\emph{Evaluate each product title in the given list, assess its relevance to the given query, and then arrange the products in descending order of relevance, with the most relevant product at the top of the ranked list.}

\textbf{Generated Instruction 2}
\emph{Rank the products A, B, C, ... based on their relevance to the provided query, and produce a ranked list with the most relevant product positioned at the top of the list.}

\textbf{Generated Instruction 3}
\emph{Analyze the query and each product title. Sort the products in descending order based on their relevance to the query. The most relevant product should be at the top of the list, and output the ranked list.}

\textbf{Generated Instruction 4}
\emph{Evaluate the relevance of each product title in the input to the given query, and then sort the products in descending order of relevance, placing the most relevant product at the top of the ranked list.}

\textbf{Unseen Instruction}
\emph{Evaluate the query against each product's title, determine the relevance between the query and the product, and organize the products in descending order of relevance, ensuring that the product with the highest relevance is positioned at the top of the list.}

\subsection{Instructions on Answerability Prediction (\AP)}
\textbf{Seed Instruction}
{\emph{Given a question and the related document, predict if the question is answerable based on the information provided in the document. Output only yes or no.}}

\textbf{Generated Instruction 1}
\emph{Evaluate the answerability of a question by analyzing the related document, outputting yes if the document contains information addressing the question, and no otherwise.}

\textbf{Generated Instruction 2}
\emph{Analyze a question and its supporting document. Predicting answerability based on the information provided in the document. Output yes if the document contains relevant information to answer the question, otherwise output no.}

\textbf{Generated Instruction 3}
\emph{Given a question and its related document, determine if the question is answerable by analyzing the information in the document. Output yes if the document addresses the question, or no otherwise.}

\textbf{Generated Instruction 4}
\emph{Output yes if the supporting document can answer the given question. Otherwise, output no.}

\textbf{Unseen Instruction}
\emph{Predict whether it is possible to answer the given question using the supporting document, and output a yes or no response.}

\subsection{Instructions on Answer Generation (\AG)}
\textbf{Seed Instruction}
{\emph{Given a question and the related document, and generate the answer to the question based on the information provided in the document.}}

\textbf{Generated Instruction 1}
\emph{Generate an answer to the question by utilizing the information contained in the document.}

\textbf{Generated Instruction 2}
\emph{Extract information from the supporting document to answer the given question.}

\textbf{Generated Instruction 3}
\emph{Answer the given question using the supporting document.}

\textbf{Generated Instruction 4}
\emph{Answer the given question by extracting information from the supporting document.}

\textbf{Unseen Instruction}
\emph{Utilize the information provided in the supporting document to generate an answer to the given question.}

\section{Training and Evaluation Details}
\label{sec:appendix:training}

The training details of general-purpose and e-commerce baseline LLMs are shown in Table~\ref{tbl:training_llms_baseline}. 
The general-purpose LLMs undergo the 1-shot evaluation, which measures the sample with one in-context example and one test case.
For each task, the 1-shot evaluation dataset is composed of all 1K test samples and the same amount of training samples (except for {\EM}, 
which uses all 253 test samples and 253 training samples) from its own test and training sets.
The e-commerce LLM undergoes both 1-shot and  0-shot evaluation.

\begin{table*}[h]
\footnotesize
  \caption{Training Details of General-purpose and \mbox{E-commerce} Baseline LLMs}
  \centering
  \label{tbl:training_llms_baseline}
  \begin{threeparttable}
      \begin{tabular}{
	@{\hspace{2pt}}p{0.25\textwidth}@{\hspace{2pt}}
    @{\hspace{2pt}}p{0.55\textwidth}@{\hspace{2pt}}
    @{\hspace{2pt}}p{0.1\textwidth}@{\hspace{0pt}}
    }
    \toprule
    General-purpose LLMs & URL & Accessibility\\	
	\midrule
    GPT-4 Turbo & \href{https://platform.openai.com/}{https://platform.openai.com/} & API
    \\
    Gemini Pro & \href{https://ai.google.dev/tutorials/python\_quickstart}{https://ai.google.dev/tutorials/python\_quickstart} & API
    \\
    Claude 2.1 & \href{https://docs.anthropic.com/claude/reference/getting-started-with-the-api}{https://docs.anthropic.com/claude/reference/getting-started-with-the-api} & API
    \\
    Llama-2 13B-chat & \href{https://huggingface.co/meta-llama/Llama-2-13b-chat-hf}{https://huggingface.co/meta-llama/Llama-2-13b-chat-hf} & Checkpoint
    \\
    Mistral-7B-Instruct-v0.2 & \href{https://huggingface.co/mistralai/Mistral-7B-Instruct-v0.2}{https://huggingface.co/mistralai/Mistral-7B-Instruct-v0.2} & Checkpoint
    \\
    \midrule
    EcomGPT & \href{https://github.com/Alibaba-NLP/EcomGPT}{https://github.com/Alibaba-NLP/EcomGPT} & Checkpoint \\
    \bottomrule
    \end{tabular}
  \vspace{-10pt}
  \end{threeparttable}
\end{table*}



\begin{table*}[!h]
\footnotesize
  \caption{Training Details of SoTA Task-specific Models}
  \centering
  \label{tbl:baseline_training}
  \begin{threeparttable}
      \begin{tabular}{
	@{\hspace{0pt}}p{0.06\textwidth}@{\hspace{4pt}}
	@{\hspace{4pt}}p{0.3\textwidth}@{\hspace{0pt}}
    @{\hspace{4pt}}p{0.36\textwidth}@{\hspace{4pt}}
    @{\hspace{4pt}}p{0.1\textwidth}@{\hspace{0pt}}
    @{\hspace{4pt}}p{0.12\textwidth}@{\hspace{0pt}}
    }
    \toprule
	Tasks & Baseline Models & Model Source & Mode & Parameters\\	
	\midrule
	\multirow{3.5}{*}{\AVE}  
	& SUOpenTag~\cite{xu-etal-2019-scaling} & \href{https://github.com/hackerxiaobai/OpenTag_2019}{https://github.com/hackerxiaobai/OpenTag\_2019} & training & epoch: 100 \\
    \cmidrule(lr){2-5}
	& AVEQA~\cite{AVEQA} & \href{https://github.com/Zinc-30/aveqa}{https://github.com/Zinc-30/aveqa} & training & epoch: 89; batch size: 16 \\
	\midrule
	\multirow{3.5}{*}{\IRP} 
	& RGCN~\cite{schlichtkrull2018modeling} & \href{https://github.com/JinheonBaek/RGCN}{https://github.com/JinheonBaek/RGCN} & training & epochs: 500K \\
    \cmidrule(lr){2-5}
    & DeBERTaV3~\cite{he2021debertav3} & \href{https://huggingface.co/microsoft/deberta-v3-base}{https://huggingface.co/microsoft/deberta-v3-base} & fine-tuning & epoch: 3; ~~~~batch size: 8 \\ 
    \midrule
	\multirow{6.5}{*}{\SA} 
	& \multirow{1}{*}{BERTweet~\cite{nguyen2020bertweet}} & \href{https://huggingface.co/finiteautomata/bertweet-base-sentiment-analysis}{https://huggingface.co/finiteautomata/bertweet-base-sentiment-analysis} & training  & epoch: 3; ~~~~batch size: 8 \\
    \cmidrule(lr){2-5}
    & DeBERTaV3~\cite{he2021debertav3} & \href{https://huggingface.co/microsoft/deberta-v3-base}{https://huggingface.co/microsoft/deberta-v3-base} & fine-tuning & epoch: 3; ~~~~batch size: 8 \\ 
    \cmidrule(lr){2-5}
	& P5~\cite{geng2022recommendation} & \href{https://github.com/jeykigung/P5}{https://github.com/jeykigung/P5} & evaluate on checkpoint & checkpoint: toy base \\
    \midrule
	\multirow{4.5}{*}{\SR}
	& gSASRec~\cite{petrov2023gsasrec} & \href{https://github.com/asash/gSASRec-pytorch}{https://github.com/asash/gSASRec-pytorch} & training & sequence length: 50 \\
    \cmidrule(lr){2-5}
    & Recformer~\cite{li2023text} & \href{https://github.com/AaronHeee/RecFormer}{https://github.com/AaronHeee/RecFormer} & fine-tuning & epoch: 16; ~~~~batch size 8\\
    \midrule
	\multirow{2}{*}{\AG} 
	& GPT-4 Turbo~\cite{GPT4} & \href{https://platform.openai.com/}{https://platform.openai.com/} & 1-shot ~~~learning & prompt template in~\ref{sec:appendix:Template:others} \\
    \midrule
    \multirow{5}{0.06\textwidth}{\EM, \MPC, \PSI, \QPR, \AP} 
    & BERT~\cite{devlin2018bert} & \href{https://huggingface.co/bert-base-multilingual-cased}{https://huggingface.co/bert-base-multilingual-cased} & fine-tuning & epoch: 3; ~~~~batch size: 8 \\
    \cmidrule(lr){2-5}
    & DeBERTaV3~\cite{he2021debertav3} & \href{https://huggingface.co/microsoft/deberta-v3-base}{https://huggingface.co/microsoft/deberta-v3-base} & fine-tuning & epoch: 3; ~~~~batch size: 8 \\
    \bottomrule
    \end{tabular}
    \begin{tablenotes}[normal,flushleft]
    \begin{footnotesize}
    \item In this table, ``training" means training from scratch, ``evaluate on checkpoint" means that we evaluate using the checkpoint provided by the link, ``fine-tuning" means that we fine-tune the checkpoint provided by the link on specific tasks, and ``1-shot learning" indicates that we directly use the model checkpoint from the link, and prompt the model with one in-context example.
    \par
    \end{footnotesize}
    \end{tablenotes}
  \vspace{-10pt}
  \end{threeparttable}
\end{table*}


The training details of SoTA task-specific models are presented in Table \ref{tbl:baseline_training}. 
For each task, we train and test on the SoTA task-specific models of each task using its own training, validation, and IND test sets (i.e., task-specific).
For the tasks with the OOD test set, we save the trained model and test them on the OOD test set.
%
%
For SUOpenTag~\cite{xu-etal-2019-scaling} and AVEQA~\cite{AVEQA}, 
we evaluate the \AVE task according to the equation~\ref{eqn:Fave}.
%
Note that we do not consider MAVEQA~\cite{yang2022mave} as the baseline of \AVE, as it has demonstrated similar performance to AVEQA. 
For RGCN~\cite{schlichtkrull2018modeling}, we construct a product-product graph, in which the nodes are products, and the edges are specified by the relations between products (e.g., similar products). We initialize the node embeddings in the RGCN using the embeddings of product titles generated from BERT~\cite{devlin2018bert}.
The BERT, DeBERTa, and BERTweet baselines are implemented through \mbox{sentence-transformer}~\cite{reimers-2019-sentence-bert}.
For P5~\cite{geng2022recommendation}, we conduct zero-shot evaluations on beauty, sports, and toys base checkpoints, and report the best result (toy base).
Similar to the RGCN, we use the BERT-generated embeddings of product titles to initialize item embeddings in the gSASRec.
%
For both gSASRec and Recformer~\cite{li2023text}, 
we evaluate the results of the next product of interest within the candidate list as detailed in Appendix~\ref{sec:appendix:preprocessing:sr}

\section{Prompt Templates}
\label{sec:appendix:Template}
The following prompts are used for evaluating general-purpose and e-commerce LLMs.
For Claude 2.1, the prompt begins with \{HUMAN PROMPT\} and ends with \{AI PROMPT\}.
For Mistral-7B Instruct-v0.2, the prompt is wrapped with ``[INST]".
\subsection{Prompt Templates for Zero-shot Evaluation}
\label{sec:appendix:Template:ecomgpt}
\paragraph{Prompt with Options}
\begin{itemize}
    \item System prompt: Below is an instruction that describes a task. Write a response that appropriately completes the request.
    \item Instruction: \{instruction\}
        \begin{itemize}
            \item input: \{input\}
            \item options: \{options\}
            \item response: 
        \end{itemize}
\end{itemize}
 
\paragraph{Prompt without Options}
\begin{itemize}
    \item System prompt: Below is an instruction that describes a task. Write a response that appropriately completes the request.
    \item Instruction: \{instruction\}
        \begin{itemize}
            \item input: \{input\}
            \item response: 
        \end{itemize}
\end{itemize}

\subsection{Prompt Templates for One-shot Evaluation}
\label{sec:appendix:Template:others}
\paragraph{Prompt with Options}
\begin{itemize}
    \item System prompt: Below is an instruction that describes a task, paired with an example that provides further context for the task.
    \item Instruction: \{instruction\}
    \item Example:
        \begin{itemize}
            \item input: \{example.input\}
            \item options: \{example.options\}
            \item response: \{example.response\}
        \end{itemize}
    \item Now write a response that appropriately completes the following example.
        \begin{itemize}
            \item input: \{input\}
            \item options: \{options\}
            \item response: 
        \end{itemize}
\end{itemize}
 
\paragraph{Prompt without Options}
\begin{itemize}
    \item System prompt: Below is an instruction that describes a task, paired with an example that provides further context for the task.
    \item Instruction: \{instruction\}
    \item Example:
        \begin{itemize}
            \item input: \{example.input\}
            \item response: \{example.response\}
        \end{itemize}
    \item Now write a response that appropriately completes the following example.
        \begin{itemize}
            \item input: \{input\}
            \item response: 
        \end{itemize}
\end{itemize}

\section{Analysis on Base Models}
\label{sec:appendix:base_model}


\begin{table*}[!h]
  \caption{Performance of Various Base Models in IND Evaluation}
  \centering
  \label{tbl:base_indomain}
  \footnotesize
  \begin{threeparttable}
      \begin{tabular}{
        @{\hspace{0pt}}l@{\hspace{3pt}}
        @{\hspace{3pt}}c@{\hspace{2pt}}
	  @{\hspace{3pt}}c@{\hspace{3pt}}
	  @{\hspace{4pt}}c@{\hspace{4pt}}
	  @{\hspace{3pt}}c@{\hspace{4pt}}
	  @{\hspace{3pt}}c@{\hspace{3pt}}
	  @{\hspace{3pt}}c@{\hspace{3pt}}
        @{\hspace{3pt}}c@{\hspace{2pt}}
        @{\hspace{1pt}}c@{\hspace{1pt}}
        @{\hspace{3pt}}c@{\hspace{4pt}}
        @{\hspace{6pt}}c@{\hspace{6pt}}
        @{\hspace{6pt}}c@{\hspace{1pt}}
      }
      \toprule
      \multirow{2.5}{*}{Model} & \multirow{2.5}{*}{Base Model} & 
      \AVE & \IRP & \EM & \SA & \SR & \MPC & \PSI & \QPR & \AP & \AG \\ 
      \cmidrule(lr){3-12}
      && F1* & Macro F1 & F1 & Macro F1 & HR@1 & Accuracy & F1 & NDCG & F1 & F$_{\text{BERT}}$ \\ 
      \midrule
      \multirow{2}{*}{\methodL} & Flan-T5 XXL & 0.447 & 0.522 & \textbf{0.995} & 0.628 & 0.512 & {0.680} & {0.376} & \textbf{0.885} & 0.836 & \textbf{0.844} \\ 
      & Llama-2 13B-chat & \textbf{0.582} & \textbf{0.611} & \textbf{0.995} & \textbf{0.648} & \textbf{0.526} & \textbf{0.684} & \textbf{0.501} & 0.870 & \textbf{0.851} & 0.841 \\ 
      \midrule
      \multirow{2}{*}{\methodB}
      & Llama-2 7B-chat & 0.577 & \textbf{0.562} & \textbf{0.995} & 0.613 & 0.503 & {0.695} & \textbf{0.444} & 0.864 & \textbf{0.859} & {0.839} \\ 
      & Mistral-7B Instruct-v0.2 & \textbf{0.662} & 0.558 & \textbf{0.995} & \textbf{0.639} & \textbf{0.542} & \textbf{0.696} & 0.305 & \textbf{0.876} & 0.846 & \textbf{0.842} \\ 
      \midrule
      \multirow{2}{*}{\methodS}
      & Flan-T5 XL & 0.376 & 0.511 & \textbf{0.995} & \textbf{0.648} & 0.463 & \textbf{0.663} & 0.042 & 0.868 & 0.827 & \textbf{0.843} \\ 
      & Phi-2 & \textbf{0.509} & \textbf{0.518} & 0.991 & 0.596 & \textbf{0.479} & 0.650 & \textbf{0.392} & \textbf{0.870} & \textbf{0.846} & 0.842 \\ 

      \bottomrule
      \end{tabular}
      \begin{tablenotes}[normal, flushleft]
      \begin{footnotesize}
      \item
      In this table, ``Base Model" presents the base models used in \method models.
      On each task, the best performance of \methodL, \methodB, and \methodS when using different base models is in \textbf{bold}.
      \par
      \end{footnotesize}
      \end{tablenotes}
  \vspace{-10pt}
  \end{threeparttable}
\end{table*}


We compare the 6 base models considered for the series of \method models (i.e., \methodL, \methodB, and \methodS) 
and show the performance comparison in Table~\ref{tbl:base_indomain}.
From Table~\ref{tbl:base_indomain}, 
we observe that Llama-2 13B-chat is the \mbox{best-performing} base model for \methodL.
The instruction-tuned Llama-2 13B-chat model demonstrates considerable improvement compared to instruction-tuned Flan-T5 XXL on 7 out of the 10 tasks.
We also observe that Mistral-7B Instruct-v0.2 and Phi-2 are the \mbox{best-performing} base models for \methodB and \methodS, respectively.
Particularly, 
with instruction tuning, Mistral-7B Instruct-v0.2 achieves a notable average improvement of 2.2\% over Llama-2 7B-chat across the 10 tasks.
Similarly, 
Instruction-tuned Phi-2 also outperforms Flan-T5 XL on 6 out of the 10 tasks and achieves similar performance with Flan-T5 XL on the rest 4 tasks.
The variations of trainable size and focused aspect contribute to the distinct inherent capabilities of the base models, which play a crucial role in adapting LLMs to e-commerce scenarios.

\section{Complete Experimental Results}
\label{sec:appendix:performance}


The complete results for both IND and OOD evaluations for tasks 
\textbf{(1)} attribute value extraction (\AVE), 
\textbf{(2)} product relation prediction (\IRP), 
\textbf{(3)} product matching (\EM), 
\textbf{(4)} sentiment analysis (\SA), 
\textbf{(5)} sequential recommendation (\SR), 
\textbf{(6)} multi-class product classification (\MPC), 
\textbf{(7)} product substitute identification (\PSI), 
\textbf{(8)} query product ranking (\QPR),   
\textbf{(9)} answerability prediction (\AP), and 
\textbf{(10)} answer generation (\AG)
are presented in Table~\ref{tbl:ave}, ~\ref{tbl:irp}, ~\ref{tbl:em}, ~\ref{tbl:sa}, ~\ref{tbl:sr_qpr}, ~\ref{tbl:mpc}, ~\ref{tbl:psi}, ~\ref{tbl:sr_qpr}, 
~\ref{tbl:ap},  and~\ref{tbl:ag}, respectively.  
Overall, these tables show that \method models fine-tuned on {\dataset} outperform the general-purpose LLMs and SoTA task-specific models in the IND test.
Meanwhile, the \method exhibits good generalizability to the OOD data.
Because of the high-quality \dataset, \method achieves remarkable performance
with different base models. 
%
Note that \#failed in tables represents the number of failure cases for which we cannot extract meaningful results from the model output.
These failure cases are counted as wrong predictions when calculating the metrics. NPR refers to the negative prediction rate.

Regarding the hallucination issue, we observe very limited hallucination from \method. As shown in the \#failed column of Table~\ref{tbl:ave} and ~\ref{tbl:sr_qpr}, in \AVE and \SR tasks, \method could slightly suffer from hallucination and does not output results in the desired format for several testing samples. For the other tasks except for the \AG task, as shown in Table~\ref{tbl:irp}-\ref{tbl:ap}, \method is robust to hallucination and could always output options in a desired format. For the \AG task, we randomly sample 50 answers generated by \method and do not observe hallucination with manual checks.

\begin{table*}[!h]
  \caption{Performance on \AVE}
  \centering
	\footnotesize
  \label{tbl:ave}
  \begin{threeparttable}
      \begin{tabular}{
        @{\hspace{0pt}}l@{\hspace{2pt}} 
        @{\hspace{2pt}}l@{\hspace{2pt}} 
        @{\hspace{2pt}}l@{\hspace{2pt}} 
	@{\hspace{2pt}}c@{\hspace{5pt}} 
	@{\hspace{5pt}}c@{\hspace{5pt}} 
	@{\hspace{5pt}}c@{\hspace{2pt}} 
	@{\hspace{2pt}}r@{\hspace{2pt}} 
	 @{\hspace{2pt}}c@{\hspace{2pt}} 
 	  @{\hspace{2pt}}c@{\hspace{5pt}} 
	  @{\hspace{5pt}}c@{\hspace{5pt}} 
	  @{\hspace{5pt}}c@{\hspace{2pt}} 
	  @{\hspace{2pt}}r@{\hspace{0pt}} 
      }
      \toprule
        \multicolumn{3}{c}{\multirow{2.5}{*}{Model}} & \multicolumn{4}{c}{IND} && \multicolumn{4}{c}{OOD} \\
      \cmidrule(lr){4-7}       \cmidrule(lr){9-12}
       & & & Recall* & Precision* & F1* & \#failed  & & Recall* & Precision* & F1* & \#failed\\
      \midrule
     \multicolumn{2}{l}{\multirow{5}{*}{General-purpose LLMs}}
         & GPT-4 Turbo & 0.422 & 0.598 & 0.495 & 6 & & 0.317 & 0.529 & 0.397 & 1 \\
       && Gemini Pro & 0.318 & 0.523 & 0.396 & 4 & & 0.203 & 0.426 & 0.275 & 6 \\
       && Claude 2.1 & 0.310 & 0.494 & 0.381 & 59 & & 0.312 & 0.600 & 0.410 & 66 \\
       && Llama-2 13B-chat & 0.002 & 0.002 & 0.002 & 0 & & 0.000 & 0.000 & 0.000 & 0 \\
       && Mistral-7B-Instruct-v0.2 & 0.321 & 0.435 & 0.369 & 69 & & 0.217 & 0.337 & 0.264 & 52\\
      \midrule
     \multicolumn{2}{l}{{E-commerce LLM}}
       & EcomGPT & 0.000 & 0.000 & 0.000 & 905 & & 0.001 & 0.042 & 0.001 & 869\\
      \midrule
      \multicolumn{2}{l}{\multirow{2}{*}{SoTA task-specific model}}
        & SUOpenTag & 0.603 & 0.500 & 0.546 & 0 & & 0.124 & 0.173 & 0.144 & 0 \\
      && AVEQA & 0.425 & 0.491 & 0.456 & 0 & & {0.283} & {0.257} & {0.269} & 0\\
      \midrule
      \multirow{12}{*}{\method}
      & \multirow{6}{*}{Task-specific} 
      & Flan-T5 XXL & 0.298 & 0.519 & 0.378 & 7 
      && 0.362 & \textbf{0.701} & 0.477 & 0 
      \\
      && Llama-2 13B-chat & 0.544 & 0.666 & 0.599 & 3 & & \textbf{0.448} & {0.613} & \textbf{0.518} & 2 \\
      & & Llama-2 7B-chat & 0.531 & 0.660 & 0.588 & 1 &&
      0.323 & 0.499 & 0.392 & 0\\
      & & Mistral-7B Instruct-v0.2 & \textbf{0.720} & \textbf{0.799} & \textbf{0.757} & 5 & & 0.374 & 0.544 & 0.443 & 0 \\
      & & Flan-T5 XL & 0.258 & 0.449 & 0.328 & 7 && 0.276 & 0.538 & 0.365 & 3\\
      & & Phi-2 & 0.304 & 0.570 & 0.397 & 265 && 0.288 & 0.488 & 0.362 & 0 \\
      \cmidrule(lr){2-12}
      & \multirow{6}{*}{Generalist} 
      & Flan-T5 XXL & 0.353 & 0.611 & 0.447 & 2 & & {0.360} & {0.699} & {0.476} & 0 \\
      & & Llama-2 13B-chat & 0.530 & 0.646 & 0.582 & 1 & & 0.276 & 0.425 & 0.335 & 0 \\
      & & Llama-2 7B-chat & 0.514 & 0.641 & 0.571 & 4 & & 0.236 & 0.392 & 0.294 & 0 \\
      & & Mistral-7B Instruct-v0.2 & {0.612} & {0.722} & {0.662} & 0 & & 0.304 & 0.463 & 0.367 & 0 \\
      & & Flan-T5 XL & 0.297 & 0.514 & 0.376 & 1 & & 0.267 & 0.518 & 0.352 & 1 \\
      & & Phi-2 & 0.455 & 0.578 & 0.509 & 0 & & 0.237 & 0.417 & 0.302 & 0 \\
      \bottomrule
      \end{tabular}
      \begin{tablenotes}[normal, flushleft]
      \begin{footnotesize}
      \item
      In this table, ``IND" and ``OOD" indicates in-domain evaluation and out-of-domain evaluation, respectively;
      ``Task-specific" indicates that the \method models are tuned on individual tasks;
      ``Generalist" represents tuning \method models using all tasks together.
      Recall*, Precision* and F1* are defined as equation~\ref{eqn:Fave} in Appendix~\ref{sec:appendix:preprocessing:ave}, and
      \#failed refers to the number of failure cases that we cannot extract meaningful results from the model output.
      On each task, the best performance is in \textbf{bold}.
      \par
      \end{footnotesize}
      \end{tablenotes}
  \end{threeparttable}
\end{table*}


\begin{table*}[!h]
  \caption{Performance on \IRP}
  \centering
\footnotesize
  \label{tbl:irp}
  \begin{threeparttable}
      \begin{tabular}{
        @{\hspace{0pt}}l@{\hspace{2pt}}
        @{\hspace{-15pt}}l@{\hspace{2pt}}
        @{\hspace{2pt}}l@{\hspace{2pt}}
	@{\hspace{2pt}}c@{\hspace{2pt}}
    @{\hspace{2pt}}c@{\hspace{2pt}}
	@{\hspace{2pt}}c@{\hspace{2pt}}
	@{\hspace{2pt}}c@{\hspace{2pt}}
	@{\hspace{-2pt}}r@{\hspace{2pt}}
	 @{\hspace{2pt}}c@{\hspace{2pt}}
 	  @{\hspace{2pt}}c@{\hspace{2pt}}
        @{\hspace{2pt}}c@{\hspace{2pt}}
	  @{\hspace{2pt}}c@{\hspace{2pt}}
	  @{\hspace{2pt}}c@{\hspace{2pt}}
	  @{\hspace{-2pt}}r@{\hspace{0pt}}
      }
      \toprule
      \multicolumn{3}{c}{\multirow{2.5}{*}{Model}} & \multicolumn{5}{c}{IND} && \multicolumn{5}{c}{OOD} \\
      \cmidrule(lr){4-8}       \cmidrule(lr){10-14}
       & & & Accuracy & M-Rec & M-Pre & M-F1 & \#failed  & & Accuracy & M-Rec & M-Pre & M-F1 & \#failed\\
      \midrule
     \multicolumn{2}{l}{\multirow{5}{*}{General-purpose LLMs}}
        & GPT-4 Turbo & 0.384 & {0.487} & {0.381} & 0.326 & 0 && {0.488} & {0.496} & 0.392 & {0.392} & 0 \\
       && Gemini Pro & 0.128 & 0.385 & 0.352 & 0.136 & 1 && 0.147 & 0.359 & 0.390 & 0.123 & 0 \\
       && Claude 2.1 & {0.508} & 0.347 & 0.344 & 0.275 & 10 && 0.362 & 0.394 & {0.400} & 0.277 & 4 \\
       && Llama-2 13B-chat & 0.473 & 0.333 & 0.333 & {0.333} & 0 && 0.419 & 0.338 & 0.339 & 0.324 & 0 \\
       && Mistral-7B-Instruct-v0.2 & 0.442 & 0.323 & 0.325 & 0.324 & 0 && 0.422 & 0.338 & 0.351 & 0.327 & 0 \\
       \midrule 
       \multicolumn{2}{l}{{E-commerce LLM}}
       & EcomGPT & 0.147 & 0.101 & 0.101 & 0.091 & 444 && 0.125 & 0.125 & 0.092 & 0.096 & 455 \\
      \midrule
      \multicolumn{2}{l}{\multirow{2}{*}{SoTA task-specific model}}
      & DeBERTaV3 & {0.762} & 0.575 & 0.620 & {0.588} & 0 && 0.658 & 0.514 & 0.570 & 0.507 & 0 \\
      && RGCN & 0.615  & \textbf{0.665} & {0.637}& 0.506 & 0 && 0.576 & 0.373 & 0.372 & 0.356 & 0 \\
      \midrule
      \multirow{11}{*}{\method}
      & \multirow{6}{*}{Task-specific}
      & Flan-T5 XXL & 0.754 & 0.516 & 0.511 & 0.508 & 0 
      && 0.663 & 0.506  &0.468 & 0.466 & 0\\
      && Llama-2 13B-chat & 0.769 & 0.530 & 0.517 & 0.521 & 0 && 0.690 & 0.520 & 0.472 & 0.483 & 0 \\
      & & Llama-2 7B-chat & 0.774 & 0.541 & 0.628 & 0.537 & 0
      & & 0.695 & 0.526 & 0.803 & 0.498 & 0\\
      & & Mistral-7B Instruct-v0.2 & {0.782} & {0.547} & \textbf{0.689} & {0.543} & 0 && {0.711} & {0.532} & \textbf{0.808} & {0.502} & 0 \\
      & & Flan-T5 XL & 0.704 & 0.467 & 0.496 & 0.460 & 0 
      && 0.592 & 0.471 & 0.625 & 0.427 & 0 \\
      & & Phi-2 & 0.584 & 0.372 & 0.379 & 0.348 & 0 && 0.406 & 0.349 & 0.334 & 0.251 & 0 \\
      \cmidrule(lr){2-14}
      & \multirow{6}{*}{Generalist} 
      & Flan-T5 XXL & 0.769 & 0.531 & 0.517 & 0.522 & 0 && 0.703 & 0.533 & {0.648} & 0.499 & 0 \\
      & & Llama-2 13B-chat & 0.775 & {0.599} & 0.635 & \textbf{0.611} & 0 && \textbf{0.726} & \textbf{0.564} & 0.611 & \textbf{0.558} & 0 \\
      
      & & Llama-2 7B-chat & \textbf{0.797} & 0.586 & {0.661} & 0.595 & 0 && 0.703 & 0.533 & {0.648} & 0.499 & 0 \\
      & & Mistral-7B Instruct-v0.2 & 0.788 & 0.555 & 0.644 & 0.558 & 0 && 0.707 & 0.537 & 0.596 & 0.502 & 0 \\
      
      & & Flan-T5 XL & 0.757 & 0.517 & 0.515 & 0.511 & 0 && 0.678 & 0.521 & 0.587 & 0.489 & 0 \\
      & & Phi-2 & 0.747 & 0.524 & 0.552 & 0.518 & 0 && 0.710 & 0.541 & 0.611 & 0.520 & 0 \\
      \bottomrule
      \end{tabular}
      \begin{tablenotes}[normal, flushleft]
      \begin{footnotesize}
      \item
      In this table, ``IND" and ``OOD" indicates in-domain evaluation and out-of-domain evaluation, respectively;
      ``Task-specific" indicates that the \method models are tuned on individual tasks;
      ``Generalist" represents tuning \method models using all tasks together;
      \#failed refers to the number of failure cases that we cannot extract meaningful results from the model output.
      The metric ``M-Rec", ``M-Pre", and ``M-F1" represents macro recall, macro precision, and macro F1, respectively.
      On each task, the best performance is in \textbf{bold}.
      \par
      \end{footnotesize}
      \end{tablenotes}
  \vspace{-10pt}
  \end{threeparttable}
\end{table*}


\begin{table*}[!h]
  \caption{Performance on \EM}
  \centering
  \label{tbl:em}
  \footnotesize
  \begin{threeparttable}
      \begin{tabular}{
        @{\hspace{0pt}}l@{\hspace{2pt}}
        @{\hspace{2pt}}l@{\hspace{2pt}}
        @{\hspace{2pt}}l@{\hspace{2pt}}
	@{\hspace{2pt}}c@{\hspace{2pt}}
        @{\hspace{2pt}}c@{\hspace{2pt}}
	@{\hspace{2pt}}c@{\hspace{2pt}}
	@{\hspace{2pt}}c@{\hspace{2pt}}
	@{\hspace{2pt}}c@{\hspace{2pt}}
	@{\hspace{2pt}}c@{\hspace{2pt}}
	@{\hspace{2pt}}r@{\hspace{0pt}}
      }
      \toprule
      \multicolumn{3}{c}{\multirow{2.5}{*}{Model}} & \multicolumn{7}{c}{IND} \\
      \cmidrule(lr){4-10}
       & & & Accuracy & Recall & Precision & F1 & Specificity & NPR & \#failed \\
      \midrule
     \multicolumn{2}{l}{\multirow{5}{*}{General-purpose LLMs}}
        & GPT-4 Turbo & 0.826 & 0.604 & \textbf{1.000} & 0.753 & \textbf{1.000} & 0.763 & 0 \\
       && Gemini Pro & {0.897} & {0.766} & \textbf{1.000} & {0.867} & \textbf{1.000} & {0.845} & 0 \\
       && Claude 2.1 & 0.711 & 0.360 & 0.952 & 0.523 & 0.986 & 0.664 & 1 \\
       && Llama-2 13B-chat & 0.474 & 0.459 & 0.411 & 0.434 & 0.486 & 0.535 & 0 \\
       && Mistral-7B-Instruct-v0.2 & 0.755 & 0.441 & \textbf{1.000} & 0.613 & \textbf{1.000} & 0.696 & 0 \\
       \midrule
       \multicolumn{2}{l}{{E-commerce LLM}}
       & EcomGPT & 0.648 & 0.739 & 0.577 & 0.648 & 0.577 & 0.739 & 0 \\
      \midrule
      \multicolumn{2}{l}{\multirow{2}{*}{SoTA task-specific model}}
       & BERT & \textbf{0.996} & 0.991 & \textbf{{1.000}} & \textbf{0.995} & \textbf{{1.000}} & 0.993 & 0 \\
      && DeBERTaV3 & 0.577 & \textbf{{1.000}} & 0.509 & 0.675 & 0.246 & \textbf{{1.000}} & 0 \\
      \midrule
      \multirow{12.5}{*}{\method}
      & \multirow{6}{*}{Task-specific}
      & Flan-T5 XXL & \textbf{0.996} & 0.991 & \textbf{1.000} & \textbf{0.995} & \textbf{1.000} & 0.993 & 0\\
      & & Llama-2 13B-chat & \textbf{0.996} & {0.991} & \textbf{1.000} & \textbf{0.995} & \textbf{1.000} & {0.993} & 0 \\
      & & Llama-2 7B-chat & 0.992 & 0.991 & 0.991 &  0.991 & 0.993 & 0.993 & 0\\
      & & Mistral-7B Instruct-v0.2 & 0.988 & {0.991} & 0.982 & 0.987 & 0.986 & {0.993} & 0 \\
      & & Flan-T5 XL & 0.960 & 0.910 & \textbf{1.000} & 0.953 & \textbf{1.000} & 0.934 & 0\\
      & & Phi-2 & 0.992 & 0.991 & 0.991 & 0.991 & 0.993 & 0.993 & 0 \\
      \cmidrule(lr){2-10}
      & \multirow{6}{*}{Generalist} 
      & Flan-T5 XXL & \textbf{0.996} & {0.991} & \textbf{1.000} & \textbf{0.995} & \textbf{1.000} & {0.993} & 0 \\
      & & Llama-2 13B-chat & \textbf{0.996} & {0.991} & \textbf{1.000} & \textbf{0.995} & \textbf{1.000} & {0.993} & 0 \\
      & & Llama-2 7B-chat & \textbf{0.996} & {0.991} & \textbf{1.000} & \textbf{0.995} & \textbf{1.000} & {0.993} & 0 \\
      & & Mistral-7B Instruct-v0.2 & \textbf{0.996} & {0.991} & \textbf{1.000} & \textbf{0.995} & \textbf{1.000} & {0.993} & 0 \\
      & & Flan-T5 XL & \textbf{0.996} & {0.991} & \textbf{1.000} & \textbf{0.995} & \textbf{1.000} & {0.993} & 0 \\
      & & Phi-2 & 0.992 & {0.991} & 0.991 & 0.991 & 0.993 & {0.993} & 0 \\
      \bottomrule
      \end{tabular}
      \begin{tablenotes}[normal, flushleft]
      \begin{footnotesize}
      \item
      In this table, ``IND" indicates in-domain evaluation;
      ``Task-specific" indicates that the \method models are tuned on individual tasks;
      ``Generalist" represents tuning \method models using all tasks together;
      \#failed refers to the number of failure cases that we cannot extract meaningful results from the model output;
      ``NPR" is the negative prediction rate. 
      On each task, the best performance is in \textbf{bold}.
      \par
      \end{footnotesize}
      \end{tablenotes}
  \vspace{-10pt}
  \end{threeparttable}
\end{table*}


\begin{table*}[!h]
\footnotesize
  \caption{Performance on \SA}
  \centering
  \label{tbl:sa}
  \begin{threeparttable}
      \begin{tabular}{
        @{\hspace{0pt}}l@{\hspace{2pt}}
        @{\hspace{-15pt}}l@{\hspace{2pt}}
        @{\hspace{2pt}}l@{\hspace{2pt}}
	@{\hspace{2pt}}c@{\hspace{2pt}}
    @{\hspace{3pt}}c@{\hspace{5pt}}
	@{\hspace{5pt}}c@{\hspace{5pt}}
	@{\hspace{5pt}}c@{\hspace{2pt}}
	@{\hspace{-2pt}}r@{\hspace{2pt}}
	 @{\hspace{2pt}}c@{\hspace{2pt}}
 	  @{\hspace{2pt}}c@{\hspace{2pt}}
        @{\hspace{5pt}}c@{\hspace{5pt}}
	  @{\hspace{5pt}}c@{\hspace{5pt}}
	  @{\hspace{5pt}}c@{\hspace{2pt}}
	  @{\hspace{-2pt}}r@{\hspace{0pt}}
      }
      \toprule
      \multicolumn{3}{c}{\multirow{2.5}{*}{Model}} & \multicolumn{5}{c}{IND} && \multicolumn{5}{c}{OOD} \\
      \cmidrule(lr){4-8}       \cmidrule(lr){10-14}
       & & & Acc & M-Rec & M-Pre & M-F1 & \#failed  & & Acc & M-Rec & M-Pre & M-F1 & \#failed\\
      \midrule
     \multicolumn{2}{l}{\multirow{5}{*}{General-purpose LLMs}}
        & GPT-4 Turbo & 0.595 & {0.575} & 0.544 & {0.516} & 0 && 0.556 & {0.586} & {0.544} & {0.510} & 0 \\
       && Gemini Pro & 0.609 & 0.521 & 0.453 & 0.470 & 2 && 0.572 & 0.511 & 0.444 & 0.454 & 1 \\
       && Claude 2.1 & 0.375 & 0.510 & 0.474 & 0.415 & 2 && 0.328 & 0.466 & 0.447 & 0.369 & 1 \\
       && Llama-2 13B-chat & 0.406 & 0.188 & 0.191 & 0.188 & 0 && 0.384 & 0.179 & 0.180 & 0.178 & 0 \\
       && Mistral-7B-Instruct-v0.2 & {0.633} & 0.532 & {0.551} & 0.470 & 0 && 0.594 & 0.531 & 0.494 & 0.438 & 0 \\
       \midrule
       \multicolumn{2}{l}{\multirow{1}{*}{E-commerce LLM}}        
       & EcomGPT & 0.191 & 0.362 & 0.341 & 0.188 & 6 && 0.196 & 0.375 & 0.336 &  0.178 & 13 \\
      \midrule
      \multicolumn{2}{l}{\multirow{3}{*}{SoTA task-specific model}}
      & BERTweet & {0.733} & {0.503} & {0.530} & {0.511} & 0 && {0.729} & {0.507} & {0.524} & {0.513} & 0 \\
      && DeBERTaV3 & {0.768} & {0.567} & {0.607} & {0.573} & 0 && {0.764} & {0.565} & {0.591} & {0.567} & 0 \\
      && P5 & 0.611 & 0.199 & 0.157 & 0.156 & 0 && 0.620 & 0.200 & 0.124 & 0.153 & 0 \\
      \midrule
      \multirow{11.5}{*}{\method}
      & \multirow{6}{*}{Task-specific}
      & Flan-T5 XXL & 0.783 & 0.619 & 0.618 & 0.612 & 0 
      && 0.770 & 0.604 & 0.601 &  0.600 & 0\\
      & & Llama-2 13B-chat & 0.791 & 0.616 & 0.641 & 0.616 & 0 && 0.781 & {0.627} & 0.645 & 0.629 & 0 \\
      & & Llama-2 7B-chat & 0.790 & 0.620 & 0.652 & 0.634 & 0 
      && 0.769 & 0.583 & 0.599 & 0.589 & 0\\
      & & Mistral-7B Instruct-v0.2 & \textbf{0.801} & {0.643} & \textbf{0.676} & \textbf{0.655} & 0 && \textbf{0.789} & 0.619 & {0.650} & {0.632} & 0 \\
      & & Flan-T5 XL & 0.771 & 0.645 & 0.638 & 0.620 & 0 
      && 0.743 & 0.594 & 0.592 & 0.582 & 0\\
      & & Phi-2 & 0.779 & 0.611 & 0.618 & 0.608 & 0 && 0.754 & 0.576 & 0.594 & 0.583 & 0 \\
      \cmidrule(lr){3-14}
      & \multirow{6}{*}{Generalist} 
      & Flan-T5 XXL & {0.797} & 0.629 & 0.646 & 0.628 & 0 && {0.787} & 0.619 & 0.624 & 0.619 & 0 \\
      & & Llama-2 13B-chat & 0.796 & 0.641 & {0.661} & {0.648} & 0 && 0.785 & 0.621 & 0.638 & 0.629 & 0 \\
      
      & & Llama-2 7B-chat & 0.768 & 0.579 & 0.589 & 0.580 & 0 && 0.776 & 0.599 & 0.626 & 0.606 & 0 \\
      & & Mistral-7B Instruct-v0.2 & 0.781 & 0.630 & 0.654 & 0.639 & 0 && 0.784 & \textbf{0.630} & \textbf{0.653} & \textbf{0.640} & 0 \\
      
      & & Flan-T5 XL & 0.782 & \textbf{0.654} & 0.655 & {0.648} & 0 && 0.753 & 0.604 & 0.598 & 0.598 & 0 \\
      & & Phi-2 & 0.780 & 0.588 & 0.619 & 0.596 & 0 && 0.758 & 0.552 & 0.590 & 0.565 & 0 \\
      \bottomrule
      \end{tabular}
      \begin{tablenotes}[normal, flushleft]
      \begin{footnotesize}
      \item
      In this table, ``IND" and ``OOD" indicates in-domain evaluation and out-of-domain evaluation, respectively;
      ``Task-specific" indicates that the \method models are tuned on individual tasks;
      ``Generalist" represents tuning \method models using all tasks together;
      \#failed refers to the number of failure cases that we cannot extract meaningful results from the model output.
      The metric ``Acc'', ``M-Rec", ``M-Pre", and ``M-F1" represents accuracy, macro recall, macro precision, and macro F1, respectively.
      On each task, the best performance is in \textbf{bold}.
      \par
      \end{footnotesize}
      \end{tablenotes}
  \vspace{-10pt}
  \end{threeparttable}
\end{table*}


\begin{table*}[!h]
\footnotesize
  \caption{Performance on \MPC}
  \centering
  \label{tbl:mpc}
  \begin{threeparttable}
      \begin{tabular}{
        @{\hspace{0pt}}l@{\hspace{2pt}}
        @{\hspace{2pt}}l@{\hspace{2pt}}
        @{\hspace{2pt}}l@{\hspace{2pt}}
	@{\hspace{2pt}}c@{\hspace{2pt}}
    @{\hspace{2pt}}c@{\hspace{4pt}}
	@{\hspace{4pt}}c@{\hspace{4pt}}
	@{\hspace{6pt}}c@{\hspace{2pt}}
	@{\hspace{2pt}}r@{\hspace{0pt}}
      }
      \toprule
      \multicolumn{3}{c}{\multirow{2.5}{*}{Model}} & \multicolumn{5}{c}{IND} \\
      \cmidrule(lr){4-8}
       & & & Accuracy & M-Rec & M-Pre & M-F1 & \#failed \\
      \midrule
     \multicolumn{2}{l}{\multirow{5}{*}{General-purpose LLMs}}
        & GPT-4 Turbo & 0.611 & \textbf{0.527} & \textbf{0.540} & \textbf{0.487} & 0 \\
       && Gemini Pro & 0.584 & 0.471 & 0.414 & 0.425 & 2 \\
       && Claude 2.1 & {0.655} & 0.464 & 0.419 & 0.435 & 13 \\
       && Llama-2 13B-chat & 0.504 & 0.250 & 0.251 & 0.250 & 0 \\
       && Mistral-7B-Instruct-v0.2 & 0.529 & 0.395 & 0.384 & 0.365 & 0 \\
       \midrule
       \multicolumn{2}{l}{{E-commerce LLM}}
       & EcomGPT & 0.540 & 0.265 & 0.218 & 0.223 & 2 \\
      \midrule
      \multicolumn{2}{l}{\multirow{2}{*}{SoTA task-specific model}}
       & BERT & 0.661 & 0.381 & 0.423 & 0.393 & 0 \\
      && DeBERTaV3 & \textbf{0.703} & {0.436} & {0.472} & {0.448} & 0 \\
      \midrule
      \multirow{12}{*}{\method}
      & \multirow{6}{*}{Task-specific}
      & Flan-T5 XXL & 0.666 & 0.438 & 0.412 & 0.346 & 0 \\
      & & Llama-2 13B-chat & 0.655 & 0.399 & 0.410 & 0.349 & 0 \\
      & & Llama-2 7B-chat & 0.659 & 0.399 & 0.531 & 0.330 & 0\\
      & & Mistral-7B Instruct-v0.2 & {0.681} & {0.406} & {0.423} & {0.387} & 0 \\
      & & Flan-T5 XL & 0.648 & 0.425 & 0.361 & 0.327 & 0\\
      & & Phi-2 & 0.646 & 0.387 & 0.316 & 0.321 & 0 \\
      \cmidrule(lr){2-8}
      & \multirow{6}{*}{Generalist} 
      & Flan-T5 XXL & 0.680 & 0.431 & 0.416 & 0.364 & 0 \\
      & & Llama-2 13B-chat & 0.684 & 0.440 & 0.435 & 0.414 & 0 \\
      
      & & Llama-2 7B-chat & 0.679 & 0.427 & 0.434 & 0.398 & 0 \\
      & & Mistral-7B Instruct-v0.2 & {0.696} & {0.450} & 0.456 & {0.443} & 0 \\
      
      & & Flan-T5 XL & 0.663 & 0.395 & {0.533} & 0.332 & 0 \\
      & & Phi-2 & 0.650 & 0.397 & 0.410 & 0.335 & 0 \\
      \bottomrule
      \end{tabular}
      \begin{tablenotes}[normal, flushleft]
      \begin{footnotesize}
      \item
      In this table, ``IND" indicates in-domain evaluation;
      ``Task-specific" indicates that the \method models are tuned on individual tasks;
      ``Generalist" represents tuning \method models using all tasks together;
      \#failed refers to the number of failure cases that we cannot extract meaningful results from the model output.
      The metric ``M-Rec", ``M-Pre", and ``M-F1" represents macro recall, macro precision, and macro F1, respectively.
      On each task, the best performance is in \textbf{bold}.
      \par
      \end{footnotesize}
      \end{tablenotes}
  \vspace{-10pt}
  \end{threeparttable}
\end{table*}


\begin{table*}[!h]
\footnotesize
  \caption{Performance on \PSI}
  \centering
  \label{tbl:psi}
  \begin{threeparttable}
      \begin{tabular}{
        @{\hspace{0pt}}l@{\hspace{2pt}}
        @{\hspace{2pt}}l@{\hspace{2pt}}
        @{\hspace{2pt}}l@{\hspace{2pt}}
	@{\hspace{2pt}}c@{\hspace{2pt}}
    @{\hspace{2pt}}c@{\hspace{2pt}}
	@{\hspace{4pt}}c@{\hspace{4pt}}
	@{\hspace{4pt}}c@{\hspace{4pt}}
	@{\hspace{4pt}}c@{\hspace{4pt}}
	@{\hspace{2pt}}c@{\hspace{2pt}}
	@{\hspace{-2pt}}r@{\hspace{0pt}}
      }
      \toprule
      \multicolumn{3}{c}{\multirow{2.5}{*}{Model}} & \multicolumn{7}{c}{IND} \\
      \cmidrule(lr){4-10}
       & & & Accuracy & Recall & Precision & F1 & Specificity & NPR & \#failed \\
      \midrule
     \multicolumn{2}{l}{\multirow{5}{*}{General-purpose LLMs}}
       & GPT-4 Turbo & 0.289 & 0.374 & 0.132 & 0.195 & 0.264 & 0.585 & 0 \\
       && Gemini Pro & 0.296 & 0.504 & 0.164 & 0.248 & 0.234 & 0.612 & 0 \\
       && Claude 2.1 & 0.291 & 0.578 & 0.179 & 0.273 & 0.205 & 0.620 & 1 \\
       && Llama-2 13B-chat & {0.649} & 0.257 & {0.247} & 0.252 & 0.766 & {0.775} & 0 \\
       && Mistral-7B-Instruct-v0.2 & 0.361 & \textbf{0.609} & 0.203 & {0.305} & 0.287 & 0.711 & 0 \\
       \midrule      
       \multicolumn{2}{l}{{E-commerce LLM}}        
       & EcomGPT & 0.630 & 0.165 & 0.176 & 0.170 & {0.769}  & 0.755 & 13 \\
      \midrule
      \multicolumn{2}{l}{\multirow{2}{*}{SoTA task-specific model}}
       & BERT & 0.761 & {0.330} & {0.472} & {0.389} & 0.890 & {0.816} & 0 \\
      && DeBERTaV3 & {0.769} & 0.000 & 0.000 & 0.000 & {0.999} & 0.770 & 0 \\
      \midrule
      \multirow{12}{*}{\method}
      & \multirow{6}{*}{Task-specific}
      & Flan-T5 XXL & 0.766 & 0.013 & 0.300 & 0.025 & 0.991 & 0.771 & 0\\
      & & Llama-2 13B-chat & {0.770} & 0.000 & 0.000 & 0.000 & \textbf{1.000} & {0.770} & 0 \\
      & & Llama-2 7B-chat & 0.770 & 0.017 & 0.500 & 0.034 & 0.995 & 0.772 & 0\\
      & & Mistral-7B Instruct-v0.2 & {0.770} & 0.000 & 0.000 & 0.000 & \textbf{1.000} & {0.770} & 0 \\
      & & Flan-T5 XL & 0.768 & 0.000 & 0.000 & 0.000 & 0.997 & 0.770 & 0 \\
      & & Phi-2 & 0.770 & 0.000 & 0.000 & 0.000 & \textbf{1.000} & 0.770 & 0 \\
      \cmidrule(lr){2-10}
      & \multirow{6}{*}{Generalist} 
      & Flan-T5 XXL & 0.771 & 0.300 & 0.504 & 0.376 & 0.912 & 0.813 & 0 \\
      & & Llama-2 13B-chat & \textbf{0.795} & {0.448} & 0.569 & \textbf{0.501} & 0.899 & \textbf{0.845} & 0 \\
      
      & & Llama-2 7B-chat & 0.781 & 0.283 & 0.546 & 0.372 & 0.930 & 0.813 & 0 \\
      & & Mistral-7B Instruct-v0.2 & 0.790 & 0.200 & 0.639 & 0.305 & 0.966 & 0.802 & 0 \\
      
      & & Flan-T5 XL & 0.773 & 0.022 & \textbf{0.714} & 0.042 & {0.997} & 0.773 & 0 \\
      & & Phi-2 & 0.777 & 0.313 & 0.526 & 0.392 & 0.916 & 0.817 & 0 \\
      \bottomrule
      \end{tabular}
      \begin{tablenotes}[normal, flushleft]
      \begin{footnotesize}
      \item
      In this table, ``IND" indicates in-domain evaluation;
      ``Task-specific" indicates that the \method models are tuned on individual tasks;
      ``Generalist" represents tuning \method models using all tasks together;
      \#failed refers to the number of failure cases that we cannot extract meaningful results from the model output;
      ``NPR" is the negative prediction rate. 
      On each task, the best performance is in \textbf{bold}.
      \par
      \end{footnotesize}
      \end{tablenotes}
  \vspace{-10pt}
  \end{threeparttable}
\end{table*}


\begin{sidewaystable*}
\footnotesize
  \caption{Performance on \AP}
  \centering
  \label{tbl:ap}
  \begin{threeparttable}
      \begin{tabular}{
        @{\hspace{0pt}}l@{\hspace{2pt}}
        @{\hspace{-3pt}}l@{\hspace{2pt}}
        @{\hspace{2pt}}l@{\hspace{2pt}}
	@{\hspace{2pt}}c@{\hspace{2pt}}
    @{\hspace{2pt}}c@{\hspace{2pt}}
	@{\hspace{2pt}}c@{\hspace{2pt}}
	@{\hspace{2pt}}c@{\hspace{2pt}}
    @{\hspace{2pt}}c@{\hspace{2pt}}
	@{\hspace{2pt}}c@{\hspace{2pt}}
	@{\hspace{-2pt}}r@{\hspace{2pt}}
	 @{\hspace{0pt}}c@{\hspace{0pt}}
 	@{\hspace{2pt}}c@{\hspace{2pt}}
	 @{\hspace{2pt}}c@{\hspace{2pt}}
	 @{\hspace{2pt}}c@{\hspace{2pt}}
	 @{\hspace{2pt}}c@{\hspace{2pt}}
	 @{\hspace{2pt}}c@{\hspace{2pt}}
	 @{\hspace{2pt}}c@{\hspace{2pt}}
	 @{\hspace{-2pt}}r@{\hspace{0pt}}
      }
      \toprule
      \multicolumn{3}{c}{\multirow{2}{*}{Model}} & \multicolumn{7}{c}{IND} && \multicolumn{7}{c}{OOD} \\
      \cmidrule(lr){4-10}       \cmidrule(lr){12-18}
       & & & Accuracy & Recall & Precision & F1 & Specificity & NPR & \#failed  & & Accuracy & Recall & Precision & F1 & Specificity & NPR & \#failed\\
      \midrule
     \multicolumn{2}{l}{\multirow{5}{*}{General-purpose LLMs}}
        & GPT-4 Turbo & {0.623} & 0.550 & 0.791 & {0.649} & 0.749 & {0.491} & 5 && {0.641} & 0.555 & \textbf{0.876} & {0.680} & 0.828 & {0.460} & 2 \\
       && Gemini Pro & 0.542 & 0.371 & {0.797} & 0.506 & 0.837 & 0.435 & 13 && 0.543 & 0.410 & 0.844 & 0.552 & 0.834 & 0.393 & 6 \\
       && Claude 2.1 & 0.424 & 0.177 & 0.671 & 0.280 & \textbf{0.850} & 0.375 & 90 && 0.384 & 0.146 & 0.769 & 0.245 & \textbf{0.904} & 0.326 & 73 \\
       && Llama-2 13B-chat & 0.534 & {0.608} & 0.638 & 0.623 & 0.406 & 0.375 & 0 && 0.541 & {0.605} & 0.688 & 0.644 & 0.401 & 0.317 & 0 \\
       && Mistral-7B-Instruct-v0.2 & 0.522 & 0.539 & 0.647 & 0.588 & 0.493 & 0.383 & 120 && 0.537 & 0.523 & 0.725 & 0.608 & 0.567 & 0.352 & 126 \\
       \midrule
       \multicolumn{2}{l}{{E-commerce LLM}}
       & EcomGPT & 0.318 & 0.051 & 0.283 &  0.086 & 0.779 & 0.322 & 254 && 0.286 & 0.085 & 0.403 & 0.140 & 0.726 & 0.266 & 245 \\
      \midrule
      \multicolumn{2}{l}{\multirow{2}{*}{SoTA task-specific model}}
       & BERT & {0.749} & \textbf{0.970} & 0.726 & {0.830} & 0.368 & \textbf{0.877} & 0 && {0.803} & {0.831} & \textbf{0.876} & {0.853} & 0.742 & {0.668} & 0 \\
      && DeBERTaV3 & 0.504 & 0.310 & {0.769} & 0.441 & {0.839} & 0.413 & 0 && 0.501 & 0.345 & 0.826 & 0.487 & {0.841} & 0.370 & 0 \\
      \midrule
      \multirow{12}{*}{\method}
      & \multirow{6}{*}{Task-specific}
      & Flan-T5 XXL & 0.749 & 0.875 & 0.763 & 0.815 & 0.531 & 0.712 & 0 
      && 0.799 & 0.889 & 0.830 & 0.859 & 0.602 & 0.713 & 0\\
      & & Llama-2 13B-chat & 0.801 & {0.919} & 0.797 & 0.854 & 0.597 & {0.811} & 0 && \textbf{0.838} & {0.924} & 0.852 & \textbf{0.887} & 0.650 & {0.797} & 0 \\
      & & Llama-2 7B-chat & 0.741 & 0.889 & 0.749 & 0.813 & 0.485 & 0.718 & 0 
      && 0.761 & 0.864 & 0.802 & 0.832 & 0.535 & 0.644 & 0\\
      
      & & Mistral-7B Instruct-v0.2 & \textbf{0.821} & 0.896 & {0.834} & {0.864} & {0.692} & 0.794 & 0 && 0.835 & 0.891 & {0.872} & 0.881 & {0.713} & 0.749 & 0 \\
      & & Flan-T5 XL & 0.693 & 0.684 & 0.802 & 0.738 & 0.708 & 0.565 & 0
      && 0.707 & 0.682 & 0.862 & 0.762 & 0.761 & 0.523 & 0\\
      & & Phi-2 & 0.765 & 0.942 & 0.751 & 0.835 & 0.460 & 0.820 & 0 && 0.781 & 0.950 & 0.779 & 0.856 & 0.411 & 0.791 & 0 \\
      \cmidrule(lr){2-18}
      & \multirow{6}{*}{Generalist} 
      & Flan-T5 XXL       & 0.774 & 0.910 & 0.773 & 0.836 & 0.540 & 0.776 & 0 && 0.814 & 0.914 & 0.832 & 0.871 & 0.596 & 0.760 & 0 \\
      & & Llama-2 13B-chat & 0.808 & 0.864 & \textbf{0.838} & 0.851 & {0.711} & 0.752 & 0 && 0.818 & 0.862 & {0.872} & 0.867 & {0.723} & 0.705 & 0 \\
      & & Llama-2 7B-chat & {0.817} & {0.921} & 0.814 & \textbf{0.864} & 0.638 & {0.824} & 0 && {0.836} & 0.931 & 0.845 & {0.886} & 0.627 & \textbf{0.807} & 0 \\
      & & Mistral-7B Instruct-v0.2 & 0.797 & 0.880 & 0.814 & 0.846 & 0.654 & 0.759 & 0 && 0.832 & 0.885 & {0.872} & 0.878 & 0.717 & 0.740 & 0 \\
      & & Flan-T5 XL      & 0.765 & 0.888 & 0.774 & 0.827 & 0.553 & 0.741 & 0 && 0.819 & 0.891 & 0.852 & 0.871 & 0.662 & 0.735 & 0 \\
      & & Phi-2 & 0.794 & 0.897 & 0.801 & 0.846 & 0.616 & 0.777 & 0 && 0.823 & \textbf{0.937} & 0.828 & 0.879 & 0.573 & \textbf{0.807} & 0 \\
      \bottomrule
      \end{tabular}
      \begin{tablenotes}[normal, flushleft]
      \begin{footnotesize}
      \item
      In this table, ``IND" and ``OOD" indicates in-domain evaluation and out-of-domain evaluation, respectively;
      ``Task-specific" indicates that the \method models are tuned on individual tasks;
      ``Generalist" represents tuning \method models using all tasks together;
      \#failed refers to the number of failure cases that we cannot extract meaningful results from the model output;
      ``NPR" is the negative prediction rate. 
      On each task, the best performance is in {bold}.
      \par
      \end{footnotesize}
      \end{tablenotes}
  \vspace{-10pt}
  \end{threeparttable}
\end{sidewaystable*}


\begin{table*}[!h]
  \caption{Performance on \AG}
  \footnotesize
  \centering
  \label{tbl:ag}
  \begin{threeparttable}
      \begin{tabular}{
        @{\hspace{0pt}}l@{\hspace{2pt}}
        @{\hspace{-15pt}}l@{\hspace{2pt}}
        @{\hspace{2pt}}l@{\hspace{2pt}}
	@{\hspace{2pt}}c@{\hspace{3pt}}
    @{\hspace{3pt}}c@{\hspace{3pt}}
	@{\hspace{3pt}}c@{\hspace{3pt}}
	@{\hspace{3pt}}c@{\hspace{3pt}}
	@{\hspace{-3pt}}r@{\hspace{3pt}}
	 @{\hspace{3pt}}c@{\hspace{1pt}}
 	  @{\hspace{2pt}}c@{\hspace{3pt}}
        @{\hspace{3pt}}c@{\hspace{3pt}}
	  @{\hspace{3pt}}c@{\hspace{3pt}}
	  @{\hspace{3pt}}c@{\hspace{3pt}}
	  @{\hspace{-3pt}}r@{\hspace{0pt}}
      }
      \toprule
      \multicolumn{3}{c}{\multirow{2.5}{*}{Model}} & \multicolumn{5}{c}{IND} && \multicolumn{5}{c}{OOD} \\
      \cmidrule(lr){4-8}       \cmidrule(lr){10-14}
       & & & R$_{\text{BERT}}$ & P$_{\text{BERT}}$ & F$_{\text{BERT}}$ & BLEURT & \#failed  & & R$_{\text{BERT}}$ & P$_{\text{BERT}}$ & F$_{\text{BERT}}$ & BLEURT & \#failed\\
      \midrule
     \multicolumn{2}{l}{\multirow{5}{*}{General-purpose LLMs}}
        & GPT-4 Turbo & 0.847 & \textbf{0.869} & \textbf{0.858} & 0.280 & 0 && \textbf{0.852} & \textbf{0.868} & \textbf{0.860} & 0.283 & 0 \\
       && Gemini Pro & 0.844 & 0.867 & 0.855 & 0.269 & 0 && 0.847 & 0.866 & 0.856 & 0.264 & 0 \\
       && Claude 2.1 & 0.848 & 0.835 & 0.841 & \textbf{0.314} & 0 && 0.851 & 0.833 & 0.842 & \textbf{0.325} & 0 \\
       && Llama-2 13B-chat & 0.845 & 0.780 & 0.811 & 0.261 & 0 && 0.845 & 0.775 & 0.808 & 0.260 & 0 \\
       && Mistral-7B-Instruct-v0.2 & \textbf{0.850} & 0.856 & 0.853 & 0.288 & 0 && \textbf{0.852} & 0.851 & 0.851 & 0.290 & 0 \\
       \midrule
       \multicolumn{2}{l}{{E-commerce LLM}}
       & EcomGPT & 0.675 & 0.665 & 0.669 & 0.290 & 0 && 0.729 & 0.716 & 0.722 &  0.296 & 0 \\
      \midrule
      \multicolumn{2}{l}{{SoTA task-specific model}} & GPT-4 Turbo & 0.847 & \textbf{0.869} & \textbf{0.858} & 0.280 & 0 && \textbf{0.852} & \textbf{0.868} & \textbf{0.860} & 0.283 & 0 \\
      \midrule
      \multirow{11}{*}{\method}
      & \multirow{6}{*}{Task-specific}
      & Flan-T5 XXL & 0.822 & 0.864 & 0.842 & 0.310 & 0
      && 0.824 & 0.865 & 0.843 & 0.302 & 0\\
      & & Llama-2 13B-chat & {0.824} & {0.861} & {0.841} & 0.309 & 0 && 0.821 & 0.860 & 0.840 & 0.289 & 0 \\
      & & Llama-2 7B-chat & 0.822 & 0.861 & 0.841 & 0.301 & 0 
      && 0.820 & 0.861 & 0.840 &  0.289 & 0\\
      & & Mistral-7B Instruct-v0.2 & 0.823 & 0.860 & 0.841 & {0.310} & 0 && {0.823} & {0.861} & {0.842} & {0.298} & 0 \\
      & & Flan-T5 XL & 0.823 & 0.864 & 0.843 & 0.320 & 0
      && 0.824 & 0.864 & 0.843 & 0.307 & 0\\
      & & Phi-2 & 0.817 & 0.855 & 0.835 & 0.283 & 0 && 0.817 & 0.856 & 0.835 & 0.270 & 0 \\
      \cmidrule(lr){2-14}
      & \multirow{6}{*}{Generalist} 
      & Flan-T5 XXL & {0.824} & {0.865} & {0.844} & 0.224 & 0 && 0.823 & 0.864 & 0.843 & 0.206 & 0 \\
      & & Llama-2 13B-chat & 0.823 & 0.861 & 0.841 & 0.215 & 0 && 0.822 & 0.861 & {0.841} & 0.195 & 0 \\
      & & Llama-2 7B-chat & 0.822 & 0.860 & 0.840 & 0.208 & 0 && 0.819 & 0.859 & 0.838 & 0.188 & 0 \\
      & & Mistral-7B Instruct-v0.2 & 0.822 & 0.864 & 0.842 & 0.213 & 0 && 0.821 & 0.862 & 0.840 & 0.194 & 0 \\
      & & Flan-T5 XL & 0.823 & 0.864 & 0.843 & {0.227} & 0 && {0.824} & {0.865} & {0.844} & {0.211} & 0 \\
      & & Phi-2 & 0.823 & 0.861 & 0.842 & 0.222 & 0 && 0.821 & 0.859 & 0.840 &0.198 & 0 \\
      \bottomrule
      \end{tabular}
      \begin{tablenotes}[normal, flushleft]
      \begin{footnotesize}
      \item
      In this table, ``IND" and ``OOD" indicates in-domain evaluation and out-of-domain evaluation, respectively;
      ``Task-specific" indicates that the \method models are tuned on individual tasks;
      ``Generalist" represents tuning \method models using all tasks together;
      \#failed refers to the number of failure cases that we cannot extract meaningful results from the model output.
      The metrics ``R$_{\text{BERT}}$", ``P$_{\text{BERT}}$", ``F$_{\text{BERT}}$" and ``BLEURT" are detailed in Table~\ref{tbl:task_data}.
      On each task, the best performance is in \textbf{bold}.
      Note that we use GPT-4 Turbo as the SoTA task-specific model in this task.
      \par
      \end{footnotesize}
      \end{tablenotes}
  \vspace{-10pt}
  \end{threeparttable}
\end{table*}


\begin{table*}[!h]
\footnotesize
  \caption{Performance on \SR and \QPR}
  \centering
  \label{tbl:sr_qpr}
  \begin{threeparttable}
      \begin{tabular}{
        @{\hspace{0pt}}l@{\hspace{2pt}}
        @{\hspace{2pt}}l@{\hspace{2pt}}
        @{\hspace{2pt}}l@{\hspace{2pt}}
	@{\hspace{2pt}}c@{\hspace{2pt}}
    	@{\hspace{2pt}}r@{\hspace{2pt}}
	@{\hspace{10pt}}c@{\hspace{10pt}}
	 @{\hspace{2pt}}c@{\hspace{2pt}}
      	 @{\hspace{2pt}}r@{\hspace{2pt}}
	 @{\hspace{10pt}}c@{\hspace{10pt}}
	 @{\hspace{2pt}}c@{\hspace{2pt}}
	 @{\hspace{2pt}}r@{\hspace{2pt}}
	 @{\hspace{2pt}}c@{\hspace{0pt}}
      }
      \toprule
      \multicolumn{3}{c}{\multirow{4}{*}{Model}} & \multicolumn{5}{c}{\SR}  && \multicolumn{2}{c}{\QPR} \\
            \cmidrule(lr){4-8}    \cmidrule(lr){10-11}
      && & \multicolumn{2}{c}{IND} &&  \multicolumn{2}{c}{OOD}  && \multicolumn{2}{c}{IND} \\
      \cmidrule(lr){4-5} \cmidrule(lr){6-8}   \cmidrule(lr){10-11}
       & & & HR@1 & \#failed && HR@1 & \#failed && NDCG & \#failed\\
      \midrule
     \multicolumn{2}{l}{\multirow{5}{*}{General-purpose LLMs}}
        & GPT-4 Turbo & {0.387} & 0 && {0.198} & 0 && {0.875} & 14 \\
       && Gemini Pro & 0.269 & 2 && 0.116 & 3 && 0.821 & 52 \\
       && Claude 2.1 & 0.066 & 34 && 0.036 & 42 && 0.821 & 26 \\
       && Llama-2 13B-chat & 0.056 & 0 && 0.050 & 0 && 0.815 & 0 \\
       && Mistral-7B-Instruct-v0.2 & 0.164 & 1 && 0.108 & 0 && 0.842 & 4 \\
       \midrule      
       \multicolumn{2}{l}{{E-commerce LLM}}        
       & EcomGPT & 0.042 & 344 && 0.023 & 391 && 0.000 & 1000 \\
      \midrule
      \multicolumn{2}{l}{\multirow{2}{*}{SoTA task-specific model}}
       & gSASRec / BERT & 0.249 & 0 && 0.065 & 0 && 0.839 & 0 \\
      && Recformer / DeBERTaV3 & {0.265} & 0 && {0.081} & 0 && {0.859} & 0 \\
      \midrule
      \multirow{12}{*}{\method}
      & \multirow{6}{*}{Task-specific}
      & Flan-T5 XXL & 0.467 & 0 &&  0.252 & 0 && 0.881 & 0\\
      & & Llama-2 13B-chat & 0.518 & 0 && 0.263 & 0 && 0.879 & 0 \\
      && Llama-2 7B-chat & 0.517 & 0 && 0.228 & 0 && 0.867 & 0\\
      && Mistral-7B Instruct-v0.2 & {0.535} & 0 && {0.268} & 0 && {0.883} & 0 \\
      && Flan-T5 XL & 0.436 & 0 && 0.226 & 0 && 0.875 & 0\\
      && Phi-2 & 0.413 & 5 && 0.219 & 10 && 0.858 & 0  \\
      \cmidrule(lr){2-11}
      & \multirow{6}{*}{Generalist} 
      & Flan-T5 XXL & 0.512 & 0 && 0.262 & 0 && \textbf{0.885} & 0 \\
      && Llama-2 13B-chat & 0.526 & 0 && 0.273 & 0 && 0.870 & 0 \\
      && Llama-2 7B-chat & 0.517 & 0 && 0.261 & 0 && 0.868 & 0 \\
      && Mistral-7B Instruct-v0.2 & \textbf{0.542} & 0 && \textbf{0.280} & 0 && 0.876 & 0 \\
      && Flan-T5 XL & 0.463 & 0 && 0.256 & 0 && 0.868 & 0 \\
      && Phi-2 & 0.479 & 5 && 0.241 & 8 && 0.870 & 0 \\
      \bottomrule
      \end{tabular}
      \begin{tablenotes}[normal, flushleft]
      \begin{footnotesize}
      \item
      In this table, ``IND" and ``OOD" indicates in-domain evaluation and out-of-domain evaluation, respectively;
      ``Task-specific" indicates that the \method models are tuned on individual tasks;
      ``Generalist" represents tuning \method models using all tasks together;
      \#failed refers to the number of failure cases that we cannot extract meaningful results from the model output.
      The metrics ``HR@1" and ``NDCG" are detailed in Appendix~\ref{sec:appendix:preprocessing}.
      On each task, the best performance is in \textbf{bold}.
      \par
      \end{footnotesize}
      \end{tablenotes}
  \vspace{-10pt}
  \end{threeparttable}
\end{table*}


\section{OOD Evaluation}
\label{sec:appendix:ood}

The following parts discuss the OOD results under different settings.

\subsection{Results on Unseen Instructions}
\label{sec:appendix:ood:unseen}
Table~\ref{tbl:instruction_ood} exhibits the results of evaluating \method on OOD test sets with unseen instruction.
As shown in table~\ref{tbl:instruction_ood}, \method models perform better when training on \dataset with diverse instructions. 
\method fine-tuned on \dataset with diverse instruction enhances the generalizability of models to the unseen instruction.

\begin{table}[!h]
\footnotesize
  \caption{Performance on Unseen Instructions in OOD Evaluation}
  \centering
  \label{tbl:instruction_ood}
  \begin{threeparttable}
      \begin{tabular}{
        @{\hspace{4pt}}l@{\hspace{8pt}}
        @{\hspace{0pt}}c@{\hspace{8pt}}
	  @{\hspace{8pt}}c@{\hspace{8pt}}
	  @{\hspace{8pt}}c@{\hspace{4pt}}
	  @{\hspace{4pt}}c@{\hspace{6pt}}
	  @{\hspace{6pt}}c@{\hspace{8pt}}
	  @{\hspace{10pt}}c@{\hspace{8pt}}
        @{\hspace{8pt}}c@{\hspace{4pt}}
      }
      \toprule
      \multirow{2.5}{*}{Model} & \multirow{2.5}{0.1\textwidth}{\centering{Training Instructions}} & \AVE & \IRP & \SA & \SR & \AP & \AG \\ 
      \cmidrule(lr){3-8}
      & & F1* & Macro F1 & Macro F1 & HR@1 & F1 & F$_{\text{BERT}}$ \\ 
      \midrule
      \multirow{2}{*}{\methodL}
      & single & 0.001 & 0.561 & 0.636 & 0.260 & \textbf{0.890} & 0.839 \\ 
      & diverse & \textbf{0.276} & \textbf{0.577} & \textbf{0.652} & \textbf{0.266} & 0.877 & \textbf{0.840} \\ 
      \midrule
      \multirow{2}{*}{\methodB} 
      & single & 0.000 & \textbf{0.527} & 0.584 & \textbf{0.284} & \textbf{0.877} & \textbf{0.849} \\ 
      & diverse & \textbf{0.366} & 0.507 & \textbf{0.628} & 0.275 & 0.863 & 0.841 \\ 
      \midrule
      \multirow{2}{*}{\methodS} 
      & single & \textbf{0.305} & 0.497 & 0.555 & \textbf{0.249} & 0.866 & 0.838 \\ 
      & diverse & 0.275 & \textbf{0.513} & \textbf{0.574} & 0.248 & \textbf{0.880} & \textbf{0.841} \\ 

      \bottomrule
      \end{tabular}
      \begin{tablenotes}[normal,flushleft]
      \begin{footnotesize}
      \item
      In this table, ``single" and ``diverse" indicate that the \method models are tuned over single and diverse instructions, respectively.
      The best performance of each \method model tuned over single and diverse instructions on each task is in \textbf{bold}.
      \par
      \end{footnotesize}
      \end{tablenotes}
  \end{threeparttable}
\end{table}


\subsection{Results on Different Base Models}
\label{sec:appendix:ood:base}
Table~\ref{tbl:base_ood} shows the OOD evaluation results of \method with different base models.
With high-quality \dataset dataset, \method models achieve good performance under different base models.

\begin{table*}[!h]
  \caption{Performance on Various Base Models in OOD Evaluation}
  \centering
  \label{tbl:base_ood}
  \footnotesize
  \begin{threeparttable}
      \begin{tabular}{
        @{\hspace{4pt}}l@{\hspace{8pt}}
	  @{\hspace{8pt}}c@{\hspace{8pt}}
        @{\hspace{8pt}}c@{\hspace{8pt}}
	  @{\hspace{8pt}}c@{\hspace{4pt}}
	  @{\hspace{4pt}}c@{\hspace{6pt}}
	  @{\hspace{6pt}}c@{\hspace{8pt}}
	  @{\hspace{10pt}}c@{\hspace{8pt}}
        @{\hspace{8pt}}c@{\hspace{4pt}}
      }
      \toprule
      \multirow{2.5}{*}{Model} & \multirow{2.5}{*}{Base Model}
      & \AVE & \IRP & \SA & \SR & \AP & \AG \\ 
      \cmidrule(lr){3-8}
      && F1* & Macro F1 & Macro F1 & HR@1 & F1 & F$_{\text{BERT}}$ \\ 
      \midrule
      \multirow{2}{*}{\methodL} 
      & Flan-T5 XXL & \textbf{0.476} & 0.499 & 0.619 & 0.262 & \textbf{0.871} & \textbf{0.843} \\ 
      & Llama-2 13B-chat & 0.335 & \textbf{0.558} & \textbf{0.629} & \textbf{0.273} & 0.867 & 0.841 \\ 
      \midrule
      \multirow{2}{*}{\methodB}
      & Llama-2 7B-chat & 0.314 & \textbf{0.511} & 0.618 & 0.266 & \textbf{0.894} & 0.837 \\ 
      & Mistral-7B Instruct-v0.2 & \textbf{0.367} & 0.502 & \textbf{0.640} & \textbf{0.280} & 0.878 & \textbf{0.840} \\ 
      \midrule
      \multirow{2}{*}{\methodS}
      & Flan-T5 XL & \textbf{0.352} & 0.489 & \textbf{0.598} & \textbf{0.256} & 0.871 & \textbf{0.844} \\ 
      & Phi-2 & 0.302 & \textbf{0.520} & 0.565 & 0.241 & \textbf{0.879} & 0.840 \\ 
      \bottomrule
      \end{tabular}
      \begin{tablenotes}[normal, flushleft]
      \begin{footnotesize}
      \item
      In this table, ``Base Model" presents the base models used in \method models.
      On each task, the best performance of \methodL, \methodB, and \methodS when using different base models is in \textbf{bold}.
      \par
      \end{footnotesize}
      \end{tablenotes}
  \vspace{-10pt}
  \end{threeparttable}
\end{table*}


\subsection{Results on Models Tuned using All Tasks and Individual Tasks}
\label{sec:appendix:ood:joint}

As demonstrated in Table~\ref{tbl:model_ood}, when generalizing \method to OOD test, \method fine-tuned on every single task can better transfer the knowledge to the specific task than \method fine-tuned on the task combination.

\begin{table*}[!h]
\footnotesize
  \caption{Performance of Generalist and Task-specific {\method} Models in OOD Evaluation}
  \centering
  \label{tbl:model_ood}
  \begin{threeparttable}
      \begin{tabular}{
        @{\hspace{4pt}}l@{\hspace{8pt}}
        @{\hspace{8pt}}c@{\hspace{8pt}}
        @{\hspace{8pt}}c@{\hspace{8pt}}
        @{\hspace{8pt}}c@{\hspace{4pt}}
        @{\hspace{4pt}}c@{\hspace{8pt}}
        @{\hspace{6pt}}c@{\hspace{8pt}}
        @{\hspace{10pt}}c@{\hspace{8pt}}
        @{\hspace{8pt}}c@{\hspace{4pt}}
      }
      \toprule
      \multirow{2.5}{*}{Model} & \multirow{2.5}{*}{Training Tasks} & \AVE & \IRP & \SA & \SR & \AP & \AG \\ 
      \cmidrule(lr){3-8}
      && F1* & Macro F1 & Macro F1 & HR@1 & F1 & F$_{\text{BERT}}$ \\ 
      \midrule
      \multirow{2}{*}{\methodL} & Task-specific & \textbf{0.518} & 0.483 & \textbf{0.629} & 0.263 & \textbf{0.887} & 0.840 \\ 
       & Generalist & 0.335 & \textbf{0.558} & \textbf{0.629} & \textbf{0.273} & 0.867 & \textbf{0.841} \\ 
      \midrule
      \multirow{2}{*}{\methodB} & Task-specific & 
      \textbf{0.443} & \textbf{0.502} & 0.632 & 0.268 & \textbf{0.881} & \textbf{0.842}\\
      & Generalist & 0.367 & \textbf{0.502} & \textbf{0.640} & \textbf{0.280} & 0.878 & 0.840\\
      \midrule
      \multirow{2}{*}{\methodS} & Task-specific 
      & \textbf{0.362} & 0.251 & \textbf{0.583} & 0.219 & 0.856 & 0.835\\
      & Generalist & 0.302 & \textbf{0.520} & 0.565 & \textbf{0.241} & \textbf{0.879} & \textbf{0.840}\\
      \bottomrule
      \end{tabular}
      \begin{tablenotes}[normal, flushleft]
      \begin{footnotesize}
      \item
      In this table, ``Task-specific" indicates that the \method models are tuned on individual tasks;
      ``Generalist" represents tuning \method models using all tasks together.
      The best performance of generalist and task-specific \method models on each task is in \textbf{bold}.
      \par
      \end{footnotesize}
      \end{tablenotes}
  \end{threeparttable}
\end{table*}





\section{Complete Results for the Analysis of Training Data Size}
\label{sec:appendix:data}

\begin{table*}[h]
  \caption{Performance with Different Data Sizes in IND Evaluation}
  \centering
  \label{tbl:scaling_IND}
  \footnotesize
  \begin{threeparttable}
      \begin{tabular}{
        @{\hspace{4pt}}l@{\hspace{4pt}}
        @{\hspace{4pt}}r@{\hspace{4pt}}
	  @{\hspace{4pt}}c@{\hspace{4pt}}
	  @{\hspace{4pt}}c@{\hspace{5pt}}
	  @{\hspace{5pt}}c@{\hspace{4pt}}
	  @{\hspace{4pt}}c@{\hspace{4pt}}
	  @{\hspace{4pt}}c@{\hspace{4pt}}
        @{\hspace{5pt}}c@{\hspace{4pt}}
        @{\hspace{1pt}}c@{\hspace{1pt}}
        @{\hspace{6pt}}c@{\hspace{4pt}}
        @{\hspace{8pt}}c@{\hspace{8pt}}
        @{\hspace{8pt}}c@{\hspace{2pt}}
      }
      \toprule
      \multirow{2.5}{*}{Model}  
      & \multirow{2.5}{*}{Data Size}
      & \AVE & \IRP & \EM & \SA & \SR & \MPC & \PSI & \QPR & \AP & \AG \\
      \cmidrule(lr){3-12}
      && F1* & Macro F1 & F1 & Macro F1 & HR@1 & Accuracy & F1 & NDCG & F1 & F$_{\text{BERT}}$ \\
      \midrule
      \multirow{4}{*}{\methodL} 
      & 1K & 0.000 & 0.309 & 0.874 & 0.309 & 0.085 & 0.576 & 0.194 & 0.803 & 0.632 & 0.813 \\
      & 10K & 0.391 & 0.500 & \textbf{0.995} & 0.601 & 0.445 & 0.656 & 0.009 & 0.856 & 0.806 & 0.843 \\
      & 47K & 0.544 & 0.549 & \textbf{0.995} & 0.618 & 0.506 & 0.675 & 0.424 & 0.869 & \textbf{0.854} & \textbf{0.844} \\
      & 92K & \textbf{0.582} & \textbf{0.611} & \textbf{0.995} & \textbf{0.648} & \textbf{0.526} & \textbf{0.684} & \textbf{0.501} & \textbf{0.870} & 0.851 & 0.841 \\
       \midrule
      \multirow{4}{*}{\methodB} 
      & 1K  & 0.046 & 0.327 & 0.982 & 0.550 & 0.318 & 0.633 & 0.156 & 0.817 & 0.792 & 0.828 \\
      & 10K & 0.478 & 0.534 & 0.991 & 0.595 & 0.468 & 0.661 & 0.252 & \textbf{0.877} & 0.804 & 0.842 \\
      & 47K & 0.618 & \textbf{0.610} & \textbf{0.995} & \textbf{0.639} & 0.507 & 0.672 & \textbf{0.404} & 0.872 & 0.843 & \textbf{0.846} \\
      & 92K & \textbf{0.662} & 0.558 & \textbf{0.995} & \textbf{0.639} & \textbf{0.542} & \textbf{0.696} & 0.305 & 0.876 & \textbf{0.846} & 0.842 \\
       \midrule
      \multirow{4}{*}{\methodS} 
      & 1K  & 0.000 & 0.296 & 0.411 & 0.286 & 0.046 & 0.507 & 0.356 & 0.745 & 0.767 & 0.748 \\
      & 10K & 0.223 & 0.330 & 0.987 & 0.510 & 0.287 & 0.636 & 0.000 & 0.838 & 0.772 & 0.840 \\
      & 47K & 0.311 & 0.503 & \textbf{0.995} & 0.571 & 0.422 & 0.653 & 0.017 & 0.863 & 0.837 & 0.837 \\
      & 92K & \textbf{0.509} & \textbf{0.518} & 0.991 & \textbf{0.596} & \textbf{0.479} & \textbf{0.650} & \textbf{0.392} & \textbf{0.870} & \textbf{0.846} & \textbf{0.842} \\
      \bottomrule
      \end{tabular}
      \begin{tablenotes}[normal,flushleft]
      \begin{footnotesize}
      \item
      The best performance of \methodL, \methodB, and \methodS over different data sizes is in \textbf{bold}.
      \par
      \end{footnotesize}
      \end{tablenotes}
  \end{threeparttable}
\end{table*}


\begin{table*}[h]
  \caption{Performance with Different Data Sizes in OOD Evaluation}
  \centering
  \label{tbl:scaling_OOD}
  \footnotesize
  \begin{threeparttable}
      \begin{tabular}{
        @{\hspace{4pt}}l@{\hspace{4pt}}
        @{\hspace{4pt}}r@{\hspace{16pt}}
	  @{\hspace{4pt}}c@{\hspace{8pt}}
	  @{\hspace{4pt}}c@{\hspace{2pt}}
	  @{\hspace{3pt}}c@{\hspace{6pt}}
	  @{\hspace{4pt}}c@{\hspace{8pt}}
	  @{\hspace{8pt}}c@{\hspace{7pt}}
        @{\hspace{7pt}}c@{\hspace{0pt}}
      }
      \toprule
      \multirow{2.5}{*}{Model}  
      & \multirow{2.5}{*}{Data Size}
      & \AVE & \IRP & \SA & \SR & \AP & \AG \\
      \cmidrule(lr){3-8}
      && F1* & Macro F1 & Macro F1 & HR@1 & F1 & F$_{\text{BERT}}$ \\
      \midrule
      \multirow{4}{*}{\methodL} 
      & 1K  & 0.000 & 0.299 & 0.329 & 0.059 & 0.547 & 0.818 \\
      & 10K & \textbf{0.389} & 0.464 & 0.602 & 0.227 & 0.833 & \textbf{0.842} \\
      & 47K & 0.346 & 0.529 & 0.628 & 0.271 & \textbf{0.880} & 0.841 \\
      & 92K & 0.335 & \textbf{0.558} & \textbf{0.629} & \textbf{0.273} & 0.867 & 0.841 \\
       \midrule
      \multirow{4}{*}{\methodB} 
      & 1K  & 0.082 & 0.304 & 0.523 & 0.156 & 0.833 & 0.830 \\
      & 10K & 0.356 & 0.483 & 0.570 & 0.261 & 0.826 & 0.841 \\
      & 47K & \textbf{0.401} & \textbf{0.529} & \textbf{0.650} & 0.267 & \textbf{0.891} & \textbf{0.846} \\
      & 92K & 0.367 & 0.502 & 0.640 & \textbf{0.280} & 0.878 & 0.840 \\
       \midrule
      \multirow{4}{*}{\methodS} 
      & 1K  & 0.000 & 0.278 & 0.278 & 0.052 & 0.801 & 0.758 \\
      & 10K & 0.278 & 0.296 & 0.490 & 0.170 & 0.794 & \textbf{0.848} \\
      & 47K & \textbf{0.317} & 0.479 & 0.537 & 0.217 & 0.855 & 0.839 \\
      & 92K & 0.302 & \textbf{0.520} & \textbf{0.565} & \textbf{0.241} & \textbf{0.879} & 0.840 \\
      \bottomrule
      \end{tabular}
      \begin{tablenotes}[normal,flushleft]
      \begin{footnotesize}
      \item
      The best performance of \methodL, \methodB, and \methodS over different data sizes is in \textbf{bold}.
      \par
      \end{footnotesize}
      \end{tablenotes}
  \end{threeparttable}
\end{table*}


This section exhibits the comprehensive results for analyzing how the training data size influences the performance of \method models. 
With the total 92K samples in \dataset, we assess data scaling with sizes of 1K, 10K, and 47K.
The 1K and 10K samples are collected by randomly selecting 0.1K and 1K samples, respectively, from the training set of each of the 10 tasks. 
The 47K samples are assembled by randomly selecting 5K samples from each task except for \EM, for which all its 2K training samples are included. 
As shown in Table~\ref{tbl:scaling_IND} and Table~\ref{tbl:scaling_OOD}, performances exhibit an upward trend along with the increase in data size. 
Notably, the F1 score of the \PSI task experiences an initial drop followed by a subsequent rise with the data size increasing.
This behavior could be attributed to the imbalanced \PSI data, making the models tend to randomly guess when the data size is small. 
As the data size increases, the models would predict the dominant label, leading to a temporarily low F1 score. 
With continued data growth, the models acquire sufficient knowledge to generate accurate predictions. 
In general, the large-scale training data plays a crucial role in developing effective \mbox{e-commerce} LLMs, underscoring the significance of our extensive, comprehensive, and high-quality e-commerce instruction dataset \dataset.

\clearpage

\section{Case Study}
\label{sec:appendix:case}

We show the predictions from \methodL and the \mbox{best-performing} \mbox{general-purpose} LLM GPT-4 Turbo on each of the 10 tasks.
From the following examples, we note that by instruction tuning using our high-quality \dataset, \methodL could outperform GPT-4 Turbo in distinguishing products (e.g. \EM), estimating users' preferences (e.g. \SA), and understanding products (e.g. \AVE).
For example, as shown in Section~\ref{sec:appendix:case:em}, \methodL could accurately predict whether two products are the same given their title, description, manufacturer, and price, while GPT-4 Turbo struggles.
In addition, the example in Section~\ref{sec:appendix:case:sr} indicates that compared to GPT-4 Turbo, \methodL could better estimate users’ preferences based on their historical interactions. 
Moreover, the example in Section~\ref{sec:appendix:case:mpc} shows that \methodL could correctly understand that 
DJ light is equivalent to light that will change based on the music. 
In contrast, without instruction tuning, GPT-4 Turbo may lack this knowledge and fail the example.
%

\subsection{Attribute Value Extraction (\AVE)}
\label{sec:appendix:case:ave}

\begin{itemize}
    \item Instruction: 
        \begin{itemize}
            \item Given the title, description, feature, price, and brand of a product and a set of target attributes, extract the value of each target at tribute from the product information. Output the extracted value and the corresponding source (e.g., title or feature) denoting where the value is extracted.
        \end{itemize}
    \item Input: 
        \begin{itemize}
            \item Product title: LibbySue-Airy, Lightweight Knit Winter Scarf in Opulent Earth Tones
            \item Product description: The lightweight weave of this soft and airy knit scarf will keep you stylishly ready for any brisk day
            \item Product feature: 100\% Acrylic, A lightweight weave, Soft and airy knit scarf, Will keep you stylishly ready for any brisk day, Opulent Earth Tones, 78 inches long by 12 inches wide with 7 inch fringe, Package Dimensions: 5.8 x 5.7 x 3.9 inches, Shipping Weight: 4 ounces, ASIN: B00FM6HPAU, Date first listed on Amazon: November 6, 2011, Average Customer Review: 5.0 out of 5 stars 1 customer review
            \item Product brand: LibbySue
            \item Target attributes: Specialty Style
        \end{itemize}
    \item Output: 
        \begin{itemize}
            \item Ground truth: Attribute: specialty style; Value: None, Source: None.
            \item \method: Attribute: specialty style; Value: None, Source: None.
            \item GPT-4 Turbo: Attribute: specialty style; Value: airy, lightweight knit, Source: product title.
        \end{itemize}
\end{itemize}

\subsection{Product Matching (\EM)}
\label{sec:appendix:case:em}
\begin{itemize}
    \item Instruction: 
        \begin{itemize}
            \item Compare the details of two given products to determine if they are identical. Output yes if they are identical or no otherwise.
        \end{itemize}
    \item Input: 
        \begin{itemize}
            \item Product 1: title - omniweb 5.0, description - NaN, manufacturer - omni group, price - 29.99
            \item Product 2: title - omni web 5.0, description - sure you can use a standard web browser with standard features. but you didn't choose a standard software experience - you chose the mac. why not try a browser built just for discriminating people with fabulous taste like yourself? omniweb5..., manufacturer - NaN, price - 23.99
        \end{itemize}
    \item Output: 
        \begin{itemize}
            \item Ground truth: Yes
            \item \method: Yes
            \item GPT-4 Turbo: No
        \end{itemize}
\end{itemize}

\subsection{Product Relation Prediction (\IRP)}
\label{sec:appendix:case:irp}
\begin{itemize}
    \item Instruction: 
        \begin{itemize}
            \item Given the title of two products, predict if the two products are similar, if the two products will be purchased or viewed together. Answer only from the options.
        \end{itemize}
    \item Input: 
        \begin{itemize}
            \item Product 1: Kenable Internal Memory Card Reader for 5.25 CD/DVD Bay With USB Port BLACK
            \item Product 2: CORSAIR Carbide 100R Mid-Tower Case
        \end{itemize}
    \item Options: 
        \begin{itemize}
            \item A: Users who buy product 1 may also buy product 2.
            \item B: Users who view product 1 may also view product 2.
            \item C: The product 1 is similar with the product 2.
        \end{itemize}
    \item Output: 
        \begin{itemize}
            \item Ground truth: B
            \item \method: B
            \item GPT-4 Turbo: A
        \end{itemize}
\end{itemize}

\subsection{Sentiment Analysis (\SA)}
\label{sec:appendix:case:sa}
\begin{itemize}
    \item Instruction: 
        \begin{itemize}
            \item Carefully assess the user's review for any strong expressions of sentiment, either positive or negative. Based on your analysis, select the most fitting sentiment option from the provided list as output.
        \end{itemize}
    \item Input: 
        \begin{itemize}
            \item This visor CD holder is a great addition for any car.  It folds shut so CDs don't slide out.  It fits well on any visor.
        \end{itemize}
    \item Options: 
        \begin{itemize}
            \item A: very positive
            \item B: positive
            \item C: neutral
            \item D: negative
            \item E: very negative
        \end{itemize}
    \item Output: 
        \begin{itemize}
            \item Ground truth: A
            \item \method: A
            \item GPT-4 Turbo: B
        \end{itemize}
\end{itemize}

\subsection{Sequential Recommendation (\SR)}
\label{sec:appendix:case:sr}
\begin{itemize}
    \item Instruction: 
        \begin{itemize}
            \item Given the products the user has purchased in history, rank the items in the listed options and output the item that the user is most likely to purchase next. Answer from one of the options.
        \end{itemize}
    \item Input: 
        \begin{itemize}
            \item 1st: Bbox A392-10CP Dual 10" Sealed Carpeted Subwoofer Enclosure - Fits 1999-2007 Ford F250/350/450 Crew Cab. Electronics. Car \& Vehicle Electronics. b.box.
            \item 2nd: Smatree Batteries Charger Kit for GoPro Hero 1/2 Digital Camera. Electronics. Camera \& Photo. Smatree.
            \item 3rd: Dolica WT-1003 67-Inch Lightweight Monopod. Electronics. Camera \& Photo. Dolica.
            \item 4th: SunFounder Sidekick Basic Starter Kit w/Breadboard, Jumper wires, Color Led, Resistors, Buzzer For Arduino UNO R3 Mega2560 Mega328 Nano - Including 42 Page Instructions Book...
            \item 5th: Winait 5MP Mini 5mp Worlds Smallest Hd Digital Video Camera Spy Camera Video Recorder Hidden Cam DV DVR with 1280 X 960 Resolution. Electronics. Camera...
            \item 6th: RoadPro RPPS-220 Platinum Series 12V 3-Pin Plug Fused Replacement CB Power Cord. Electronics. Accessories \& Supplies. RoadPro.
            \item 7th: Transcend USB 3.0 SDHC / SDXC / microSDHC / SDXC Card Reader, TS-RDF5K (Black). Electronics. Computers \& Accessories. Transcend.\
        \end{itemize}        
    \item Options: 
        \begin{itemize}
            \item A: Sony Bloggie Live(MHS-TS55) Video Camera with 4x Digital Zoom, 3.0-Inch Touchscreen LCD and WiFi Connectivity (2012 Model). Electronics. Camera \& Photo. Sony.
            \item B: Gold Tip Expedition Hunter 5575 Dozen Black Shafts. Sports \& Outdoors. Sports \& Fitness. Gold Tip.
            \item C: One Direction Purple Zebra Print Fleece Throw Blanket. Home \& Kitchen. Bedding. 1D Media Ltd.
            \item D: Absolute DAG15 Dual 15-Inch Angle Ported MDF Enclosure. Electronics. Car \& Vehicle Electronics. Absolute.
            \item E: Adesso iMouseE1 - Vertical Ergonomic Illuminated Optical 6-Button USB Mouse - Right Hand Orientation. Electronics. Computers \& Accessories. Adesso.
            \item F: AGPtek AC Power Home Wall Charger Adapter For Microsoft Surface Windows RT Surface 2 Tablet. Electronics. Computers \& Accessories. AGPTEK.
            \item ...
            \item O: SanDisk Ultra 32GB UHS-I/Class 10 Micro SDHC Memory Card With Adapter - SDSDQUAN-032G-G4A. Electronics. Computers \& Accessories. SanDisk.
            \item P: UCEC 5.25 Inch Front Panel USB Hub with 2-Port USB 3.0 \& 2-Port USB 2.0 \& HD Audio Output Port \& Microphone Input Port for...
            \item Q: GreenLife Soft Grip 11" Ceramic Non-Stick Open Wok, Burgundy. Home \& Kitchen. Kitchen \& Dining. GreenLife.
            \item R: Bulova Frank Lloyd Wright Luxfer Prism Wall Clock, 14", Bronze. Home \& Kitchen. Home Dcor. Bulova.
            \item S: Ortopad Girls Eye Patching Reward Posters: 1 Princess Poster, 1 Butterfly Poster. Home \& Kitchen. Wall Art. Ortopad.
            \item T: HIS H775F1GD Radeon HD 7750 1GB (128bit) GDDR5 Displayport HDMI DVI (HDCP) PCI Express X16 3.0 Graphics Cards. Electronics. Computers \& Accessories. HIS.
        \end{itemize}
    \item Output: 
        \begin{itemize}
            \item Ground truth: O
            \item \method: O
            \item GPT-4 Turbo: A
        \end{itemize}
\end{itemize}

\subsection{Multiclass Product Classification (\MPC)}
\label{sec:appendix:case:mpc}
\begin{itemize}
    \item Instruction: 
        \begin{itemize}
            \item Determine the relevance between the query and the product title provided, and select your response from one of the available options.
        \end{itemize}
    \item Input: 
        \begin{itemize}
            \item Query: dj lights in the car
            \item Product title: Sanhezhong Interior Car Lights, LED Car Strip Lights with Waterproof Design, 48 LED Remote Control Car Light Kit, Music Sync Under Dash Car Lighting with Car Charger, DC 12V
        \end{itemize}
    \item Options: 
        \begin{itemize}
            \item A: The product is relevant to the query, and satisfies all the query specifications.
            \item B: The product is somewhat relevant. It fails to fulfill some aspects of the query but the product can be used as a functional substitute.
            \item C: The product does not fulfill the query, but could be used in combination with a product exactly matching the query.
            \item D: The product is irrelevant to the query.
        \end{itemize}
    \item Output: 
        \begin{itemize}
            \item Ground truth: A
            \item \method: A
            \item GPT-4 Turbo: B
        \end{itemize}
\end{itemize}

\subsection{Product Substitute Identification (\PSI)}
\label{sec:appendix:case:psi}
\begin{itemize}
    \item Instruction: 
        \begin{itemize}
            \item Assess the relevance of a product to a given query by determining if it can function as a substitute, despite not fully meeting certain aspects of the query. Provide a binary output of yes or no based on this evaluation.
        \end{itemize}
    \item Input: 
        \begin{itemize}
            \item Query: iphone 7 plus case otterbox.
            \item Product: OtterBox SYMMETRY CLEAR SERIES Case for iPhone 8 Plus \& iPhone 7 Plus (ONLY) - Retail Packaging - EASY BREEZY (CLEAR/EASY BREEZY).
        \end{itemize}
    \item Output: 
        \begin{itemize}
            \item Ground truth: No
            \item \method: No
            \item GPT-4 Turbo: Yes
        \end{itemize}
\end{itemize}

\subsection{Query Product Ranking (\QPR)}
\label{sec:appendix:case:qpr}
\begin{itemize}
    \item Instruction: 
        \begin{itemize}
            \item Rank the products A, B, C, ... based on their relevance to the provided query, and produce a ranked list with the most relevant product positioned at the top of the list.
        \end{itemize}
    \item Input: 
        \begin{itemize}
            \item Query: everyone loves raymond dvd complete series
            \item Product A: Everybody Loves Raymond: The Complete Series - Seasons 1-9
            \item Product B: Perry Mason: The Ninth and Final Season, Vol. 2
        \end{itemize}
    \item Output: 
        \begin{itemize}
            \item Ground truth: A, B
            \item \method: A, B
            \item GPT-4 Turbo: A
        \end{itemize}
\end{itemize}

\subsection{Answerability Prediction (\AP)}
\label{sec:appendix:case:ap}
\begin{itemize}
    \item Instruction: 
        \begin{itemize}
            \item Given a question and its related document, determine if the question is answerable by analyzing the information in the document. Output yes if the document addresses the question, or no otherwise.
        \end{itemize}
    \item Input: 
        \begin{itemize}
            \item Question: if i order 10 of these, are the keys the same for each lock? In other words, can I use any of the keys to open any of the locks? Thanks!
            \item Documents: 
            \begin{itemize}
                \item Document 1: I purchased two of these locks and within two uses with each one I had trouble opening the lock. The third time and I cannot get either opened at all. They are stuck                on the tree. I'll order a different brand and then go cut them off. They keys each turn but they won't come off. Wish I had read the reviews.
                \item Document 2:  Locks are great my only complaint is I ordered two and I wanted them keyed the same, there was no where in placing the order to specify this and they sent two different keys. Otherwise all is good.
                \item Document 3: I like the Python. It's easy to use \& I feel it is well made. The problem is I use multiple game cameras \& you can not buy the locks keyed alike. My local locksmith who is a Master Lock dealer said can not order them alike. I called Master Locks corporate office-and was told the same. If you need one they're great. I'm forced to look elsewhere. 
                \item Document 4: ...
            \end{itemize}
        \end{itemize}
    \item Output: 
        \begin{itemize}
            \item Ground truth: Yes
            \item \method: Yes
            \item GPT-4 Turbo: No
        \end{itemize}
\end{itemize}

\subsection{Answer Generation (\AG)}
\label{sec:appendix:case:ag}
\begin{itemize}
    \item Instruction: 
        \begin{itemize}
            \item Answer the given question by extracting information from the supporting document.
        \end{itemize}
    \item Input: 
        \begin{itemize}
            \item Question: It says it can support 2 DIMMs and 8GB of RAM.  Does this mean I can use a single 8gb stick, or do I have to use 2x4?
            \item Documents: 
            \begin{itemize}
                \item Document 1: Not going to write a full review here, it's a cheap motherboard, it's \$50, expect to get what you pay for. I bought this from another retailer, but wanted everyone to know that this board does in-fact support 16GB of RAM. I am using this board in a home lab VMware ESXi host, and I needed 16GB at least. Here is what you need to do: (1) Update bios to at least 1303 (here is linky) [...] (2) Use RAM on supported RAM list (here is linky)[...] I am using 16GB of Corsair VengeanceCorsair Vengeance  16GB (2x8GB)  DDR3 1600 MHz (PC3 12800) Desktop Memory (CMZ16GX3M2A1600C10)and it works like a champ.
                \item Document 2: I bought this mobo to replace my ASUS MVA2 HDMI series when upgrading my system. As said in other reviews it can definitely use a 125 watt CPU as I checked the manual before I started my upgrade. Originally this was designed for Phenom 2 X6 cpus which run over 95 watts typically on load.On my rig I'm using an AMD FX8350, 8gb of Kingston DDR31600 ram (2 sticks of 4gb in each slot dual channel), ASUS AMD Radeon 7750 1gb, Kingston 120gb ssd, 1tb Western Digital hdd, Windows XP 32bit and 7 64 bit, in an Antec desktop case with an upgraded Antec 660 watt power supply.
                \item Document 3: This is the PLUS version and will take a 125W processor (I have a AMD 965 quad core Deneb). It will take 8GB RAM in each of the two slots. It's a good board with all the Asus extras on it. Some considerations - the legacy connectors on the back might be useful for you or not. Most people would prefer extra USB's (or a couple of fast USB). The front panel connector is on the side next to the USB connectors so if you have a short panel connector cord(s) that *might* be an issue for you.
                \item Document 4: Always check the manufacturer's website for reliable specs. I built a computer using this board and it works perfectly.Reviewer who called this mobo ``junk" and gave it a ``1" was attempting to use a CPU this board isn't rated for- board rated for up to 95W cpu. No mobo will run with wrong RAM, CPU, inadequate power supply, etc. Double check all components, OS, peripherals, BIOS version and settings, etc. for compatibility.Make sure you know what you're doing before attempting to build your own computer or you can end up wasting a lot of money.
                \item Document 5: This board is an excellent choice if you are working on a budget and are making a relatively low end system. I used it along with an FX-6100 processor and have had no problems.I would recommend this to anyone building their first system and/or hoping to keep it a cheap build.I would not recommend this as an upgrade but rather a starting point.pros:cheaphigh connectabilitysmallgood SATA port placement.excellent support and manualsimplecons:no USB 3.0 or SATA IIIno IDEonly 4 USB ports on boardlimited ram supportonly PCIe 2.0 x16 @ 8xonly 2 fan connectors (one for cpu and one for case)
                \item Document 6: ...
            \end{itemize}
        \end{itemize}
    \item Output: 
        \begin{itemize}
            \item Ground truth: You can use 8Gb modules, there's only 2 DIMMS. I'm using Corsair DDR 3 16GB kit of 2 8Gb1600Mhz
            \item \method: You can use 2 sticks of 4GB or 1 stick of 8GB.
            \item GPT-4 Turbo: You can use a single 8GB stick.
        \end{itemize}
\end{itemize}

%

\end{document}